\documentclass{article}

\PassOptionsToPackage{numbers, compress}{natbib}
 \usepackage[preprint]{neurips_2026}

\usepackage[utf8]{inputenc} 
\usepackage[T1]{fontenc}    
\usepackage{hyperref}       
\usepackage{url}            
\usepackage{booktabs}       
\usepackage{amsfonts}       
\usepackage{nicefrac}       
\usepackage{microtype}      
\usepackage{xcolor}         

\usepackage{amsmath}
\usepackage{amssymb}
\usepackage{mathtools}
\usepackage{amsthm}
\usepackage{subcaption}
\usepackage{graphicx}
\usepackage{adjustbox} 
\usepackage{listings}
\usepackage{tablefootnote}
\usepackage[many]{tcolorbox} 
\usepackage{xcolor}
\usepackage{longtable}
\usepackage{array}
\usepackage{calc}
\usepackage{pifont}
\usepackage{multirow} 
\usepackage{makecell}
\usepackage{placeins}
\usepackage{algorithm}
\usepackage{algorithmic}
\usepackage{enumitem}
\usepackage[table]{xcolor}
\newtcolorbox{examplecard_2}[1][]{colframe=black, colback=black!3!white, title=#1}
\newtcolorbox{correctcard}[1][]{
    colframe=brown!50!black, 
    colback=yellow!10!white, 
    title=#1
}
\newcommand{\cmark}{\textbf{\textcolor{teal}{\ding{51} Correct}}} 
\newcommand{\xmark}{\textbf{\textcolor{red}{\ding{55} Incorrect}}} 
\usepackage[T1]{fontenc}  
\usepackage[varqu]{zi4}
\newcommand{\llmtext}[1]{%
    \begin{minipage}[t]{\linewidth}%
      \fontsize{8.5pt}{10.2pt}\selectfont\ttfamily\raggedright #1
    \end{minipage}%
}
\definecolor{oprange}{HTML}{E2F0D9} 
\newcommand{\cg}{\cellcolor{oprange}}
\newcommand{\ccmark}{\ding{51}}
\newcommand{\xxmark}{\ding{55}}
\usepackage{etoc}
\usepackage{amssymb}
\usepackage{tikz}
\usepackage{dsfont}
\usepackage{xspace}
\usepackage{hyperref}
\DeclareRobustCommand{\slant}[2]{%
  \tikz[baseline=(X.base), xslant=tan(#1)]
  \node[inner sep=0pt, outer sep=0pt](X){#2};
  \kern-0.05em
}

\hypersetup{hidelinks}

\newcommand{\disableaddcontentsline}{%
  \let\savedaddcontentsline\addcontentsline 
  \renewcommand{\addcontentsline}[3]{}
}
\newcommand{\enableaddcontentsline}{%
  \let\addcontentsline\savedaddcontentsline
}
\usepackage{color}
\usepackage{wrapfig}
\makeatletter
\newcommand\blfootnote[1]{%
  \begingroup
  \def\@thefnmark{}%
  \@footnotetext{#1}%
  \endgroup
}
\usepackage{fontawesome5}   
\usepackage{xcolor}
\definecolor{linkcolor}{HTML}{0366D6}

\makeatother

\title{Learning Rate Matters: Vanilla LoRA May Suffice for LLM Fine-tuning}

\author{Yu-Ang Lee$^{1}$\quad Ching-Yun Ko$^{2}$\quad Pin-Yu Chen$^{2}$\quad Mi-Yen Yeh$^{1,3}$\\
$^{1}$National Taiwan University\quad $^{2}$IBM Research\quad $^{3}$Academia Sinica\\
\\
\centering Project Page: \href{https://github.com/yuang-lee/lr-matters-lora}%
     {\textcolor{linkcolor}{\faGithub\ \texttt{yuang-lee/lr-matters-lora}}}
}

\begin{document}

\disableaddcontentsline

\maketitle

\blfootnote{%
\parbox[t]{0.95\textwidth}{%
\noindent Yu-Ang Lee is in the Data Science Degree Program, National Taiwan University and Academia Sinica.\\
\noindent Emails: \texttt{r12946015@ntu.edu.tw}, \texttt{cyko@ibm.com}, \texttt{pin-yu.chen@ibm.com}, \texttt{miyen@iis.sinica.edu.tw}
}}

\begin{abstract}
Low-Rank Adaptation (LoRA) is the prevailing approach for efficient large language model (LLM) fine-tuning. 
Building on this paradigm, recent studies have proposed alternative initialization strategies, architectural modifications, and optimization adjustments, reporting substantial improvements over vanilla LoRA.
However, these gains are often demonstrated under fixed or narrowly tuned hyperparameter settings, despite the
known 
sensitivity of neural networks to training configurations. In this work, we systematically re-evaluate 
nine
representative LoRA variants alongside vanilla LoRA through extensive hyperparameter searches
over learning rate, batch size, rank, and training duration. 
Across tasks spanning mathematical reasoning, commonsense reasoning, code generation, and instruction following at diverse model scales, we find that different LoRA methods favor distinct learning rate ranges.
Crucially, once learning rates are properly tuned, all methods achieve similar peak performance (within 1--2\%), with only subtle rank-dependent behaviors. 
These results suggest that vanilla LoRA remains a competitive baseline and that improvements reported under a single training configuration may not reflect consistent methodological advantages.
Finally, a second-order analysis attributes the differing optimal learning rate ranges to variations in the largest Hessian eigenvalue, aligning with classical learning theories. 
\end{abstract}

\section{Introduction}

\begin{figure}[h]
    \centering
    \includegraphics[width=0.985\linewidth]{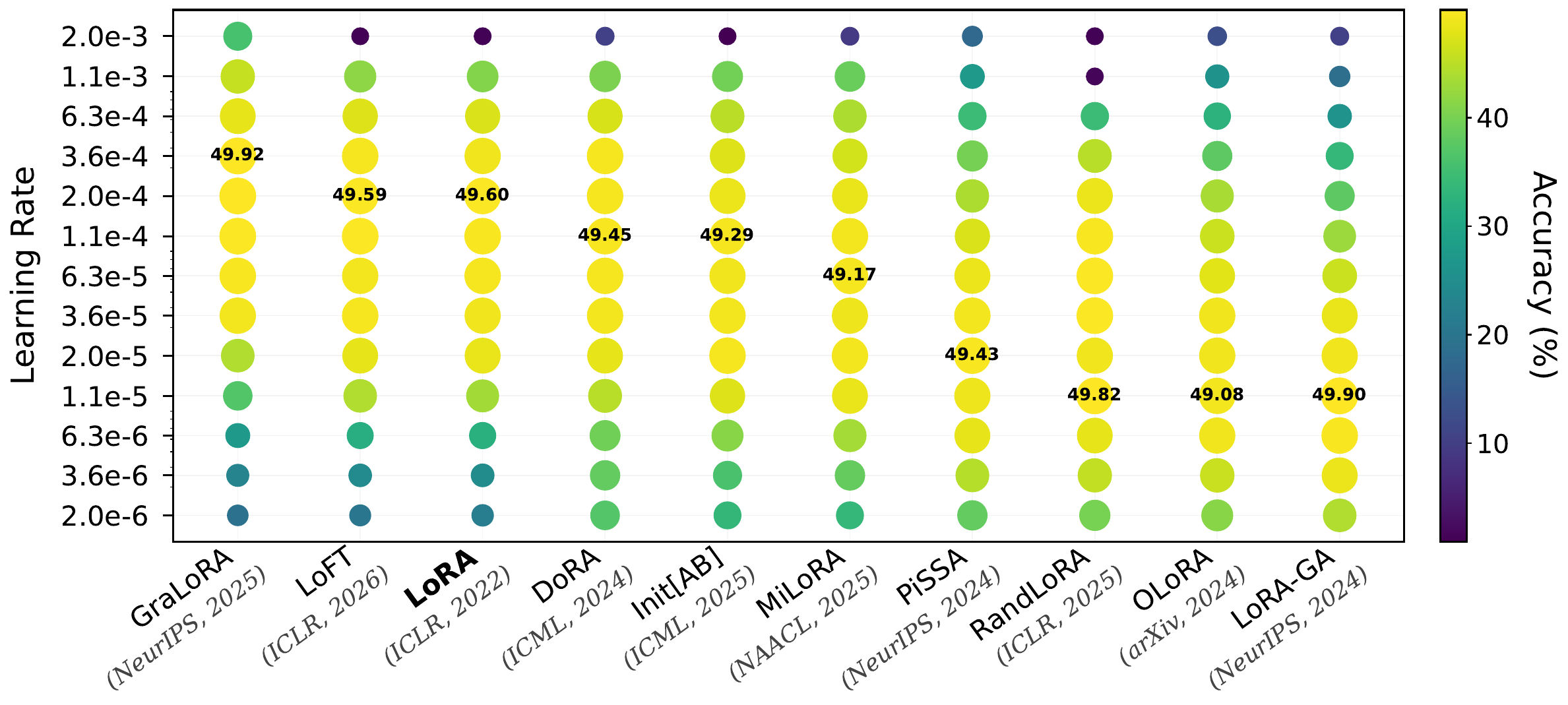}
    \vspace{-0.05em}
    \caption{Performance of Qwen3-0.6B fine-tuned on mathematical reasoning tasks under rank 128 and batch size 64 across learning rates. Different methods reach a similar performance level once the learning rate is properly tuned. Each point is averaged over three training runs. We annotate the peak accuracy of each method and sort methods by their optimal learning rate ranges. Results for other model--task combinations and training setups are reported in Sec.~\ref{sec:main_results} and Appendix Sec.~\ref{sec:additional_ablations}.} 
    \label{fig:main-evidence}
\end{figure}

Despite the rapidly growing capabilities of pretrained large language models (LLMs), fine-tuning remains a fundamental step for adapting these models to specialized applications in diverse domains such as medicine~\citep{anisuzzaman2025fine} and finance~\citep{djagba2025exploring}. However, modern LLMs typically contain billions of parameters, making full-parameter fine-tuning (Full FT) prohibitively expensive in terms of memory and computation. These constraints have motivated sustained research interest in developing parameter-efficient fine-tuning (PEFT) methods, which allow task-specific learning while updating only a small fraction of parameters.

Even though PEFT methods span diverse design paradigms, ranging from prompt-based approaches~\cite{li2021prefix,lester2021power} to adapter-based methods~\cite{houlsby2019parameter}, low-rank adaptation (LoRA), introduced by~\citet{hu2022lora}, has emerged as the de facto standard.
Inspired by the low intrinsic dimensionality observed in pretrained models~\citep{li2018measuring}, \citet{hu2022lora} hypothesized that task-specific parameter updates can be well approximated by low-dimensional structures. Consequently, they inject pairs of trainable decomposition matrices into selected layers while keeping the pretrained weights frozen. After training, these learned low-rank adapters can be merged into the original backbone, thereby incurring no additional inference latency.

Even with its popularity, LoRA has been shown to underperform Full FT on challenging tasks in programming and mathematics~\cite{biderman2024lora}. 
This gap has in turn spurred recent efforts toward advanced LoRA variants~\cite{zhu2025survey}, with promising performance improvements reported. On Llama~\cite{llama2}, for example, PiSSA~\cite{pissa} presented around a 10\% accuracy improvement on GSM8K~\citep{gsm8k} by modifying LoRA initialization strategies, while DoRA~\cite{dora} reported substantial gains of 37.2\% on commonsense reasoning tasks by separately learning magnitude and directional updates of pretrained weight matrices. More recently, LoFT~\cite{tastan2025loft} reported a further 40.0\% improvement over DoRA in the same evaluation setting by aligning LoRA's optimizer dynamics with those of full FT.

\begin{wrapfigure}{r}{0.45\columnwidth}
    \vspace{-8pt}
    \centering
    \includegraphics[width=0.45\columnwidth]{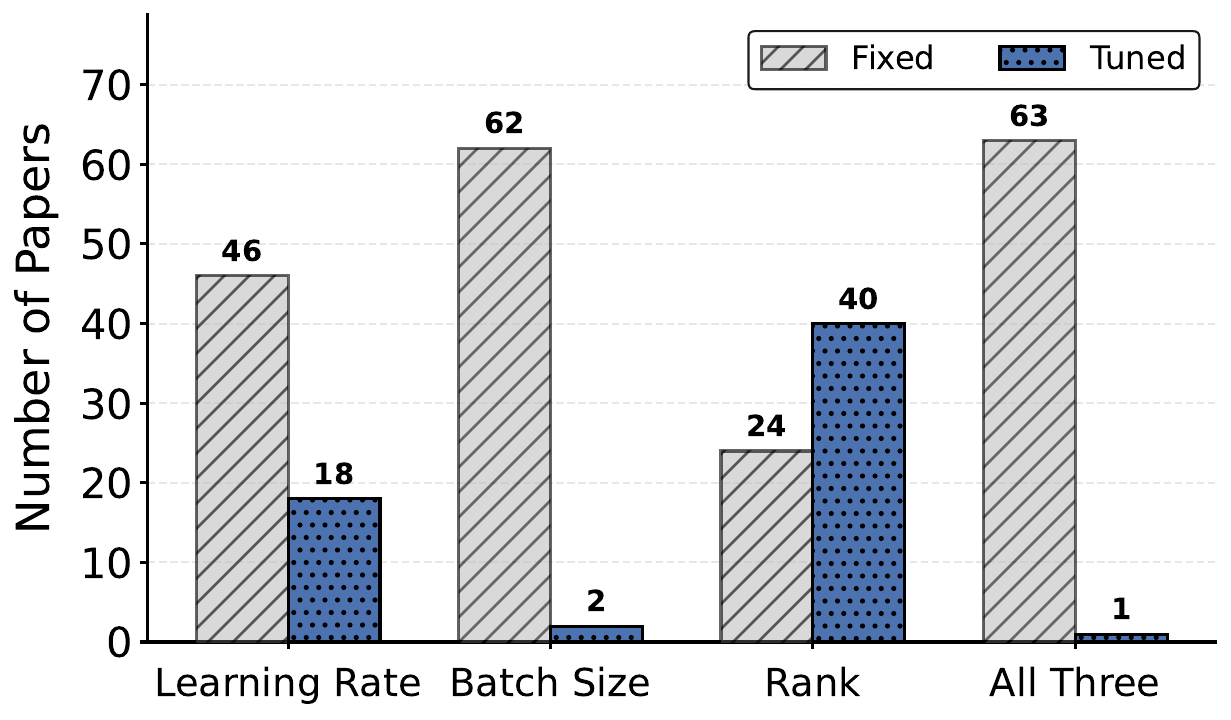}
    \caption{Frequency of advanced LoRA-based PEFT studies, categorized by whether learning rate or batch size tuning was applied and whether comparisons with vanilla LoRA across different ranks were conducted. 
    Refer to Appendix Sec.~\ref{sec:prior_studies_hyperparameter} for detailed data counts.
    } \label{fig:frequency_statistics}
    \vspace{-8pt}
\end{wrapfigure}
Yet, the results in a majority of work along this line were obtained with hyperparameters directly inherited from prior studies, or only 
tuned
in a narrow range.
To be specific, in Figure~\ref{fig:frequency_statistics}, we collect 54 LoRA publications from major AI conferences and journals over the past three years, and
additionally 
include 10
more
high-impact or recently released preprints, to investigate whether their training involved tuning key hyperparameters—namely, learning rate, batch size, and rank. 
The statistics over a total of 64 prior studies clearly reveal that hyperparameter search is not a standard practice in the field, with only one paper simultaneously considering three hyperparameters and fewer than 30\% tuning the learning rate.
These findings raise questions about the extent to which the reported gains can be attributed to genuine methodological improvements, particularly given the well-known sensitivity of neural networks to training configurations~\cite{lecun2002efficient,bengio2012practical}\footnote{Among prior LoRA studies involving hyperparameter tuning, we also observed that many reported only the final optimal performance, leaving it unclear whether the adopted search ranges covered the optimal configurations for each method.}.
This is especially critical when it comes to LoRA on LLMs, where careful learning rate tuning 
has been demonstrated
to be essential for eliciting strong performance, and optimal settings are contingent on both the base model and the target problem~\cite{biderman2024lora,schulman2025lora,yan2025plora}.

To address the above concern, we select nine representative advanced LoRA variants and conduct a large-scale hyperparameter search,
benchmarking them against vanilla LoRA in a head-to-head manner. Under a unified evaluation protocol, we surprisingly find that once the learning rate is properly tuned, all methods peak at similar performance levels, exhibiting no systematic advantages over 
vanilla LoRA. 
For example, in Figure~\ref{fig:main-evidence}, we fine-tune Qwen3~\cite{qwen3} using all ten LoRA methods, with the learning rate varied over three orders of magnitude; under a fixed rank of 128 and batch size of 64, all methods achieve accuracies within a narrow 0.84\% range.
Moreover, different methods operate within disparate learning rate ranges (e.g., a 10$\times$ difference between PiSSA and LoRA in Figure~\ref{fig:main-evidence}), suggesting that success under a single training configuration cannot be taken as evidence of robust and reliable improvements.
This phenomenon is not isolated; we consistently observe such performance parity across four task types over diverse LoRA ranks, training durations, and model scales ranging from 0.6B to 13B (e.g., Appendix Figure~\ref{fig:main-evidence-gemma-cs-r4} presents similar results on commonsense reasoning tasks with Gemma-3~\cite{gemma3}). 
Notably, within these marginal performance variations, rank-dependent behaviors emerge: 
some advanced variants may slightly outperform LoRA at higher ranks while lagging behind at lower ones (or vice versa), highlighting the importance of verifying improvements across the entire rank spectrum.
By delving into the fundamentals of learning theories~\cite{lecun1992automatic, lewkowycz2020large}, we provide an explanation for the importance of tuning learning rates during LoRA fine-tuning and uncover the reasons behind different desirable learning rate ranges among various LoRA 
methods. 
Specifically, we demonstrate that PiSSA~\cite{pissa}, OLoRA~\cite{buyukakyuz2024olora}, and LoRA-GA~\cite{loraga} exhibit significantly larger maximum Hessian eigenvalues compared to vanilla LoRA, which theoretically justifies their requirement for a lower learning rate. 
Based on the extensive tuning experiments, we also derive five~\emph{practical heuristics} for hyperparameter tuning in LoRA-based methods, 
particularly regarding how desirable learning-rate ranges interact with batch size 
(\textbf{\emph{I}}--\textbf{\emph{II}})
and how such ranges can be inferred from the eigenvalues of the loss Hessian 
(\textbf{\emph{III}}).
In addition, we show how performance improvements can be expected across LoRA ranks 
(\textbf{\emph{IV}})
and training durations (\textbf{\emph{V}}).

In summary, our work reveals the insufficiency of hyperparameter tuning in many prior LoRA studies and provides a systematic empirical re-evaluation of their best achievable performance. By explaining the observed performance trends and differences in optimal learning rate ranges via Hessian analysis, we hope to encourage future LoRA research to adopt more comprehensive hyperparameter tuning protocols.
Meanwhile, 
with the provided practical guidelines on LoRA hyperparameters, we aim to help practitioners with limited computational resources avoid unnecessarily exhaustive hyperparameter searches.
Concretely, the main contributions of this paper are organized around the following questions:


\begin{itemize}[leftmargin=*]
    \item \emph{\textbf{What is the problem?}} We conduct a comprehensive audit of advanced LoRA PEFT studies and identify a recurring issue: the majority of works lack thorough hyperparameter tuning, despite this being a standard requirement. In fact, according to Figure~\ref{fig:frequency_statistics}, only 1 out of 64 papers simultaneously considers three hyperparameters, while 46 present results under a fixed learning rate.
    \item \emph{\textbf{Why does it matter?}} 
    Without proper hyperparameter tuning, conclusions may be ungrounded.
    Concretely, through extensive experimentation, 
    we demonstrate that while different methods require distinct optimal learning rate ranges, 
    they yield comparable peak performance when configured to their optimal settings. 
    For example, on Qwen3-0.6B, averaged over three runs, 
    the top-performing method (GraLoRA) leads the runner-up (LoRA-GA) by only 0.02\%, 
    and the least effective method (OLoRA) by 0.84\% 
    (cf. Figure~\ref{fig:main-evidence}).
    \item 
    \emph{\textbf{Which learning rate to use?}}
    By analyzing the eigenvalues of the loss Hessian across various initialization-based
    LoRA variants, we find that the optimal learning rate is generally negatively correlated with the maximum eigenvalue, aligning with classical learning theories. 
    \item \emph{\textbf{How can we tune LoRA methods efficiently?}} 
    Since fair comparison requires method-specific tuning but exhaustive searches are computationally expensive,  we identify practical heuristics across learning rate, batch size, LoRA rank, and training duration.  Together with the Hessian eigenvalue analysis, these findings help narrow the hyperparameter search space.
\end{itemize}

\section{Related Work and Background}\label{sec:related_work}
\subsection{Related Work}
\subsubsection{Systematic Empirical Re-evaluation of Prior Claims}
Incomplete performance evaluation remains a persistent concern~\cite{sculley2018winner, lipton2019troubling}.
For instance, \citet{melis2017state} revealed that two published improvements to the vanilla LSTM~\cite{hochreiter1997long}, originally attributed to complex network designs, were in fact due to more careful hyperparameter tuning. Under a fair tuning protocol, the standard LSTM emerged as the best-performing architecture. 
In the same vein,~\citet{lin2023linear} pointed out that simple baselines such as linear SVM~\cite{boser1992training} are competitive with BERT~\cite{devlin2019bert}-based methods for text classification, sometimes even outperforming them with a clear gap.
More recently,~\citet{rheude2025fusion} reported a striking finding in the multimodal learning literature: although new multimodal architectures are continually proposed with claims that they outperform prior baselines, they often fail to surpass standard unimodal approaches given fair comparisons and statistical correctness.
Such systematic empirical re-evaluation has also been done across diverse machine learning subfields. Examples include not only traditional topics such as image classification~\cite{chatfield2014return}, graph neural networks~\cite{shchur2018pitfalls,luo2025can}, generative adversarial networks~\cite{lucic2018gans}, recommender systems~\cite{konstan2013toward, ferrari2019we,benigni2026diffusion}, metric learning~\cite{musgrave2020metric}, and neural network pruning~\cite{blalock2020state}, but also more recent areas like optimizers~\cite{choi2019empirical,sivaprasad2020optimizer,schmidt2021descending}, reinforcement learning~\cite{eimer2023hyperparameters}, preference optimization~\cite{ahrabian2025practical}, 
and model merging~\cite{hitit2025systematic, de2026task}. 

While empirical studies benchmarking LoRA with other PEFT methods such as prefix tuning~\cite{li2021prefix} and BitFit~\cite{zaken2022bitfit} exist~\cite{he2021towards,hu2023llm, zheng2024llamafactory, mannisto2025comparative,  belanec2026peft, xu2026parameter}, few prior studies specifically focus on comparing LoRA and its advanced variants. 
More concerning, training hyperparameters in many of these works were kept fixed without method-specific optimization.
Therefore, practitioners are left without clear and reliable guidance when choosing LoRA-based methods.

\subsubsection{LoRA Hyperparameter Tuning}
Theories regarding LoRA's lack of Lipschitz smoothness
~\cite{sun2024improving, malinovsky2024randomized} and its spurious loss landscape~\cite{liu2025optimization} point toward its intrinsic sensitivity to hyperparameter variations. 
Consequently, many research efforts have been invested in finding optimal training setups, such as learning rate~\cite{hayou2024lora+, chen2026learning}, rank~\cite{zhang2025beyond}, initializations~\cite{hayou2024impact}, scaling factor~\cite{kalajdzievski2023rank}, dropout~\cite{lin2024lora, wang2024lora_dropout}, and adapter placements~\cite{fomenko2024note, hayou2025plop}. 
Despite these insights into individual hyperparameters, establishing unified configuration guidelines remains an ongoing pursuit. For example, recent works have sought to derive practical ``rules of thumb'' through extensive, joint evaluations across multiple hyperparameter dimensions~\cite{biderman2024lora,schulman2025lora}. Addressing the computational bottleneck of such extensive searches, another line of work has focused on improving the efficiency of LoRA hyperparameter tuning, either by developing hyperparameter optimization algorithms~\cite{tribes2023hyperparameter, oliver2024crafting, sengupta2024robust, seong2026efficient} or by adopting system-level approaches that optimize hardware resources to maximize training throughput~\cite{yan2025plora,zuo2026alto}.
Although prior work has optimized LoRA hyperparameters, only a few concurrent studies have begun to examine whether LoRA and its variants require distinct hyperparameter settings~\cite{zhang2025primacy, he2026unified, lee2026beware}. These studies, however, differ from ours in scope.
Specifically,~\citet{zhang2025primacy} noted that LoRA and two of its initialization variants, PiSSA~\cite{pissa} and MiLoRA~\cite{milora}, exhibit performance shifts across two learning rates ($2e^{-4}$ and $2e^{-5}$).
\citet{he2026unified} also concluded that, with proper learning rate tuning, vanilla LoRA consistently matches or surpasses most of its variants under a single LLM scale with fixed rank and training duration.
Meanwhile, Lee et al.~\cite{lee2026beware} demonstrated that the reported advantages of PiSSA and MiLoRA could be driven by batch size bias. 
Our paper differs from these studies by expanding the investigation to a broader set of recent LoRA variants and conducting comprehensive multivariate hyperparameter tuning to identify the best-performing configuration for each method. Moreover, our work compares methods across varying ranks and training durations, and further leverages Hessian analysis to provide theoretical insight into the underlying factors driving the observed performance differences and trends.
\subsection{Vanilla LoRA and Its Variants}
\label{sec:peft}
\subsubsection{Low-Rank Adaptation}\label{sec:lora}
Given a pretrained neural network layer parameterized by $W_{\text{pre}} \in \mathbb{R}^{m \times n}$, LoRA introduced two trainable matrices: the down-projecting $A \in \mathbb{R}^{r \times n}$ and the up-projecting $B \in \mathbb{R}^{m \times r}$ ($r \ll \min(m, n)$). For layer input $x \in \mathbb{R}^{n}$, the output $h \in \mathbb{R}^{m}$ is computed as:
\begin{equation}\label{eq:lora_forward}
    h = W_{\text{pre}} x + \gamma_r B A x,
\end{equation}
where $\gamma_r=\frac{\alpha}{r}$ serves as a rank-dependent scaling factor with $\alpha$ being a tunable hyperparameter.
At initialization, the two trainable matrices $B$ and $A$ are set to $B_0 = 0$ and $A_0 \sim \mathcal{N}(0, \sigma^2)$ (i.e., Kaiming initialization~\cite{he2015delving}),
ensuring that fine-tuning starts exactly from the pretrained checkpoint.

\subsubsection{Representative LoRA Variants}\label{sec:lora_variants}
In this paper, we consider nine representative LoRA variants spanning diverse optimization mechanisms, which we organize into three categories: 
(1)~\textbf{\emph{Initialization Variants}} (OLoRA~\cite{buyukakyuz2024olora}, PiSSA~\cite{pissa}, MiLoRA~\cite{milora}, Init[AB]~\cite{initab}, LoRA-GA~\cite{loraga}), 
(2)~\textbf{\emph{Architecture Modifications}} (DoRA~\cite{dora}, GraLoRA~\cite{gralora}, RandLoRA~\cite{albert2025randlora}), 
and (3)~\textbf{\emph{Optimization Adjustments}} (LoFT~\cite{tastan2025loft}).
We describe their key design principles below and defer the detailed design rationales and  formulas
to Appendix Sec.~\ref{sec:variants-detailed-formulas}.

\paragraph{Initialization Variants.}\label{sec:init_variants}
This category comprises methods that explore improved initialization strategies for LoRA~\cite{pissa,buyukakyuz2024olora,loraga,milora,initab,yang2024corda,paischer2024parameter,lora-one}. Methods along this line can be further distinguished by whether their initialization requires 
task
data, yielding~\emph{data-free} and~\emph{data-informed} subcategories, both of which are considered in this work.
Specifically, within the data-free subcategory, 
OLoRA~\cite{buyukakyuz2024olora} applies QR decomposition to $W_{\text{pre}}$ to initialize $B$ and $A$ using 
the
first-$r$ columns of $Q$ and
the 
first-$r$ rows of $R$, respectively.
PiSSA~\cite{pissa} and MiLoRA~\cite{milora}, on the other hand, leverage the singular value decomposition (SVD) of $W_{\text{pre}}$ to inform the initialization of LoRA adapters, with PiSSA selecting the top-$r$ principal components and MiLoRA adopting the minor
ones.
Several works have also theoretically analyzed the initialization strategies of LoRA~\cite{hayou2024impact, xu2025understanding, initab}.
In particular, Init[AB]~\cite{initab} showed that randomly initializing both LoRA matrices using Kaiming initialization can 
be more advantageous
by balancing stability, training efficiency, and hyperparameter robustness. 
Turning to the~\emph{data-informed} subcategory, 
LoRA-GA~\cite{loraga} uses one-step full-gradient information to initialize LoRA adapters. 
Let $G=-\nabla_{W_{\text{pre}}}\mathcal{L}\in\mathbb{R}^{m\times n}$ denote the sampled full gradient with respect to $W_{\text{pre}}$.  LoRA-GA computes the SVD of $G$ and initializes LoRA adapters in a disjoint manner, using the top-$r$ right singular vectors for $A$ and the $(r+1)$-th through $2r$-th left singular vectors for $B$.  

Note that since $B_0A_0\neq0$ for all the initialization variants discussed above, the base weight is replaced by a~\emph{residual matrix} so that fine-tuning starts from the pretrained weights. 
Specifically, the~\emph{residual matrix} is defined as $W_{\mathrm{res}} = W_{\text{pre}} - B_0A_0$, 
and the modified forward pass becomes:
\begin{equation*}
    h = W_{\mathrm{res}}x + \gamma_r BAx.
\end{equation*}

\paragraph{Architectural Modifications.}\label{sec:arch_modifications}
Besides investigating initialization strategies, a large body of literature has 
also
focused on architecture-level improvements, 
e.g.,~\cite{gralora,albert2025randlora,vera,liu2023parameter,zhong2026peanut} available in the PEFT library~\cite{peft}. 
By modifying vanilla LoRA's forward design (i.e., Eq.~\ref{eq:lora_forward}), 
these methods improve fine-tuning effectiveness either by sustaining performance with greater parameter efficiency~\cite{vera, balazy2024lora, li2024vb, gao2024parameter, yang2024loretta}
or by achieving higher accuracy under a similar trainable-parameter budget~\cite{dora, gralora, albert2025randlora, huang2025hira, jiang2024mora}.
We focus on methods in the latter category, as large differences in trainable-parameter counts relative to LoRA make 
direct head-to-head comparisons non-trivial; e.g., VeRA~\cite{vera} substantially reduces the number of trainable parameters from $(m+n)r$ to $m+r$ per layer.
In particular, we select DoRA~\cite{dora}, RandLoRA~\cite{albert2025randlora}, and GraLoRA~\cite{gralora}, as they require no more than one additional architectural hyperparameter.
This contrasts with other modification strategies such as BoFT~\cite{liu2023parameter} and PEANuT~\cite{zhong2026peanut}, which involve multiple architectural choices, namely $(m,b)$ and depth/activation function, respectively, and could therefore rapidly expand the hyperparameter search space.
Due to space constraints, we defer the details of their forward designs to Appendix Sec.~\ref{sec:variants-arch-detailed-formulas}.


\paragraph{Optimization Adjustments.}\label{sec:optimization_adjustments}
More recent studies have started to improve LoRA by directly adjusting its optimization dynamics. 
For example, LoRA+~\cite{hayou2024lora+} assigns different learning rates to $A$ and $B$, while~\emph{scaled AdamW}~\cite{zhang2024riemannian} introduces an $r \times r$ preconditioner into each gradient step. 
LoRA-Pro~\cite{wang2024lorapro} further adjusts the gradients of LoRA so that the induced update better approximates the full fine-tuning gradient.
More recently, LoFT~\cite{tastan2025loft} aligns the optimizer's internal dynamics with full fine-tuning by projecting Adam~\cite{kingma2014adam}'s first- and second-moment estimates into the same low-rank subspace, narrowing the performance gap between LoRA and full fine-tuning.

\section{Learning Rate Matters, Really}\label{sec:exp}

\subsection{Motivation}\label{sec:motivation}
For the trainable LoRA parameters across layers, collectively denoted as $\theta$, the update rule of Stochastic Gradient Descent (SGD) at step $t$ is:
\begin{equation*}
    \boldsymbol{\theta}_{t+1} = \boldsymbol{\theta}_t - \eta \mathbf{g}(\boldsymbol{\theta}_t),
\end{equation*}
where $\eta$ is the learning rate and $\mathbf{g}(\boldsymbol{\theta}_t) \triangleq \nabla \mathcal{L}(\boldsymbol{\theta}_t)$ is the gradient of the loss function $\mathcal{L}$.
While setting $\eta$ too large causes the optimization step to overshoot, leading to instability or divergence, a value that is too small is insufficient to escape suboptimal local minima or affect the convergence rate. 
To analyze this formally, consider the local geometry characterized by the Hessian $\mathbf{H}(\boldsymbol{\theta}_t) \triangleq \nabla^2 \mathcal{L}(\boldsymbol{\theta}_t)$.
According to classical learning theories~\cite{lecun1992automatic}, the optimal learning rate $\eta^*$ for efficient learning is intrinsically tied to the curvature of the loss landscape at $\boldsymbol{\theta}$, typically scaling inversely with the Hessian's maximum eigenvalue:
\begin{equation}\label{eq:lr_and_hessian}
    \eta^* \propto \frac{1}{\lambda_{\max}(\mathbf{H}(\boldsymbol{\theta}))}.
\end{equation}
Notably, LoRA initialization variants establish specific training starting points $\boldsymbol{\theta}_0$, resulting in distinct $\mathbf{g}(\boldsymbol{\theta}_0)$, $\mathbf{H}(\boldsymbol{\theta}_0)$, and subsequent training trajectories compared to vanilla LoRA. 
Similarly, while LoRA variants based on architectural modifications or optimization adjustments could share the same $\mathbf{g}(\boldsymbol{\theta}_0)$ and $\mathbf{H}(\boldsymbol{\theta}_0)$ as vanilla LoRA, their subsequent gradient and Hessian evolution throughout training may naturally deviate from LoRA due to the unique forward designs or update rules.
Therefore, different methods theoretically require their respective calibrations of $\eta$ to ensure efficient convergence, motivating our decision to perform learning rate tuning for a fair and reliable head-to-head comparison across methods.

\subsection{Experimental Setup}\label{sec:exp_setup}
Since model choices, training configurations, and dataset partitioning vary across papers, we establish a unified experimental framework that accommodates all methods fairly. We describe the key components below, with additional implementation details deferred to Appendix~\ref{sec:implementation_details}.

\paragraph{Pretrained Models.}
We consider four decoder-only 
models 
spanning diverse scales: Qwen3-0.6B~\cite{qwen3}, Gemma-3-1B~\cite{gemma3}, Llama-2-7B, and 
Llama-2-13B~\cite{llama2}. This selection includes both recently released ones (Qwen3, Gemma-3) and an older but widely used model family (Llama-2) in prior art on this subject, enabling us to validate results across LLMs with diverse pretrained capabilities.

\paragraph{Fine-tuning Tasks.} We train models on four canonical tasks: commonsense reasoning, mathematical reasoning, code generation, and instruction following. The dataset setup follows prior LoRA studies.
Specifically, for commonsense reasoning, we leverage the 15k training examples compiled by~\citet{hu2023llm}, which comprise eight general question-answering subtasks.
For mathematical reasoning, we use 100k subsampled training examples from MetaMathQA~\cite{metamath} and evaluate models on GSM8K~\cite{gsm8k} and MATH~\cite{MATH}. 
For code generation, we use 104k subsampled training examples from CodeFeedback~\cite{codefeedback} and evaluate models on HumanEval~\cite{humaneval} and MBPP~\cite{mbpp}. 
For instruction following, we train models on 52k Alpaca~\cite{alpaca} examples and evaluate them using the IFEval framework~\cite{zhou2023instruction}.
Unless otherwise specified, we report mean accuracy 
over the testing datasets. 



\begin{table}[t]
\centering
\renewcommand{\arraystretch}{1.4} 
\setlength{\tabcolsep}{0.5pt}
\setlength{\aboverulesep}{0pt}
\setlength{\belowrulesep}{0pt}
\adjustbox{max width=\textwidth}{%
\begin{tabular}{l|c|*{12}{c}}
\toprule
\multirow{2}{*}{\centering\arraybackslash \textbf{Methods}} & 
\multirow{2}{*}{\centering\arraybackslash \makecell{\textbf{Batch}\\\textbf{Size}}} & 
\multicolumn{12}{c}{\raisebox{0.4ex}{\textbf{Learning Rate}}} \\
\cline{3-14}
& & 
\textbf{1.1e-5} & \textbf{2e-5} & \textbf{3.6e-5} & \textbf{6.3e-5} & 
\textbf{1.1e-4} & \textbf{2e-4} & \textbf{3.6e-4} & \textbf{6.3e-4} & 
\textbf{1.1e-3} & \textbf{2e-3} & \textbf{3.6e-3} & \textbf{6.3e-3} \\
\midrule

\multirow{3}{*}{\textbf{LoRA}}
& \textbf{16} 
& 9.78$_{\pm 0.36}$ 
& 11.16$_{\pm 0.28}$
& 13.58$_{\pm 0.18}$ 
& 15.48$_{\pm 0.15}$ 
& 18.43$_{\pm 0.14}$ 
& \cg \textbf{20.00$_{\pm 0.26}$ }
& \cg 19.93$_{\pm 0.65}$ 
& 17.99$_{\pm 0.55}$ 
& 11.71$_{\pm 0.49}$ 
& 1.52$_{\pm 0.19}$ 
& 1.27$_{\pm 0.59}$ 
& 1.07$_{\pm 0.27}$ \\
& \textbf{64} 
& 6.88$_{\pm 0.04}$ 
& 9.12$_{\pm 0.39}$ 
& 10.79$_{\pm 0.37}$
& 13.23$_{\pm 0.25}$
& 15.65$_{\pm 0.57}$
& 17.54$_{\pm 0.29}$ 
& \cg 19.73$_{\pm 0.16}$ 
& \cg \textbf{20.46$_{\pm 0.79}$}
& \cg 19.83$_{\pm 0.91}$ 
& 13.33$_{\pm 0.81}$ 
& 1.48$_{\pm 0.48}$ 
& 0.00$_{\pm 0.00}$ \\
& \textbf{128} 
& 5.70$_{\pm 0.34}$ 
& 6.95$_{\pm 0.23}$ 
& 9.41$_{\pm 0.44}$ 
& 11.43$_{\pm 0.40}$ 
& 13.68$_{\pm 0.77}$ 
& 15.92$_{\pm 0.45}$ 
& \cg 18.58$_{\pm 0.44}$ 
& \cg 19.60$_{\pm 0.09}$ 
& \cg \textbf{20.32$_{\pm 0.28}$ }
& 16.95$_{\pm 2.70}$ 
& 0.09$_{\pm 0.16}$ 
& 0.00$_{\pm 0.00}$ \\
\cline{1-14}

\multirow{3}{*}{\textbf{DoRA}}
& \textbf{16} 
& 9.89$_{\pm 0.24}$
& 11.16$_{\pm 0.51}$
& 13.84$_{\pm 0.41}$
& 15.61$_{\pm 0.11}$
& 18.21$_{\pm 0.45}$ 
& \cg 20.11$_{\pm 0.26}$ 
& \cg \textbf{20.96$_{\pm 0.57}$}
& 18.34$_{\pm 0.20}$ 
& 11.90$_{\pm 0.29}$ 
& 4.89$_{\pm 0.99}$ 
& 0.93$_{\pm 0.12}$ 
& 1.16$_{\pm 0.15}$ \\
& \textbf{64} 
& 6.72$_{\pm 0.09}$ 
& 9.19$_{\pm 0.19}$ 
& 10.53$_{\pm 0.20}$ 
& 13.45$_{\pm 0.31}$ 
& 15.72$_{\pm 0.32}$ 
& 17.66$_{\pm 0.20}$ 
& \cg 19.96$_{\pm 0.05}$ 
& \cg \textbf{20.82$_{\pm 0.32}$ }
& \cg 19.87$_{\pm 0.91}$ 
& 13.53$_{\pm 1.64}$ 
& 1.52$_{\pm 0.45}$ 
& 0.34$_{\pm 0.23}$ \\
& \textbf{128} 
& 5.55$_{\pm 0.11}$ 
& 7.21$_{\pm 0.18}$ 
& 9.72$_{\pm 0.17}$ 
& 11.58$_{\pm 0.25}$ 
& 13.98$_{\pm 0.33}$ 
& 16.19$_{\pm 0.46}$ 
& 18.25$_{\pm 0.23}$ 
& \cg 19.67$_{\pm 0.71}$ 
& \cg \textbf{20.33$_{\pm 0.64}$ }
& 12.86$_{\pm 10.03}$ 
& 0.13$_{\pm 0.23}$ 
& 0.02$_{\pm 0.03}$ \\
\cline{1-14}

\multirow{3}{*}{\textbf{Init[AB]}}
& \textbf{16} 
& 9.73$_{\pm 0.35}$ 
& 12.10$_{\pm 0.14}$ 
& 14.41$_{\pm 0.49}$ 
& 16.73$_{\pm 0.37}$ 
& 18.38$_{\pm 0.53}$ 
& \cg 20.39$_{\pm 0.38}$ 
& \cg \textbf{20.55$_{\pm 0.40}$ }
& 18.34$_{\pm 0.48}$ 
& 11.94$_{\pm 0.31}$ 
& 1.48$_{\pm 0.24}$ 
& 1.16$_{\pm 0.31}$ 
& 1.45$_{\pm 0.17}$ \\
& \textbf{64} 
& 6.51$_{\pm 0.22}$ 
& 9.15$_{\pm 0.12}$ 
& 11.28$_{\pm 0.20}$ 
& 13.20$_{\pm 0.24}$ 
& 15.88$_{\pm 0.39}$ 
& 17.89$_{\pm 0.30}$ 
& \cg 20.08$_{\pm 0.26}$ 
& \cg \textbf{20.98$_{\pm 0.33}$ }
& \cg 19.31$_{\pm 0.75}$ 
& 13.97$_{\pm 0.03}$ 
& 2.74$_{\pm 3.83}$ 
& 0.07$_{\pm 0.12}$ \\
& \textbf{128} 
& 6.06$_{\pm 0.35}$ 
& 7.05$_{\pm 0.33}$ 
& 9.53$_{\pm 0.22}$ 
& 11.81$_{\pm 0.08}$ 
& 13.98$_{\pm 0.79}$ 
& 16.46$_{\pm 0.39}$ 
& 18.36$_{\pm 0.21}$ 
& \cg 20.37$_{\pm 0.39}$ 
& \cg \textbf{20.66$_{\pm 0.39}$ }
& 17.85$_{\pm 0.84}$ 
& 4.40$_{\pm 7.46}$ 
& 0.00$_{\pm 0.00}$ \\
\cline{1-14}

\multirow{3}{*}{\textbf{MiLoRA}}
& \textbf{16} 
& 12.44$_{\pm 0.07}$ 
& 13.77$_{\pm 0.25}$ 
& 16.28$_{\pm 0.24}$ 
& 18.45$_{\pm 0.47}$ 
& \cg 20.04$_{\pm 0.19}$ 
& \cg \textbf{20.63$_{\pm 0.67}$}
& \cg 19.40$_{\pm 0.80}$ 
& 15.72$_{\pm 0.49}$ 
& 10.22$_{\pm 0.42}$ 
& 2.03$_{\pm 0.95}$ 
& 1.35$_{\pm 0.43}$ 
& 1.56$_{\pm 0.65}$ \\
& \textbf{64} 
& 8.82$_{\pm 0.40}$ 
& 11.25$_{\pm 0.20}$ 
& 13.16$_{\pm 0.11}$ 
& 15.54$_{\pm 0.29}$ 
& 17.43$_{\pm 0.24}$ 
& \cg 19.56$_{\pm 0.33}$ 
& \cg \textbf{20.03$_{\pm 0.59}$}
& \cg 19.60$_{\pm 0.78}$ 
& 17.93$_{\pm 0.90}$ 
& 13.65$_{\pm 0.07}$
& 4.97$_{\pm 0.40}$ 
& 0.00$_{\pm 0.00}$ \\
& \textbf{128} 
& 7.32$_{\pm 0.33}$ 
& 9.57$_{\pm 0.24}$ 
& 11.76$_{\pm 0.33}$ 
& 13.54$_{\pm 0.12}$ 
& 16.02$_{\pm 0.16}$ 
& 18.39$_{\pm 0.26}$ 
& \cg 19.70$_{\pm 0.34}$ 
& \cg \textbf{19.99$_{\pm 0.66}$} 
& \cg 19.53$_{\pm 0.47}$ 
& 16.83$_{\pm 0.73}$ 
& 7.45$_{\pm 1.00}$ 
& 0.57$_{\pm 0.81}$ \\
\cline{1-14}

\multirow{3}{*}{\textbf{PiSSA}}
& \textbf{16} 
& 14.30$_{\pm 0.18}$ 
& 16.10$_{\pm 0.27}$ 
& 18.31$_{\pm 0.12}$ 
& \cg 19.90$_{\pm 0.21}$ 
& \cg \textbf{20.61$_{\pm 0.28}$} 
& \cg 19.09$_{\pm 0.20}$ 
& 16.10$_{\pm 0.64}$ 
& 13.25$_{\pm 0.55}$ 
& 8.41$_{\pm 0.13}$ 
& 4.67$_{\pm 0.29}$ 
& 2.50$_{\pm 1.27}$ 
& 0.96$_{\pm 0.15}$ \\
& \textbf{64} 
& 11.11$_{\pm 0.05}$ 
& 13.67$_{\pm 0.17}$ 
& 15.56$_{\pm 0.33}$ 
& 18.11$_{\pm 0.23}$ 
& \cg 19.52$_{\pm 0.48}$ 
& \cg \textbf{20.68$_{\pm 0.77}$}
& \cg 20.59$_{\pm 0.32}$ 
& \cg 19.11$_{\pm 0.86}$ 
& 15.53$_{\pm 0.37}$ 
& 9.57$_{\pm 0.72}$ 
& 5.78$_{\pm 0.37}$ 
& 0.33$_{\pm 0.46}$ \\
& \textbf{128} 
& 9.42$_{\pm 0.38}$ 
& 11.80$_{\pm 0.28}$ 
& 14.40$_{\pm 0.11}$ 
& 16.23$_{\pm 0.38}$ 
& \cg 18.60$_{\pm 0.21}$ 
& \cg 19.61$_{\pm 0.44}$ 
& \cg \textbf{20.65$_{\pm 0.44}$} 
& \cg 19.21$_{\pm 1.15}$ 
& 16.91$_{\pm 0.19}$ 
& 13.87$_{\pm 0.97}$ 
& 6.28$_{\pm 0.49}$ 
& 1.19$_{\pm 0.36}$ \\
\bottomrule
\end{tabular}
}
\vspace{0.4em}
\caption{
Performance of Gemma-3-1B on mathematical reasoning task across varying batch sizes and learning rates ($r=128$). Results are reported as mean $\pm$ standard deviations over three independent runs. Best results are highlighted in \textbf{bold}, and configurations achieving $\ge18.5\%$ accuracy (i.e., $\approx 90\%$ of the maximum) are shaded in green (\colorbox{oprange}{\phantom{12}}). 
While all methods achieve comparable peak accuracies, the optimal learning rates vary depending on both the fine-tuning method and batch size.}
\label{tab:gemma-main}
\vspace{-1.8em}
\end{table}

\paragraph{Hyperparameter Settings.} 
We consider batch sizes ($B$) in \{16, 32, 64, 128, 256, 512\} and ranks ($r$) in \{4, 8, 16, 32, 64, 128, 256\}.
The learning rates ($\eta$) are tuned uniformly on a logarithmic scale from $10^{-6}$ to $10^{-3}$, with four values per order of magnitude: {1.1247$\times 10^*$, 2.0000$\times 10^*$, 3.5566$\times 10^*$, 6.3246$\times 10^*$}, yielding up to 16 grid points for the learning rate alone. 
To maintain the computational feasibility of this study, batch size and rank are tuned only for selected model-task combinations; conversely, learning rates are tuned across all combinations, with ranges defined to ensure inclusion of optimal performance. See Appendix Table~\ref{tab:model_task_config} for a summary of models, tasks, and their corresponding hyperparameter search ranges.
Other configurations, such as epoch, adapter placement, and learning rate scheduler, remain fixed across all experiments and are listed in Appendix Table~\ref{tab:other_hyperparameters}. Specifically for the scaling factor $\gamma_r$, we follow \citet{pissa} by setting $\alpha$ equal to $r$ in all our experiments. This results in $\gamma_r=1$ for all $r$, effectively factoring out the need to tune this hyperparameter 
(refer to Appendix Sec.~\ref{sec:on_scaling_factor} for further discussion). 

\subsection{Results and Observations}\label{sec:main_results}
We begin by discussing results under fixed rank
$r=128$ in Sec.~\ref{sec:similar_perf_level}, 
and 
then
in Sec.~\ref{sec:performnace_w_rank}, we analyze how the methods perform across different ranks.

\subsubsection{Similar Performance Levels}\label{sec:similar_perf_level}
\begin{figure}[t]
    \centering
    \begin{subfigure}[b]{0.497\textwidth}
        \centering
        \includegraphics[width=\linewidth]{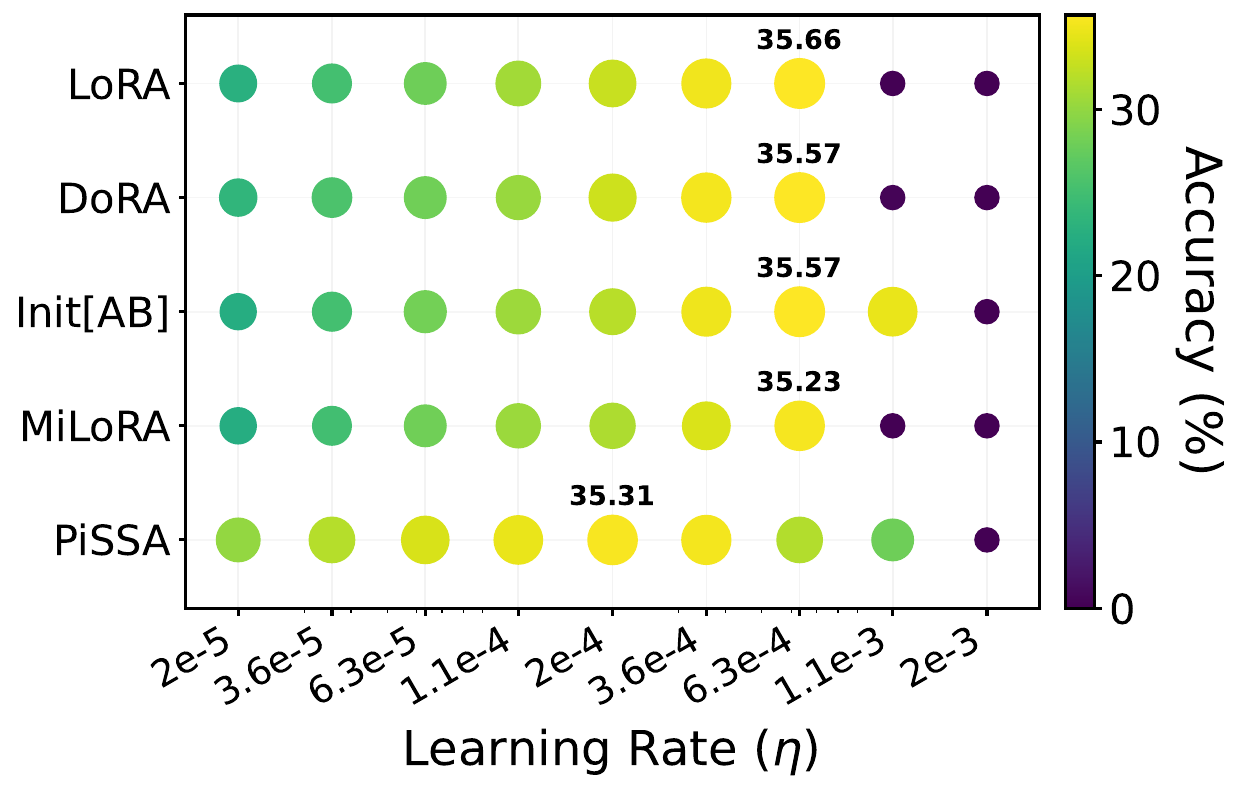} 
        \caption{
        Mathematical Reasoning}
        \label{fig:llama-main-math}
    \end{subfigure}
    \hfill
    \begin{subfigure}[b]{0.497\textwidth}
        \centering
        \includegraphics[width=\linewidth]{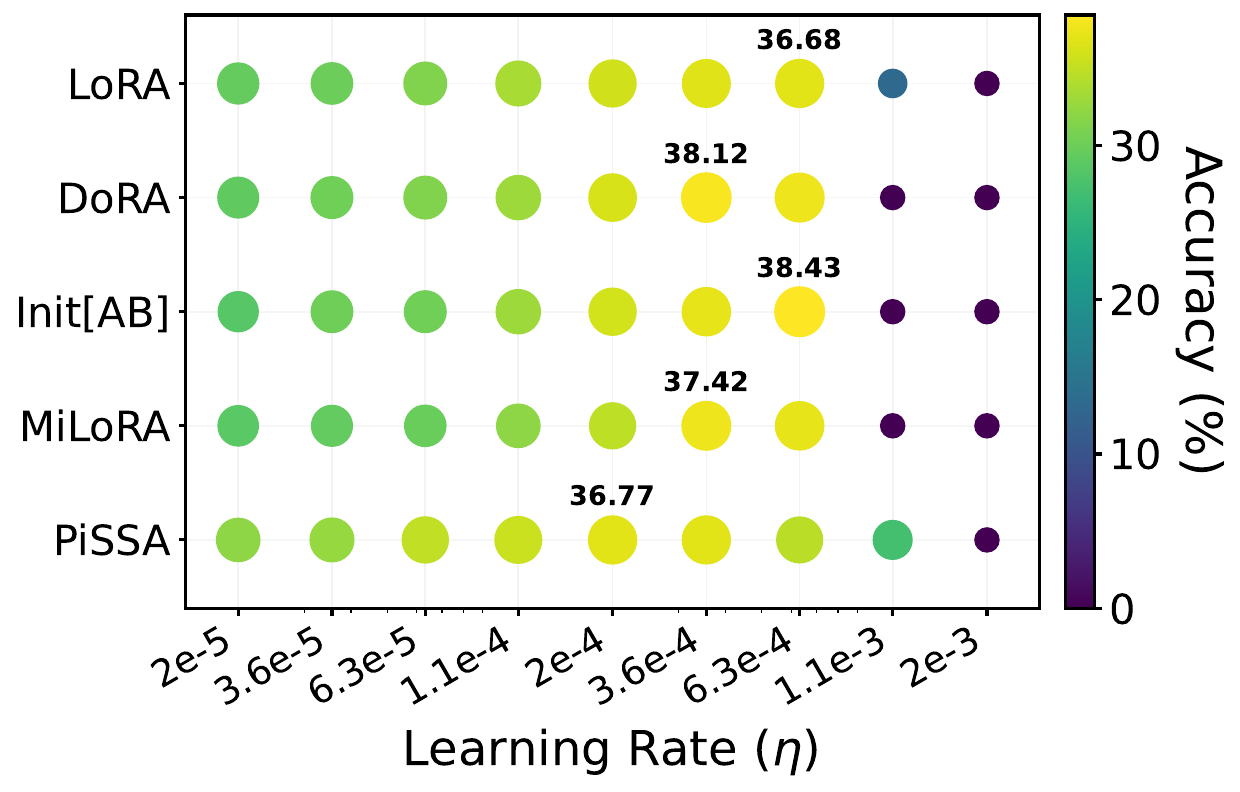} 
        \caption{
        Code Generation}
        \label{fig:llama-main-code}
    \end{subfigure}
    \caption{
    Performance of Llama-2-7B on mathematical reasoning and code generation tasks across varying learning rates ($r=128$, $B=128$). 
    Notably, PiSSA peaks at lower learning rates but remains effective at larger learning rates on both tasks (e.g., 1.1$\times 10^{-3}$), where other methods diverge. Results scaling up to Llama-2-13B are provided in Appendix Figure~\ref{fig:llama-13b-math}.
    }
    \label{fig:main-llama}
\end{figure}

To further examine whether the performance parity among LoRA methods observed on Qwen3-0.6B (Figure~\ref{fig:main-evidence} and Appendix Figure~\ref{fig:main-evidence-gemma-cs-r4}) generalizes to diverse model--task combinations, Table~\ref{tab:gemma-main}, Figure~\ref{fig:main-llama}, and Appendix Figure~\ref{fig:llama-13b-math} present the results for Gemma-3-1B, Llama-2-7B, and Llama-2-13B, respectively.
To maintain the computational feasibility of this study, 
we select four popular and recently published LoRA variants from the nine methods considered previously, namely PiSSA, MiLoRA, Init[AB], and DoRA.
Through comprehensive hyperparameter searches, we consistently observe across different model scales and tasks that these methods peak at performance levels similar to vanilla LoRA.
Specifically, the performance gaps across all methods remain small: 
0.52\% for Gemma-3-1B on math, 0.43\% and 1.75\% for Llama-2-7B on math and code, respectively, 
and 1.81\% for Llama-2-13B on math.
It is also important to note that while peak performance is similar, the optimal learning rates vary. In particular, PiSSA requires a lower learning rate compared to vanilla LoRA across all model-task combinations, while other methods fall within a similar range to LoRA, typically within the same order of magnitude.

Beyond the optimal learning rates, a closer inspection of the full learning rate spectrum reveals intriguing method-specific behaviors. 
For instance, we observe that PiSSA tends to remain effective at larger learning rates where other methods diverge: in Figure~\ref{fig:main-llama} at $\eta=$1.1$\times10^{-3}$, PiSSA maintains accuracies of 27.83\% and 26.90\% on math and code tasks, respectively, while other methods collapse to near-zero performance on at least one of the tasks.

Note that in Table~\ref{tab:gemma-main}, the joint optimization of learning rate and batch size indicates that tuning the learning rate is significantly more critical than tuning the batch size for obtaining the best performance in both LoRA and its variants, consistent with early findings for neural networks~\cite{bengio2012practical}.
For example, with
PiSSA,
fixing the learning rate at 2$\times10^{-5}$ and tuning only the batch size yields a suboptimal maximum accuracy of 
16.1\%. 
In contrast, by fixing the batch size to any value in $\{16, 64, 128\}$ and tuning the learning rate, the model achieves substantially higher performance
around 20.6\%. 
Moreover, we observe that the optimal learning rate scales proportionally with batch size, aligning with the ``scaling rule'' established in SGD literature~\cite{goyal2017accurate, hoffer2017train}. 
This offers several~\emph{practical heuristics} for LoRA hyperparameter tuning: 
\textbf{\emph{I.}} Under limited computational resources, one may consider prioritizing learning rate tuning while fixing the batch size to a small or medium value.\footnote{When the batch size is set too large, the best achievable performance of LoRA methods under comprehensive learning rate tuning starts to decay, as we show in Appendix~\ref{sec:addition-batch-size}.}
\textbf{\emph{II.}} If additional resources are available and practitioners wish to explore different batch sizes, 
even though further performance gains are likely to be marginal once the learning rate has been properly tuned, 
we remind practitioners to keep in mind the scaling relationship between batch size and learning rate, 
which can help guide initial learning rate selection when configuring different batch sizes.
More numerical results, example model responses, and practical learning heuristics (\textbf{\emph{III}}--\textbf{\emph{V}}) are provided in Appendix~\ref{sec:detail_table}, Sec.~\ref{sec:example_model_response}, and Sec.~\ref{sec:practical-heuristic}, respectively.





\begin{figure}[t] 
    \centering
    \hspace{-1em}
    \begin{subfigure}{0.475\linewidth}
        \centering
        \includegraphics[width=\linewidth]{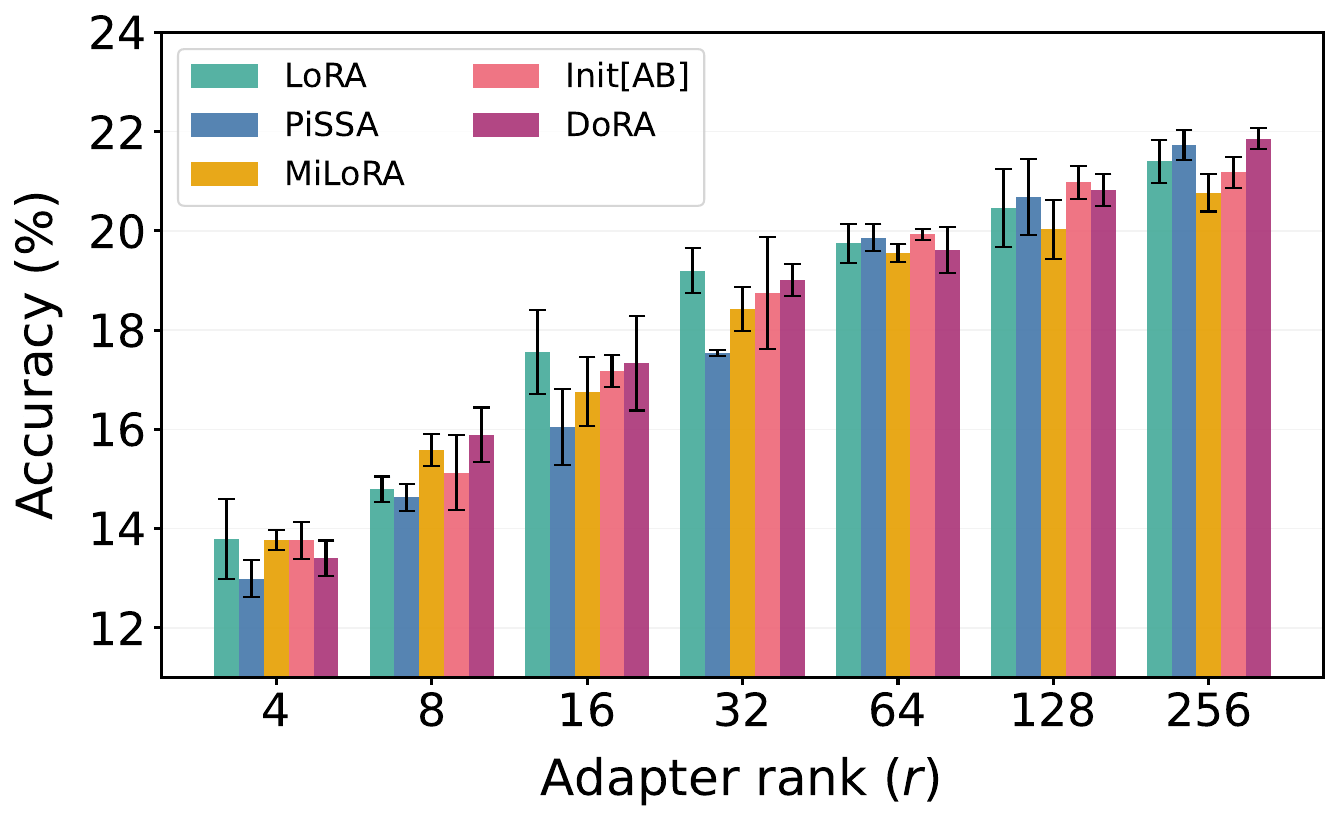}
        \caption{Mathematical Reasoning}\label{fig:performance_across_rank_math}
    \end{subfigure}
    \begin{subfigure}{0.475\linewidth}
        \centering
        \includegraphics[width=\linewidth]{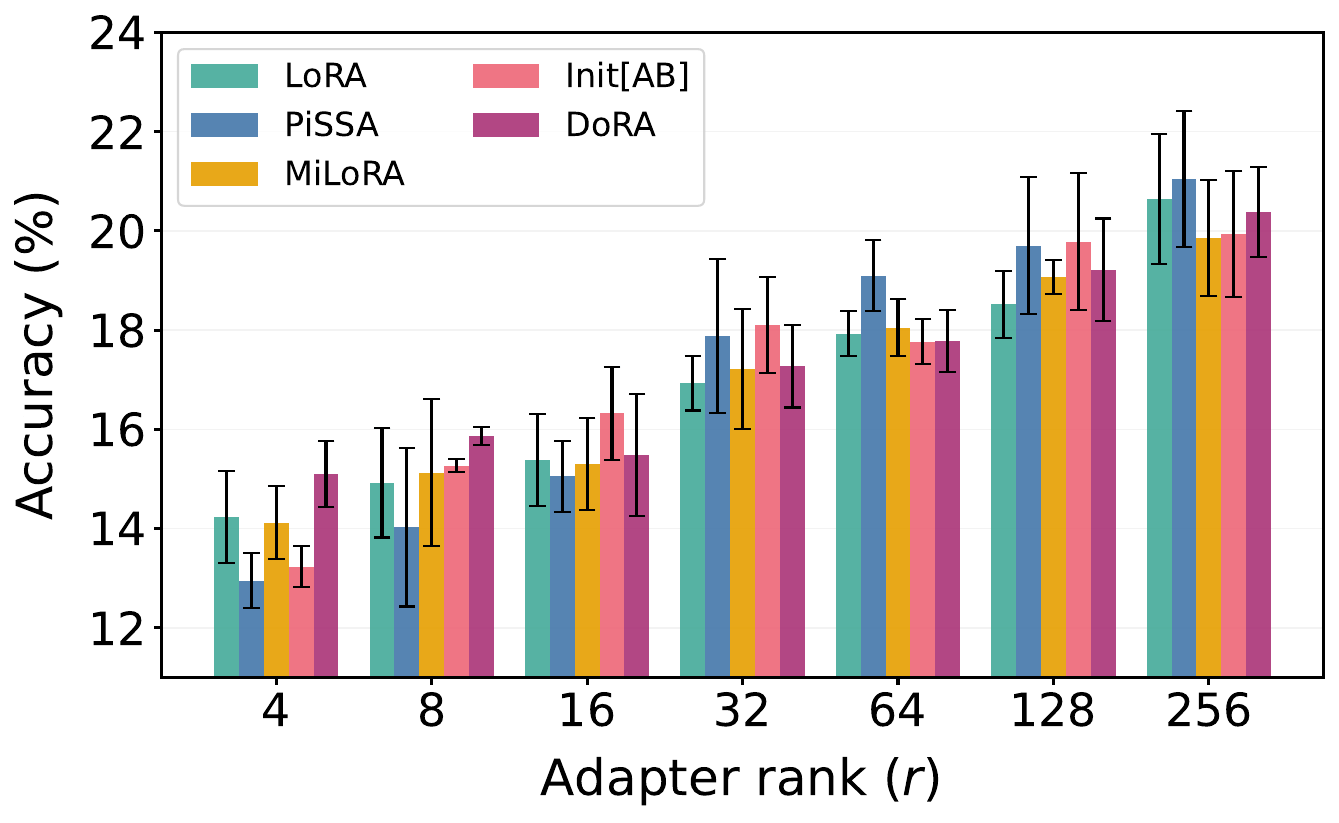}
        \caption{Code Generation}\label{fig:performance_across_rank_code}
    \end{subfigure}
    \caption{Best achievable performance of LoRA and its advanced variants across adapter ranks on Gemma-3-1B ($B=64$). With properly tuned learning rates, all methods exhibit similar performance improvement trends as the rank increases, though subtle rank-dependent behaviors emerge. 
    Results are reported with means and standard deviations over three independent runs. 
    Additional rank-tuning results for Llama-2-7B are deferred to Appendix Sec.~\ref{sec:ablation:adapter_rank}.
    }
    \label{fig:perf_across_ranks}
\end{figure}

\subsubsection{Performance Comparison Across Ranks}\label{sec:performnace_w_rank}


Next, we extend our analysis by varying adapter ranks for Gemma-3-1B, as shown in Figure~\ref{fig:perf_across_ranks}. The results indicate that the previously observed performance parity persists across a wide range of rank settings, with the maximum performance differences among methods being only 1.67\% (Math, $r=32$) and 2.15\% (Code, $r=4$).
Interestingly, however, we observe that the relative performance of variants compared to LoRA fluctuates across different ranks within these margins.

In particular, PiSSA initially underperforms vanilla LoRA before gradually overtaking it as the rank increases. 
Taking the math task as an example (Figure~\ref{fig:performance_across_rank_math}), PiSSA exhibits performance deficits of up to 1.67\% at low ranks ($r\leq32$), but narrows the gap to within 0.11\% at $r=64$ and shifts to a slight gain of 0.22\% and 0.33\% at $r=128$ and 256, respectively. 
In contrast, MiLoRA shows an opposite trend, where it tends to outperform vanilla LoRA at lower ranks but fails to sustain this advantage as the rank increases.
Figure~\ref{fig:performance_across_rank_code} indicates that these rank-dependent dynamics extend to the coding task.
For
Init[AB], we observe that it tends to outperform LoRA at medium ranks, e.g., achieving maximal gains of 0.52\% on math and 1.26\% on code at $r=128$. Yet, the success does not translate to either lower or higher rank scenarios, where Init[AB] typically performs similarly to vanilla LoRA.
As for DoRA, we observe performance gains against LoRA specifically in low-rank regimes, peaking at 1.1\% on math and 0.95\% on code at $r=8$.
Similar performance comparisons across ranks were also conducted on Llama-2-7B, with results deferred to Appendix Sec.~\ref{sec:ablation:adapter_rank}. 
Beyond adapter ranks, we also validate our findings under different numbers of training samples and training epochs, with results deferred to Appendix Sec.~\ref{sec:vary_train_duration}.

\section{Understanding the Optimal Learning Rate via Hessian Analysis}\label{sec:hessian}

\subsection{Sharpness-Learning Rate Relationship}
The Hessian of the loss function has been the subject of numerous studies. Geometrically, its top eigenvalue (denoted as $\lambda_{\max}$ for brevity) at a given point represents the maximal curvature of the loss landscape along any direction, commonly referred to as sharpness~\cite{dinh2017sharp, lyu2022understanding, luo2024explicit}. 
This metric is closely linked to the optimal learning rate, a connection that originates from the Gauss-Newton method for convex optimization and was further elucidated by~\citet{lecun1992automatic} in the context of neural networks. 
Specifically, it was shown that an efficient learning rate theoretically falls within $1/\lambda_{\max} \le \eta^* < 2/\lambda_{\max}$ under quadratic approximation, whereas rates exceeding $2/\lambda_{\max}$ lead to divergence.
More recently,~\citet{lewkowycz2020large} identified a ``catapult'' learning regime characterized by $2/\lambda_{\max} \le \eta^* \le 12/\lambda_{\max}$, in which modern architectures achieve optimal performance.
Further research has explored
the intricacies of the interplay between $\lambda_{\max}$ and $\eta^{*}$ with a consensus that these two quantities exhibit an inversely proportional relationship~\cite{pan2021eigencurve,cohen2021gradient,kalra2024warmup}.

\subsection{Sharpness Analysis in LoRA}\label{sec:hessian-analysis}
For our LoRA fine-tuning problem, we leverage the downstream MetaMathQA dataset to compute the Hessian matrix of the loss function 
and focus exclusively on the trainable LoRA parameters~\cite{yang2023bayesian, zhao2024second,yu2025prunedlora}.
Instead of concatenating LoRA parameters across all layers, we follow standard 
LLM practices
to estimate $\lambda_{\max}$ in a block-wise manner~\cite{zhang2024transformers, wang2025sharpness, ilin2025hessian} at the initialization point.
Formally, we calculate the layer-wise metric as $\lambda_{\max}^{l} = \lambda_{\max}(\mathbf{H}^l)$, where $\mathbf{H}^l$ represents the Hessian corresponding to parameters $\theta^{l}=\{B_0^{l}, A_0^{l}\}$, with $l$ indexing 
matrix types and Transformer layers. 
The Lanczos algorithm~\cite{lanczos1950iteration} and Hessian-vector products are 
used
to estimate the top eigenvalue without explicitly forming $\mathbf{H}$. Implementation details are provided in Appendix~\ref{sec:lanczos_algo_implement}. 
While LoRA architectural modifications and optimization adjustments may share LoRA's initialization, implying identical initial Hessians, their unique forward designs and update rules may lead to distinct Hessian evolution throughout training. 
We thus defer their investigation to future work.

\begin{wrapfigure}{r}{0.43\columnwidth}
    \vspace{-10pt}
    \centering
    \includegraphics[width=0.43\columnwidth]{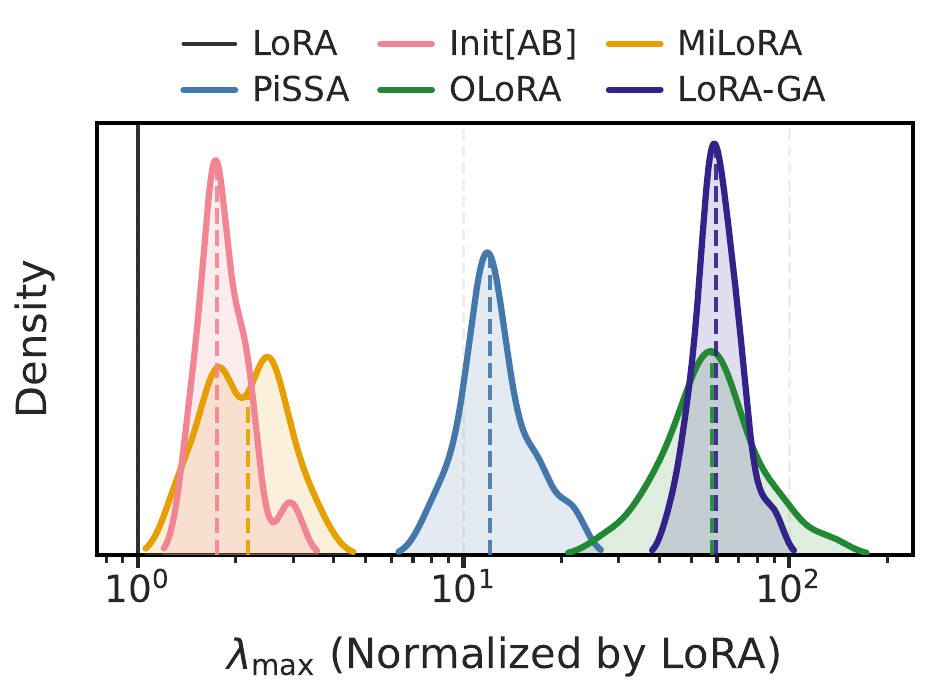}
    \caption{Distributions of the ratios of the top loss Hessian eigenvalues relative to LoRA for Query projection matrices across Transformer layers on Qwen3 ($r=128$). Dashed lines indicate the medians.
    } \label{fig:main_hessian}
    \vspace{-8pt}
\end{wrapfigure}

Specifically, let the Hessian for the Query projection matrix in the $i$-th layer be $\mathbf{H}^{Q,i}_{t}$,
where $t$ indexes different LoRA methods.
We further denote their corresponding maximum eigenvalues by $\lambda_{\max,t}^{Q,i}$. 
Then, we normalize the maximum eigenvalues from LoRA initialization variants by that from LoRA, and plot the distribution across layers in Figure~\ref{fig:main_hessian}, i.e. $\lambda_{\max,t}^{Q,i}/\lambda_{\max,\text{LoRA}}^{Q,i}$ for $t=$Init[AB], MiLoRA, PiSSA, OLoRA, LoRA-GA and $i=1,\ldots, L$.
The results reveal that all methods initialize trainable parameters in a higher curvature state than vanilla LoRA. Most notably, OLoRA and LoRA-GA exhibit up to $100\times$ higher curvature, explaining the reason behind their requirement for a much lower learning rate (18.2$\times$ lower) in Figure~\ref{fig:main-evidence}. Similar patterns apply to other methods. In particular, PiSSA exhibits $\approx 10\times$ higher curvature, which is consistent with its requirement for a $10\times$ lower learning rate. For Init[AB] and MiLoRA, however, the eigenvalue magnitudes are more similar to those of vanilla LoRA ($\approx 2\times$ higher), supporting their lower optimal learning rates by factors of $1.8\times$ and $3.2\times$ in Figure~\ref{fig:main-evidence}, respectively.
Detailed $\lambda_{\max}$ values and Hessian analyses on other models and matrix types are provided in Appendix Sec.~\ref{sec:additional_hessians_gemma_llama} and Sec.~\ref{sec:detailed_lambda}.

\section{Conclusion}\label{sec:conclusion}
Motivated by the increasing number of LoRA variants and the insufficient hyperparameter tuning in many 
studies,
in this work, we conducted a systematic re-evaluation of 
ten
LoRA PEFT methods under a unified evaluation protocol.
Based on the comprehensive hyperparameter experiments, we conclude that 
\textbf{improper learning rates give a false sense of LoRA advancements}.
In our studies, we also pointed out the scenarios where one might be in favor of a specific variant that, albeit likely to produce generally comparable performance, marginally outperforms other variants. It is worth noting that these improvements often lack universality, with vanilla LoRA frequently matching or even outperforming them.
By elucidating the disparate optimal learning rate ranges through Hessian analysis, we hope our study encourages future PEFT research to adopt a more comprehensive hyperparameter search protocol, ensuring reliable advancements in the field. 
We also hope that the five practical heuristics derived from our experimentation will help practitioners reduce the computational burden associated with LoRA tuning.
We acknowledge that this paper is subject to several limitations, primarily due to computational constraints, which we discuss in detail
below.





\section*{Limitations and Future Work}\label{sec:limitations}
In this paper,
we focused our investigation on decoder-only LLMs from 0.6B to 13B parameter scale. 
Hence, the scalability of our findings to larger foundation models remains to be verified. 
Additionally, the computational costs required for hyperparameter searches for each LoRA method on diverse model--task combinations and training durations precluded an exhaustive search over relatively minor hyperparameters. In particular, while key hyperparameters (learning rate, batch size, LoRA rank) were tuned, other secondary training setups, such as learning rate schedulers, warmup steps, and LoRA adapter placements, remained fixed.
It may be possible that fine-grained tuning of these configurations could yield further performance gains or distinct convergence behaviors.

We also highlight that our findings may not extend to untested model architectures (e.g., encoder-only LLMs~\cite{devlin2019bert}, Vision Transformers~\cite{dosovitskiy2020image}, and Vision-Language Models~\cite{alayrac2022flamingo}) or to all existing advanced LoRA variants.
For instance, several methods have originally reported higher peak performance than LoRA under comprehensive learning rate sweeps—such as LoRA-One~\cite{lora-one}, which initializes adapters via the SVD of the one-step full gradient, with $\approx2\%$ performance improvement on Llama (cf.~\citet{lora-one} Table 3).
Moreover, fine-tuned accuracy on standard benchmarks is not the sole criterion for evaluating PEFT algorithms; specific variants may offer distinct advantages in other dimensions, e.g., mitigating catastrophic forgetting of pretrained knowledge~\cite{biderman2024lora,milora,yang2024corda,zhang2025lora,xiong2025oplora, quercia2026least}.

While the landscape of LoRA variants continues to expand, our results suggest that vanilla LoRA already suffices as a competitive baseline, potentially indicating that weight-based low-rank adaptation strategies may be approaching saturation. 
Looking ahead, we posit that further investigation into alternative adaptation mechanisms may unlock new dimensions of efficiency. 
Examples of such mechanisms include hidden representation fine-tuning~\cite{wu2024reft,yin2024lofit} 
and approaches that adapt non-linear functions within layers~\cite{yin2025don}. We leave the exploration of these orthogonal paradigms as future work.

\section*{Acknowledgements}
This work was supported in part by National Science and Technology Council, Taiwan, R.O.C., under grant 113-2628-E-001-003-MY4, and by National Taiwan University and Academia Sinica Innovative Joint Program, under grant AS-NTU-114-06.
We thank the authors of prior LoRA studies for open-sourcing their work, and all contributors to the official PEFT package for their detailed documentation, continued maintenance, and prompt handling of issues.
Y.-A. Lee would like to thank the academic training and support received from the Data Science Degree Program at NTU and Academia Sinica, as well as the NTU Overseas Internship Program and IBM Research.

{\small
\bibliographystyle{unsrtnat}
\bibliography{reference}
}

\newpage
\appendix
\enableaddcontentsline
{
    \hypersetup{linkcolor=black}
    \setcounter{tocdepth}{2}
    \tableofcontents
}
\newpage

{\small
\section{Additional Experiments}\label{sec:additional_ablations}

In this section, we present fine-tuning results for various LoRA methods under additional model--task combinations (Sec.~\ref{sec:addition-model-task}), different LoRA ranks (Sec.~\ref{sec:ablation:adapter_rank}), and diverse training durations (Sec.~\ref{sec:vary_train_duration}). 
Moreover, we scale up the batch size in Sec.~\ref{sec:addition-batch-size} to examine performance under lower-stochasticity regimes and include instruction-following tasks in Sec.~\ref{sec:addition-inst-following} to cover diverse task types.

\subsection{Learning Rate Tuning on More Model--task Combinations}~\label{sec:addition-model-task}

Analogously to Figure~\ref{fig:main-evidence}, where we present all ten LoRA methods for Qwen3-0.6B on mathematical reasoning tasks, Figure~\ref{fig:main-evidence-gemma-cs-r4} presents the corresponding comparison for Gemma-3-1B on commonsense reasoning tasks. With proper learning rate tuning, all methods consistently peak at a similar performance level ($\approx37\%$). 
We note that, although the detailed ordering of the optimal learning rate ranges across methods may differ slightly from that shown in Figure~\ref{fig:main-evidence}, the overall trend remains broadly similar. For example, \{DoRA, LoFT, GraLoRA\} share slightly higher or comparable learning rate ranges relative to LoRA, whereas \{PiSSA, Init[AB], RandLoRA, OLoRA, MiLoRA\} tend to exhibit lower ranges. Notably, LoRA-GA consistently requires substantially lower learning rate ranges than LoRA, by more than 10$\times$ in both Figure~\ref{fig:main-evidence} and Figure~\ref{fig:main-evidence-gemma-cs-r4}.
This relatively stable relationship between each LoRA variant and vanilla LoRA across different model--task combinations suggests that practitioners who wish to use different LoRA variants may 
leverage
their known relative learning rate ranges with respect to LoRA as a practical prior to guide learning rate tuning, without necessarily conducting an exhaustive learning rate tuning as in our study or re-estimating Hessian using new downstream task samples as in Sec.~\ref{sec:hessian}.

\begin{figure}[h]
    \centering
    \includegraphics[width=0.985\linewidth]{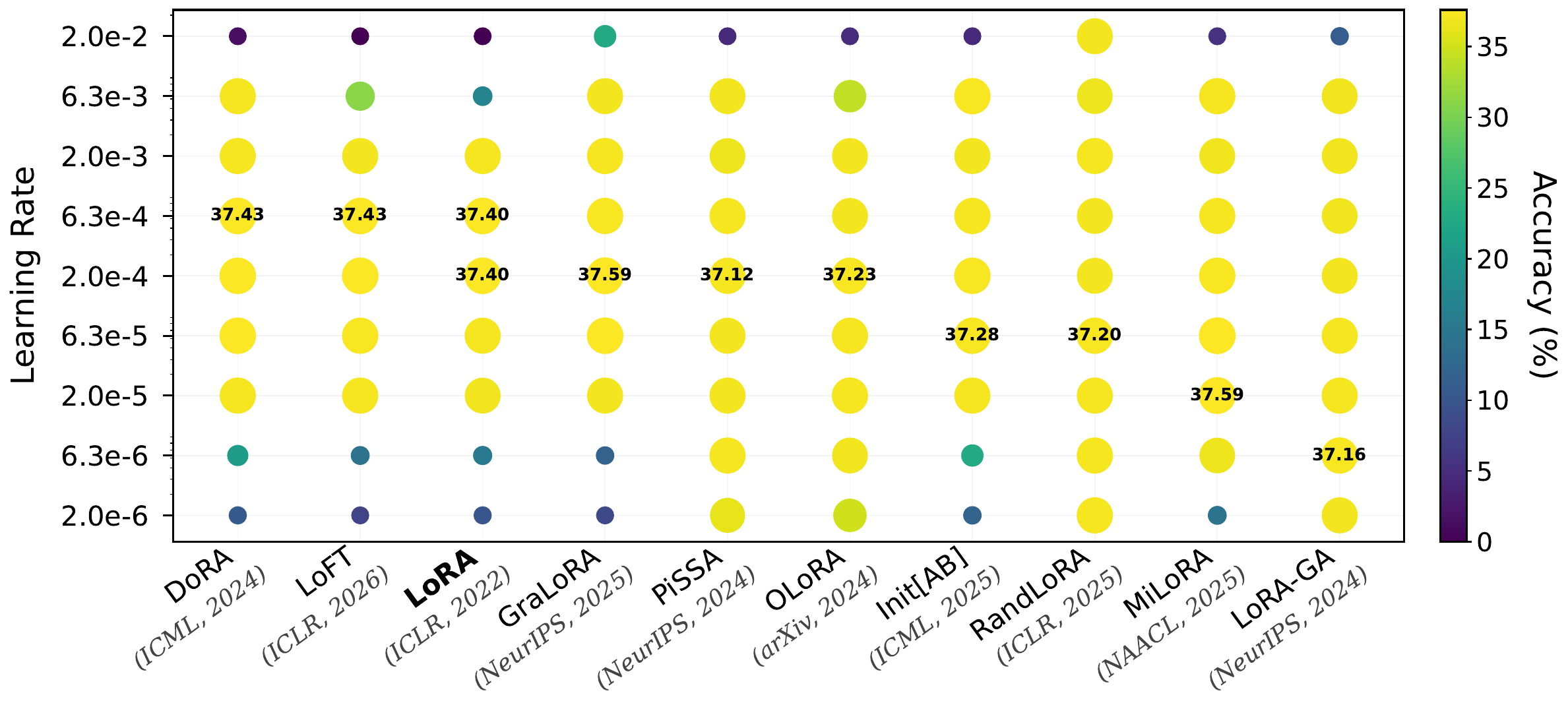}
    \vspace{-0.05em}
    \caption{Performance of Gemma-3-1B fine-tuned on commonsense reasoning tasks 
    across learning rates ($r=4, B=64$). Different methods reach a similar performance level once the learning rate is properly tuned. Each point is averaged over three independent training runs, and methods are sorted by their optimal learning rate ranges.} 
    \label{fig:main-evidence-gemma-cs-r4}
\end{figure}


Similarly to Figure~\ref{fig:main-llama}, where we present the performance of Llama-2-7B for LoRA, DoRA, Init[AB], MiLoRA, and PiSSA, Figure~\ref{fig:llama-13b-math} scales the analysis up to Llama-2-13B. The results validate the generalizability of our conclusions to larger model scales, where the largest gains over vanilla LoRA are 0.61\% for Init[AB] when $r=8$ and 0.53\% for DoRA when $r=128$.

\begin{figure}[h]
    \centering
    \begin{subfigure}[b]{0.48\linewidth}
        \centering
        \includegraphics[width=\linewidth]{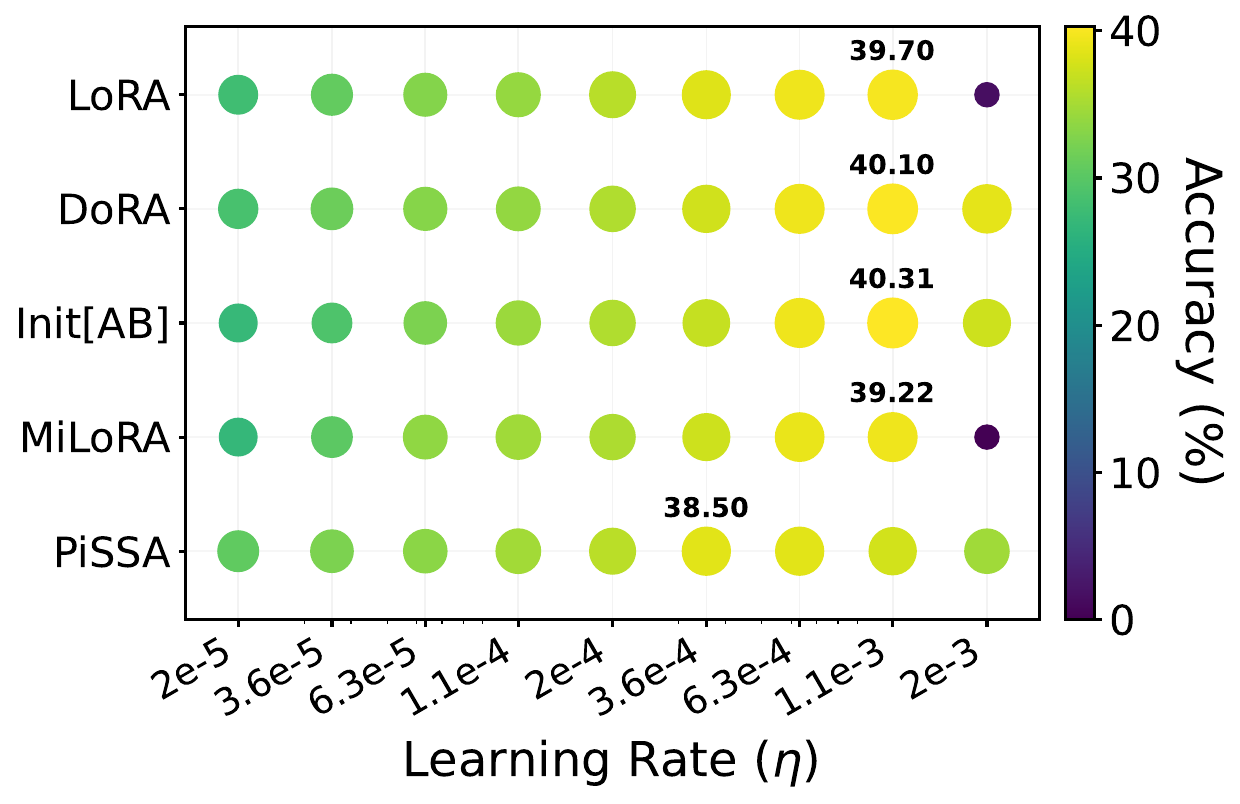}
        \caption{$r=8$}
    \end{subfigure}
    \hfill
    \begin{subfigure}[b]{0.48\linewidth}
        \centering
        \includegraphics[width=\linewidth]{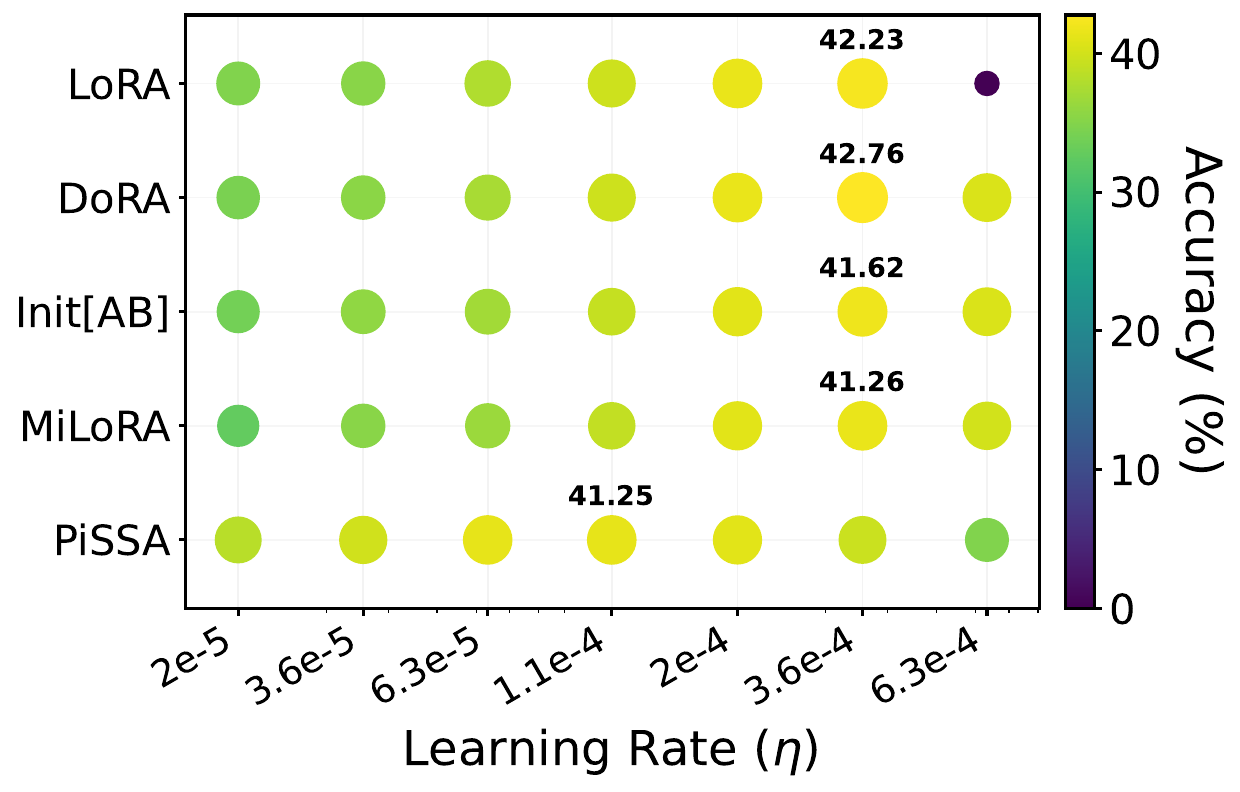}
        \caption{$r=128$}
    \end{subfigure}
    \caption{Performance of Llama-2-13B on mathematical reasoning tasks across varying learning rates and ranks $r \in\{8, 128\}$ ($B = 64$).}
    \label{fig:llama-13b-math}
\end{figure}

\FloatBarrier

\subsection{Varying Adapter Ranks on Llama}\label{sec:ablation:adapter_rank}
In Sec.~\ref{sec:performnace_w_rank}, we analyzed the behavior of LoRA PEFT methods as the adapter rank varies on Gemma across math and code tasks. Here, we present a corresponding analysis on Llama in Figure~\ref{fig:perf_across_ranks_llama}, reporting the best performance achieved under joint optimization of learning rate and batch size ($B\in\{16,128\}$). Similar trends can be observed, where all methods exhibit comparable performance improvement trends as the rank increases. Although under specific tasks or rank settings, one might favor a particular variant that marginally outperforms others, it is worth noting again that these improvements often lack universality, with vanilla LoRA frequently matching or even outperforming them.

\begin{figure*}[h] 
    \centering
    \begin{subfigure}{0.48\linewidth}
        \centering
        \includegraphics[width=\linewidth]{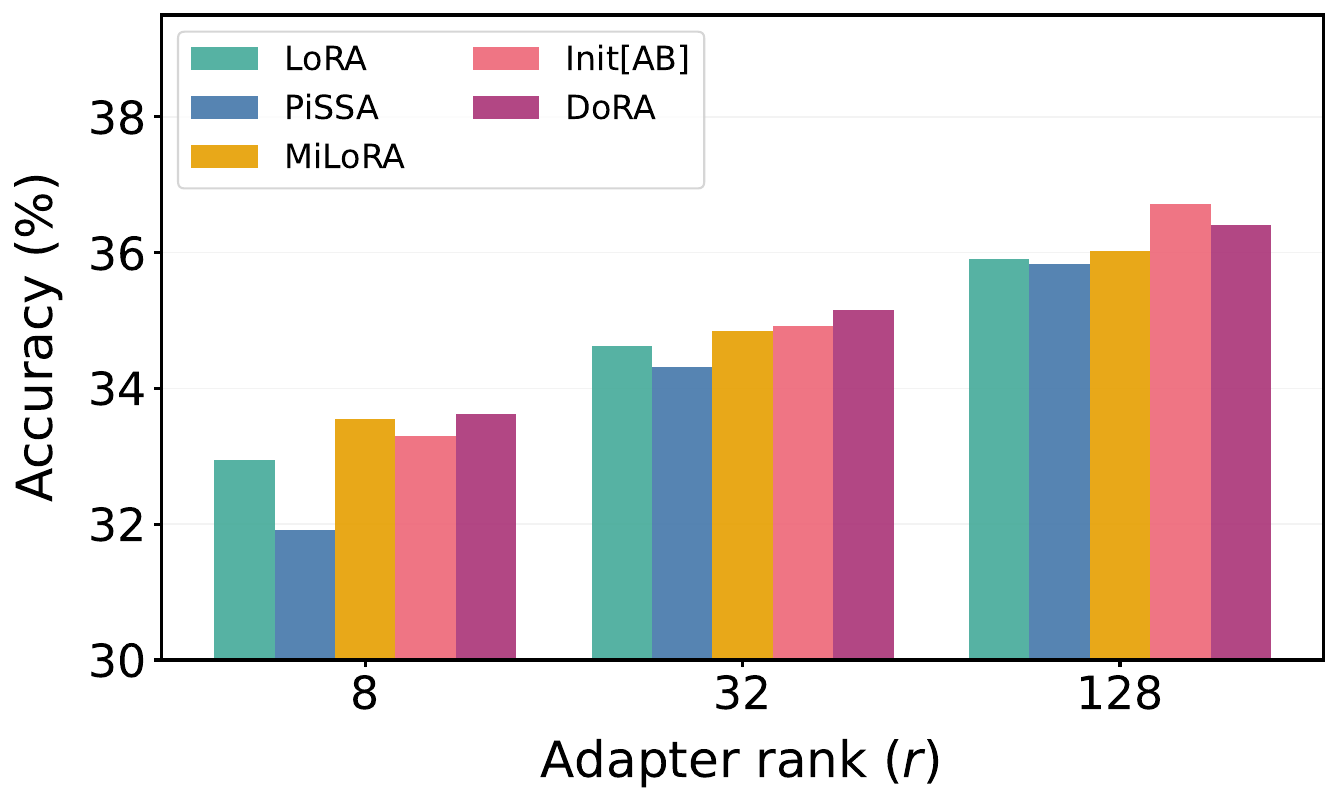}
        \caption{Mathematical Reasoning}\label{fig:perf_across_ranks_llama_math}
    \end{subfigure}
    \hfill 
    \begin{subfigure}{0.48\linewidth}
        \centering
        \includegraphics[width=\linewidth]{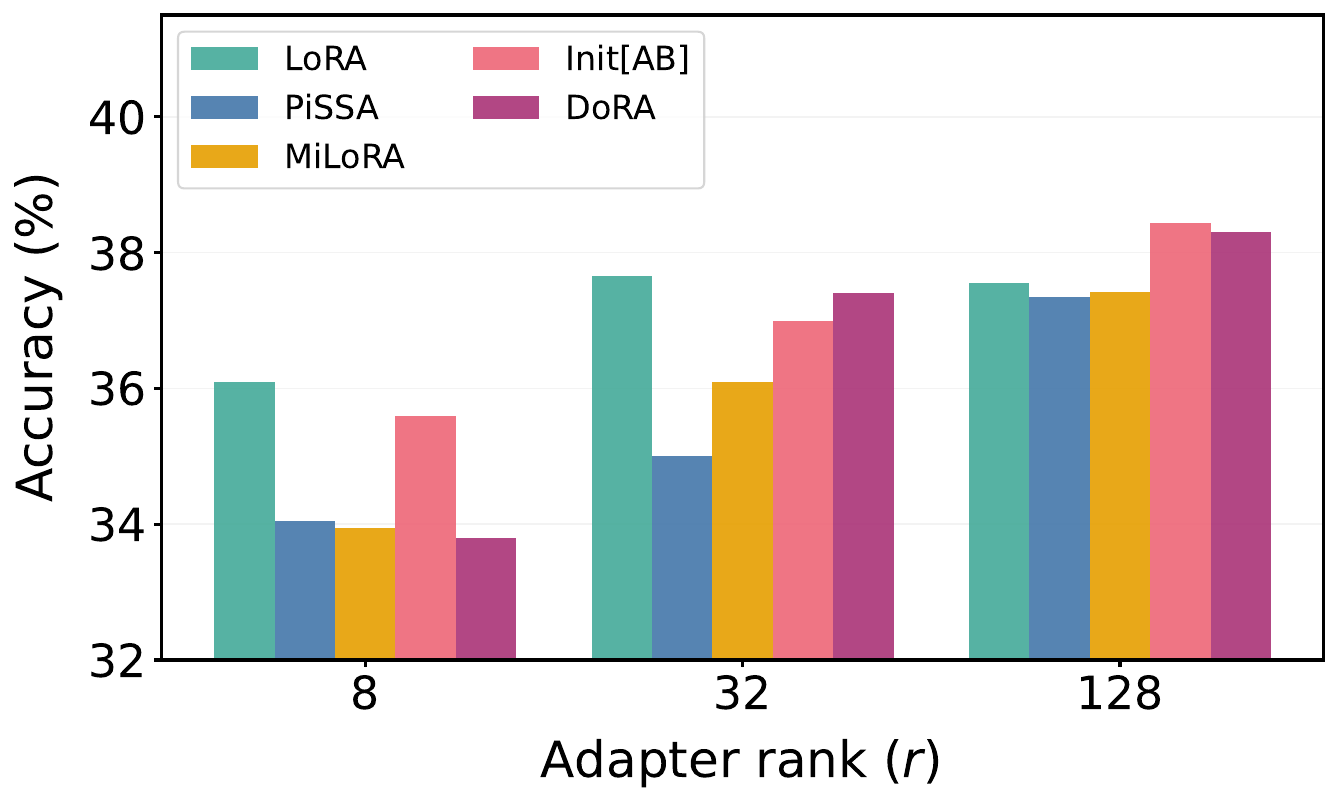}
        \caption{Code Generation}\label{fig:perf_across_ranks_llama_code}
    \end{subfigure}
    \caption{Best achievable performance of LoRA and its advanced variants across adapter ranks on Llama-2-7B ($B \in \{16, 128\}$).
    }
    \label{fig:perf_across_ranks_llama}
\end{figure*}

\FloatBarrier
\clearpage

\subsection{Varying Training Duration}\label{sec:vary_train_duration}
We examine whether different methods consistently peak at comparable performance levels under varying training durations. 
Specifically, we vary the training duration by (1) scaling the number of MetaMathQA training samples from 5k up to the full 395k (Figure~\ref{fig:number_training_samples}), and (2) varying the number of training epochs in \{1, 2, 3\} with a fixed 100k MetaMathQA training samples (Figure~\ref{fig:number_training_epochs}). 
With accuracies averaged over three independent runs, the results show that vanilla LoRA and its variants exhibit similar improvement trends as the training duration increases, with their performance generally falling within one another’s standard deviation range.

\begin{figure}[h]
    \centering    
    \includegraphics[width=0.50\linewidth]{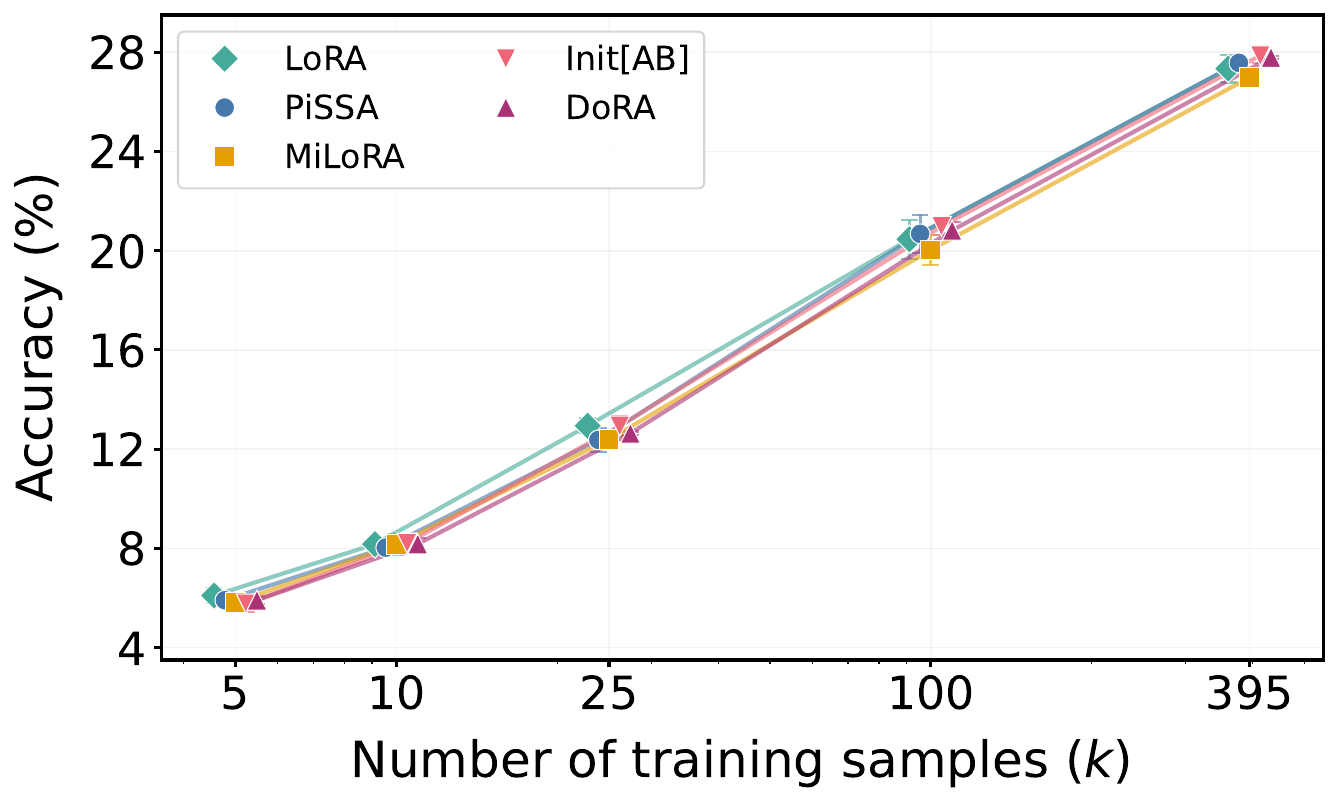}
    \caption{Best achievable performance of LoRA and its variants across different training sample sizes on mathematical reasoning with Gemma-3-1B ($r=128, B=64$). Once the learning rate is properly tuned, all methods exhibit nearly identical improvement trends as the number of training samples increases. Results are reported with mean and standard deviation over three runs.
    }
    \label{fig:number_training_samples}
\end{figure}

\begin{figure}[h]
    \centering    
    \includegraphics[width=0.50\linewidth]{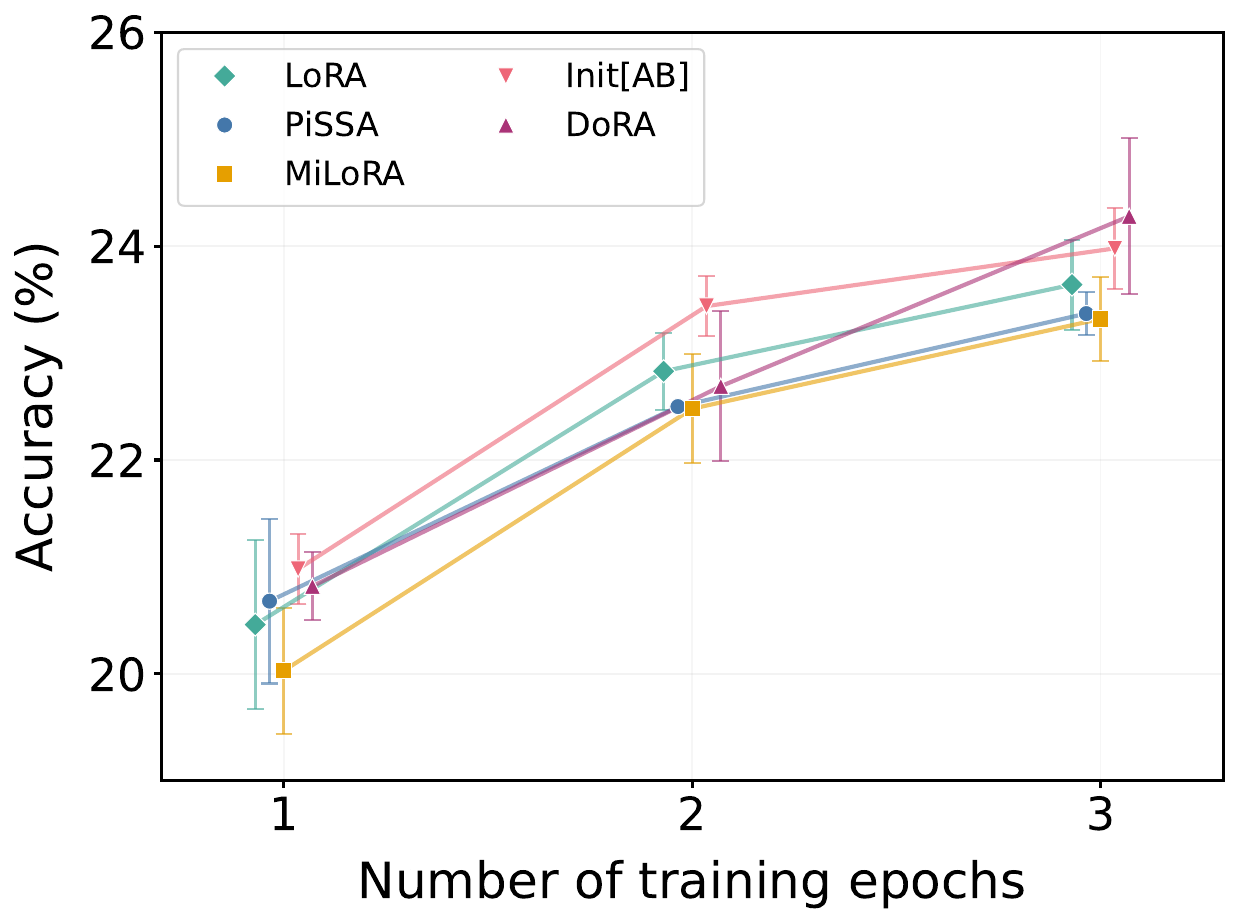}
    \caption{Best achievable performance of LoRA and its variants across different numbers of training epochs on mathematical reasoning with Gemma-3-1B ($r=128, B=64$). Results are reported with means and standard deviations over three independent runs. Once the learning rate is properly tuned, all methods exhibit nearly identical improvement trends as the number of training epochs increases, with their performance largely falling within each other's standard deviation ranges.}
    \label{fig:number_training_epochs}
\end{figure}

\FloatBarrier
\clearpage

\subsection{Scaling Batch Size to 512 for LoRA and PiSSA}\label{sec:addition-batch-size}

In Table~\ref{tab:gemma-main}, we jointly optimize the learning rate and batch size for Gemma-3-1B on mathematical reasoning tasks.  
However, the batch sizes considered there remain within standard stochastic training regimes ($B\in\{16,64,128\}$). 
To further examine the behavior of LoRA methods under larger-batch, lower-stochasticity regimes, we scale $B$ up to 512 in Figure~\ref{fig:gemma-batch-size}. 
The results reveal an intriguing phenomenon: even after learning-rate tuning, the best achievable performance of both LoRA and PiSSA begins to decay when the batch size reaches $B\geq 256$. 
This suggests that, while learning-rate tuning should be prioritized, the batch size should still be kept within a relatively small-to-medium range, as stated in \emph{practical heuristic \textbf{I}}.

\begin{figure}[h]
    \centering    
    \includegraphics[width=0.50\linewidth]{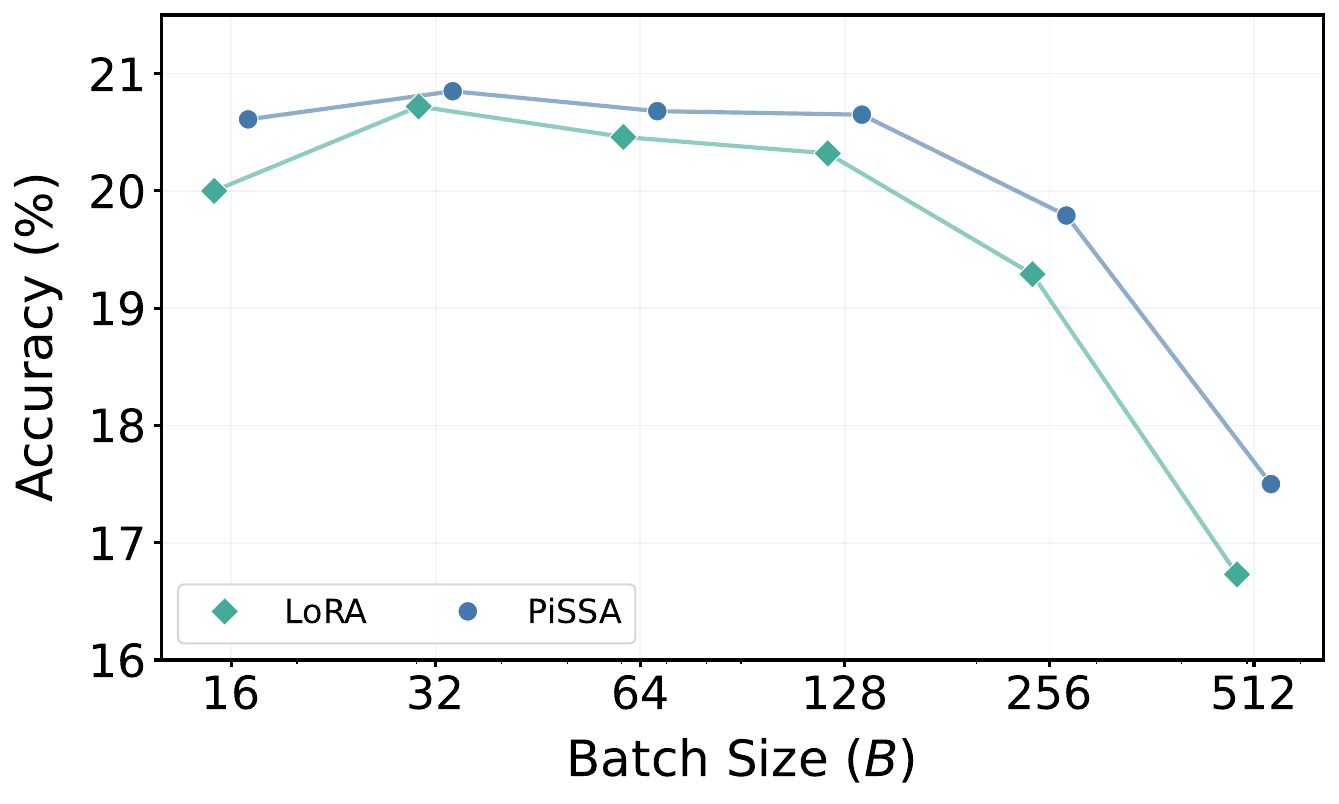}
    \caption{Best achievable performance of LoRA and PiSSA across different batch sizes on mathematical reasoning tasks with Gemma-3-1B ($r=128$). While at $B\le 128$, both LoRA and PiSSA can reach $\approx$ 20\% accuracy with proper learning rate tuning, the performance upper bound gradually decays when $B$ increases to 512.}
    \label{fig:gemma-batch-size}
\end{figure}

\FloatBarrier

\subsection{Fine-tuning on Instruction Following Tasks for LoRA and DoRA}\label{sec:addition-inst-following}

Beyond the commonsense reasoning, mathematical reasoning, and code generation tasks considered previously, we extend our evaluation to instruction-following tasks. Specifically, we train Qwen3-0.6B on Alpaca~\cite{alpaca}, while evaluating models' prompt-level strict accuracy under the  IFEval~\cite{zhou2023instruction} framework. 

\begin{table}[h!]
\centering
\setlength{\aboverulesep}{0pt}
\setlength{\belowrulesep}{0pt}
\renewcommand{\arraystretch}{1.4}
\resizebox{0.72\textwidth}{!}{%
\begin{tabular}{l|ccccccc}
\toprule
 & \textbf{2e-06} & \textbf{6.3e-06} & \textbf{2e-05} & \textbf{6.3e-05} & \textbf{2e-04} & \textbf{6.3e-04} & \textbf{2e-03} \\
\midrule
\textbf{LoRA} & 26.84 & 26.63 & 26.93 & \textbf{28.20} & 27.32 & 26.97 & 15.30 \\
\textbf{DoRA} & 26.99 & 26.41 & 26.41 & 26.44 & \textbf{27.42} & 26.54 & 16.86 \\
\bottomrule
\end{tabular}
}
\vspace{0.4em}
\caption{The prompt-level strict accuracy on IFEval with Qwen3-0.6B ($r=128$, $B=64$). Both DoRA and LoRA achieve a similar performance level under proper learning rate tuning.}
\label{tab:ifeval-qwen3}
\end{table}

\clearpage
\section{Comprehensive Study of Hyperparameters in Prior Work}\label{sec:prior_studies_hyperparameter}

\subsection{Survey Criteria}
To generate the statistics presented in Figure~\ref{fig:frequency_statistics}, we curated a dataset comprising 
64 papers, consisting of 
54 studies published at major AI conferences or journals, 6 high-impact arXiv preprints (exceeding 40 citations), and 4 recent preprints released within the last six months. For each paper, we examined whether the authors reported performance metrics under learning rate or batch size tuning, and whether comparisons across different ranks were provided. The selection criteria for inclusion were as follows:

\begin{enumerate}
    \item The primary objective of the proposed method is to enhance fine-tuning effectiveness (i.e., aiming for higher accuracy with equivalent trainable parameter counts or sustained performance with greater parameter efficiency).
    \item Vanilla LoRA is explicitly employed as a baseline for performance comparison.
\end{enumerate}

In assessing hyperparameter tuning, \textbf{our analysis focuses exclusively on decoder-only LLMs}, excluding encoder-only~\cite{devlin2019bert}, encoder-decoder~\cite{raffel2020exploring} architectures, Vision Transformers~\cite{dosovitskiy2020image}, and Vision-Language Models~\cite{alayrac2022flamingo}, as these lie outside the scope of this work. Consequently, papers lacking experiments on decoder-only LLMs are excluded from our statistics
(e.g.,~\cite{zhang2023adalora, liu2023parameter, xu2025understanding, zhang2025lora2, edalati2025krona, yuan2024bridging}). 
Moreover, we exclude papers focusing on objectives other than standard PEFT efficiency, such as parameter-efficient pretraining~\cite{lialin2023relora}, continual learning~\cite{wang2023orthogonal, zhao2024galore, luo2026keeplora}, and quantization~\cite{dettmers2023qlora, xu2023qa}.

Given that some studies may tune hyperparameters only for their proposed methods while leaving baselines untuned (e.g., by adopting settings from prior work without modification), \textbf{we rigorously verified whether the vanilla LoRA baseline underwent tuning}. Specifically, we consider the learning rate and batch size to be ``tuned'' only if they are evaluated across at least three distinct values. Consequently, studies such as MiLoRA (which compared two sets of hyperparameter setups) or LoRA-GA~\cite{wang2024lora} (which tested learning rate in \{1e-5, 5e-5\}) do not qualify as tuning under our criteria. For rank, we require comparisons across at least two distinct values. Crucially, if a study varies the rank for its proposed method but benchmarks against a fixed-rank vanilla LoRA, we do not classify the baseline rank as tuned  
(e.g.,~\cite{yuan2024bridging, balazy2024lora, li2024vb, he2026gora}).

During the data curation process, we observed that verifying specific hyperparameter tuning details can be non-trivial in some cases. The difficulty arises from discrepancies between paper versions (e.g., ablation studies on hyperparameters added to the Appendix post-publication), incomplete descriptions of experimental setups, or underspecified hyperparameter settings. Additionally, we frequently observed papers performing hyperparameter tuning only on smaller encoder-only LLMs (e.g., RoBERTa~\cite{liu2019roberta}). In strict adherence to our inclusion criteria, we do not categorize these instances as tuned. 
Moreover, we observed that many prior LoRA studies that did involve hyperparameter tuning reported only the final optimal performance, leaving unclear whether the adopted search ranges covered the optimal configurations for each method.
While these ambiguities complicate binary categorization, we have made every effort to ensure accuracy. We emphasize that the statistics is curated solely to present the current state of hyperparameter tuning practices in LoRA PEFT research, and we welcome future contributions or corrections to further refine this collection.

\subsection{Comprehensive List of Papers}
A comprehensive list of the reviewed papers is detailed in Table~\ref{tab:prior_studies}. Specifically, ``arXiv Date'' marks the release date of the first version on arXiv (denoted as ``--'' if unavailable). ``Pub. Date'' refers to the formal publication date of the venue, where ``--'' indicates the paper has not yet been formally published. Note that the table is sorted primarily by Pub. Date, followed by arXiv Date.
\newcommand{\arraystretchval}{0.5}

\begingroup
\setlength{\LTleft}{0pt}
\setlength{\LTright}{0pt}
\setlength{\LTcapwidth}{\linewidth}

\captionsetup[longtable]{
    justification=raggedright,
    singlelinecheck=false
}

{\fontsize{8pt}{24pt}\selectfont
\setlength{\tabcolsep}{1.2pt}

\begin{longtable}{@{\extracolsep{\fill}}l c c c l l c c c@{}}

\caption{Publication dates, venues, and experimental configurations in prior LoRA PEFT studies. The table summarizes decoder-only LLMs and tasks, noting whether the vanilla LoRA baseline involved learning rate (\textbf{LR}) or batch size (\textbf{BS}) tuning and offered comparisons across different \textbf{Ranks}. A positive entry (\ccmark) indicates the configuration was provided for at least one model-task combination; \xxmark~denotes otherwise.
}
\label{tab:prior_studies} \\

\toprule
\noalign{\vspace{-6pt}}
\textbf{Method} & \textbf{arXiv Date} & \textbf{Pub. Date} & \textbf{Venue} &
\textbf{Decoder-only LLM} & \textbf{Fine-tuned Task} &
\textbf{LR} & \textbf{BS} & \textbf{Rank} \\[-3pt]
\midrule
\endfirsthead

\toprule
\noalign{\vspace{-6pt}}
\textbf{Method} & \textbf{arXiv Date} & \textbf{Pub. Date} & \textbf{Venue} &
\textbf{Decoder-only LLM} & \textbf{Fine-tuned Task} &
\textbf{LR} & \textbf{BS} & \textbf{Rank} \\[-3pt]
\midrule
\endhead
    \bottomrule
    \endlastfoot

    \textbf{DyLoRA}~\cite{valipour2023dylora} & 
    2022-10 & 
    2023-05 & 
    EACL & 
    GPT-2 Medium & 
    NLG & 
    \xxmark & \xxmark & \ccmark \\[0.3em]

    \textbf{GLoRA}~\cite{chavan2023one} & 
    2023-06 & 
    -- & 
    arXiv & 
    {\renewcommand{\arraystretch}{\arraystretchval}%
    \begin{tabular}[c]{@{}l@{}}Llama-1-7B\\[-0.3em]Llama-2-7B\end{tabular}} & 
    NLG & 
    \xxmark & \xxmark & \xxmark \\[0.3em]

    \textbf{LoRA-FA}~\cite{zhang2023lora} & 
    2023-08 & 
    -- & 
    arXiv & 
    {\renewcommand{\arraystretch}{\arraystretchval}%
    \begin{tabular}[c]{@{}l@{}}Llama-1-7B\\[-0.3em]Llama-2-7B\end{tabular}} & 
    Commonsense & 
    \xxmark & \xxmark & \xxmark \\[0.3em]

    {\renewcommand{\arraystretch}{\arraystretchval}%
    \begin{tabular}[c]{@{}l@{}}
    \textbf{Laplace-LoRA}\\[-0.3em]
    \cite{yang2023bayesian}
    \end{tabular}} &
    2023-08 & 
    2024-05 & 
    ICLR & 
    {\renewcommand{\arraystretch}{\arraystretchval}%
    \begin{tabular}[c]{@{}l@{}}Llama-1-7B\\[-0.3em]Llama-2-7B\end{tabular}} & 
    Commonsense & 
    \xxmark & \xxmark & \xxmark \\[0.3em]

    \textbf{VeRA}~\cite{kopiczko2023vera} & 
    2023-10 & 
    2024-05 & 
    ICLR & 
    {\renewcommand{\arraystretch}{\arraystretchval}%
    \begin{tabular}[c]{@{}l@{}}GPT-2 Medium/Large\\[-0.3em]Llama-1-7B/13B\\[-0.3em]Llama-2-7B/13B\end{tabular}} & 
    {\renewcommand{\arraystretch}{\arraystretchval}%
    \begin{tabular}[c]{@{}l@{}}NLG\\[-0.3em]Instruction Following\end{tabular}} & 
    \ccmark & \xxmark & \ccmark \\[0.3em]

    \textbf{BoFT}~\cite{liu2023parameter} & 
    2023-11 & 
    2024-05 & 
    ICLR & 
    Llama-2-7B & 
    {\renewcommand{\arraystretch}{\arraystretchval}%
    \begin{tabular}[c]{@{}l@{}}Instruction Following\\[-0.3em]Math\end{tabular}} & 
    \xxmark & \xxmark & \ccmark \\[0.3em]

    \textbf{MoRA}~\cite{jiang2024mora} & 
    2024-05 & 
    -- & 
    arXiv & 
    {\renewcommand{\arraystretch}{\arraystretchval}%
    \begin{tabular}[c]{@{}l@{}}Llama-2-7B/13B\end{tabular}} & 
    {\renewcommand{\arraystretch}{\arraystretchval}%
    \begin{tabular}[c]{@{}l@{}}UUID\\[-0.3em]Math\\[-0.3em]Instruction Following\end{tabular}} & 
    \ccmark & \xxmark & \ccmark \\[0.3em]

    \textbf{Delta-LoRA}~\cite{zi2023delta} & 
    2023-09 & 
    -- & 
    arXiv & 
    GPT-2 Medium & 
    NLG & 
    \xxmark & \xxmark & \xxmark \\[0.3em]

    \textbf{Tied-LoRA}~\cite{renduchintala2024tied} & 
    2023-11 & 
    2024-06 & 
    NAACL & 
    {\renewcommand{\arraystretch}{\arraystretchval}%
    \begin{tabular}[c]{@{}l@{}}GPT-2B-001\\[-0.3em]Llama-2-7B\end{tabular}} & 
    {\renewcommand{\arraystretch}{\arraystretchval}%
    \begin{tabular}[c]{@{}l@{}}NLG\\[-0.3em]Commonsense\\[-0.3em]Math\end{tabular}} & 
    \xxmark & \xxmark & \ccmark \\[0.3em]

    \textbf{LoRETTA}~\cite{yang2024loretta} & 
    2024-02 & 
    2024-06 & 
    NAACL & 
    {\renewcommand{\arraystretch}{\arraystretchval}%
    \begin{tabular}[c]{@{}l@{}}Llama-2-7B/13B/70B\end{tabular}} & 
    {\renewcommand{\arraystretch}{\arraystretchval}%
    \begin{tabular}[c]{@{}l@{}}NLG\\[-0.3em]GLUE\end{tabular}} & 
    \xxmark & \xxmark & \xxmark \\[0.3em]

    \textbf{AutoLoRA}~\cite{zhang2024autolora} & 
    2024-03 & 
    2024-06 & 
    NAACL & 
    GPT-2 Medium & 
    NLG & 
    \xxmark & \xxmark & \xxmark \\[0.3em]

    \textbf{ALoRA}~\cite{liu2024alora} & 
    2024-03 & 
    2024-06 & 
    NAACL & 
    {\renewcommand{\arraystretch}{\arraystretchval}%
    \begin{tabular}[c]{@{}l@{}}GPT2-Large\\[-0.3em]Llama-2-7B\end{tabular}} & 
    {\renewcommand{\arraystretch}{\arraystretchval}%
    \begin{tabular}[c]{@{}l@{}}NLG\\[-0.3em]GLUE\\[-0.3em]Instruction Following\end{tabular}} & 
    \xxmark & \xxmark & \ccmark \\[0.3em]

    \textbf{RoSA}~\cite{nikdan2024rosa} & 
    2024-01 & 
    2024-07 & 
    ICML & 
    Llama-2-7B & 
    {\renewcommand{\arraystretch}{\arraystretchval}%
    \begin{tabular}[c]{@{}l@{}}NLG\\[-0.3em]Math\\[-0.3em]Code\\[-0.3em]Instruction Following\end{tabular}} & 
    \ccmark & \xxmark & \xxmark \\[0.3em]

    \textbf{LoRA+}~\cite{hayou2024lora+} & 
    2024-02 & 
    2024-07 & 
    ICML & 
    {\renewcommand{\arraystretch}{\arraystretchval}%
    \begin{tabular}[c]{@{}l@{}}GPT-2\\[-0.3em]Llama-1-7B\end{tabular}} & 
    {\renewcommand{\arraystretch}{\arraystretchval}%
    \begin{tabular}[c]{@{}l@{}}GLUE\\[-0.3em]Instruction Following\end{tabular}} & 
    \ccmark & \xxmark & \xxmark \\[0.3em]

    {\renewcommand{\arraystretch}{\arraystretchval}%
    \begin{tabular}[c]{@{}l@{}}
    \textbf{scaled AdamW}\\[-0.3em]
    \cite{zhang2024riemannian}
    \end{tabular}} &
    2024-02 & 
    2024-07 & 
    ICML & 
    {\renewcommand{\arraystretch}{\arraystretchval}%
    \begin{tabular}[c]{@{}l@{}}GPT-2 Medium\\[-0.3em]Mistral-7B-V0.1\end{tabular}} & 
    {\renewcommand{\arraystretch}{\arraystretchval}%
    \begin{tabular}[c]{@{}l@{}}NLG\\[-0.3em]GLUE\end{tabular}} & 
    \ccmark & \xxmark & \ccmark \\[0.3em]

    \textbf{DoRA}~\cite{dora} & 
    2024-02 & 
    2024-07 & 
    ICML & 
    {\renewcommand{\arraystretch}{\arraystretchval}%
    \begin{tabular}[c]{@{}l@{}}Llama-1-7B/13B\\[-0.3em]Llama-2-7B\\[-0.3em]Llama-3-8B\end{tabular}} & 
    Commonsense & 
    \xxmark & \xxmark & \ccmark \\[0.3em]

    \textbf{FLORA}~\cite{hao2024flora} & 
    2024-02 & 
    2024-07 & 
    ICML & 
    {\renewcommand{\arraystretch}{\arraystretchval}%
    \begin{tabular}[c]{@{}l@{}}GPT-2 -base/XL\end{tabular}} & 
    {\renewcommand{\arraystretch}{\arraystretchval}%
    \begin{tabular}[c]{@{}l@{}}Summarization\\[-0.3em]Translation\end{tabular}} & 
    \ccmark & \xxmark & \ccmark \\[0.3em]

    \textbf{FourierFT}~\cite{gao2024parameter} & 
    2024-05 & 
    2024-07 & 
    ICML & 
    {\renewcommand{\arraystretch}{\arraystretchval}%
    \begin{tabular}[c]{@{}l@{}}GPT-2 Medium/Large\\[-0.3em]Llama-1-7B/13B\\[-0.3em]Llama2-7B/13B\end{tabular}} & 
    {\renewcommand{\arraystretch}{\arraystretchval}%
    \begin{tabular}[c]{@{}l@{}}NLG\\[-0.3em]Instruction Following\end{tabular}} & 
    \ccmark & \ccmark & \ccmark \\[0.3em]

    \textbf{ResLoRA}~\cite{shi2024reslora} & 
    2024-02 & 
    2024-08 & 
    ACL & 
    Llama-2-7B & 
    {\renewcommand{\arraystretch}{\arraystretchval}%
    \begin{tabular}[c]{@{}l@{}}Math\\[-0.3em]Commonsense\end{tabular}} & 
    \xxmark & \xxmark & \ccmark \\[0.3em]

    \textbf{PLoRA}~\cite{meng2024periodiclora} & 
    2024-02 & 
    -- & 
    arXiv & 
    Llama-1-7B & 
    {\renewcommand{\arraystretch}{\arraystretchval}%
    \begin{tabular}[c]{@{}l@{}}Instruction Following\\[-0.3em]Math\end{tabular}} & 
    \ccmark & \xxmark & \ccmark \\[0.3em]

    \textbf{OLoRA}~\cite{buyukakyuz2024olora} & 
    2024-06 & 
    -- & 
    arXiv & 
    {\renewcommand{\arraystretch}{\arraystretchval}%
    \begin{tabular}[c]{@{}l@{}}Mistral-7B\\[-0.3em]LLaMA-2-7B\\[-0.3em]Tiny Llama-1.1B\\[-0.3em]Gemma-2B\\[-0.3em]OPT-1.3B\end{tabular}} & 
    {\renewcommand{\arraystretch}{\arraystretchval}%
    \begin{tabular}[c]{@{}l@{}}Commonsense\\[-0.3em]Instruction Following\end{tabular}} & 
    \xxmark & \xxmark & \ccmark \\[0.3em]

    \textbf{LamDA}~\cite{azizi2024lamda} & 
    2024-06 & 
    2024-11 & 
    EMNLP & 
    Llama-2-7B & 
    {\renewcommand{\arraystretch}{\arraystretchval}%
    \begin{tabular}[c]{@{}l@{}}NLG\\[-0.3em]Math\\[-0.3em]Commonsense\end{tabular}} & 
    \xxmark & \xxmark & \ccmark \\[0.3em]

    \textbf{PiSSA}~\cite{pissa} & 
    2024-04 & 
    2024-12 & 
    NeurIPS & 
    {\renewcommand{\arraystretch}{\arraystretchval}%
    \begin{tabular}[c]{@{}l@{}}Llama-2-7B/13B\\[-0.3em]Llama-3-8B/70B\\[-0.3em]Mistral-7B-v0.1\\[-0.3em]Gemma-7B\\[-0.3em]Qwen1.5-7B\\[-0.3em]Yi-1.5-34B\\[-0.3em]DeepSeek-MoE-16B\\[-0.3em]Mixtral-8x7B\end{tabular}} & 
    {\renewcommand{\arraystretch}{\arraystretchval}%
    \begin{tabular}[c]{@{}l@{}}Math\\[-0.3em]Code\\[-0.3em]Instruction Following\end{tabular}} & 
    \xxmark & \xxmark & \ccmark \\[0.3em]

    \textbf{VB-LoRA}~\cite{li2024vb} & 
    2024-05 & 
    2024-12 & 
    NeurIPS & 
    {\renewcommand{\arraystretch}{\arraystretchval}%
    \begin{tabular}[c]{@{}l@{}}GPT-2 Medium/Large\\[-0.3em]Llama-2-7B/13B\\[-0.3em]Mistral-7B-v0.1\\[-0.3em]Gemma-7B\end{tabular}} & 
    {\renewcommand{\arraystretch}{\arraystretchval}%
    \begin{tabular}[c]{@{}l@{}}NLG\\[-0.3em]Math\\[-0.3em]Instruction Following\end{tabular}} & 
    \xxmark & \xxmark & \xxmark \\[0.3em]

    \textbf{HRA}~\cite{yuan2024bridging} & 
    2024-05 & 
    2024-12 & 
    NeurIPS & 
    Llama-2-7B & 
    Math & 
    \xxmark & \xxmark & \xxmark \\[0.3em]

    \textbf{CorDA}~\cite{yang2024corda} & 
    2024-06 & 
    2024-12 & 
    NeurIPS & 
    {\renewcommand{\arraystretch}{\arraystretchval}%
    \begin{tabular}[c]{@{}l@{}}Llama-2-7B/13B\\[-0.3em]Llama-3-8B\\[-0.3em]Gemma-2-9B\end{tabular}} & 
    {\renewcommand{\arraystretch}{\arraystretchval}%
    \begin{tabular}[c]{@{}l@{}}Math\\[-0.3em]Code\\[-0.3em]Instruction Following\\[-0.3em]World Knowledge\end{tabular}} & 
    \xxmark & \xxmark & \ccmark \\[0.3em]

    \textbf{LoRA-GA}~\cite{wang2024lora} & 
    2024-07 & 
    2024-12 & 
    NeurIPS & 
    Llama-2-7B & 
    {\renewcommand{\arraystretch}{\arraystretchval}%
    \begin{tabular}[c]{@{}l@{}}Math\\[-0.3em]Code\\[-0.3em]Instruction Following\end{tabular}} & 
    \xxmark & \xxmark & \xxmark \\[0.3em]

    \textbf{RoAd}~\cite{liao20243} & 
    2024-09 & 
    2024-12 & 
    NeurIPS & 
    {\renewcommand{\arraystretch}{\arraystretchval}%
    \begin{tabular}[c]{@{}l@{}}Llama-1-7B/13B\\[-0.3em]Llama-2-7B\\[-0.3em]Llama-3-8B\end{tabular}} & 
    {\renewcommand{\arraystretch}{\arraystretchval}%
    \begin{tabular}[c]{@{}l@{}}Math\\[-0.3em]Commonsense\end{tabular}} & 
    \xxmark & \xxmark & \ccmark \\[0.3em]

    \textbf{LoRA-drop}~\cite{zhou2025lora} & 
    2024-02 & 
    2025-01 & 
    COLING & 
    Llama-2-7B & 
    {\renewcommand{\arraystretch}{\arraystretchval}%
    \begin{tabular}[c]{@{}l@{}}NLG\\[-0.3em]Summarization\\[-0.3em]GLUE\\[-0.3em]Math\end{tabular}} & 
    \xxmark & \xxmark & \xxmark \\[0.3em]

    \textbf{AG-LoRA}~\cite{wang2025activation} & 
    -- & 
    2025-01 & 
    IEEE Access & 
    Llama-1-7B & 
    Commonsense & 
    \xxmark & \xxmark & \ccmark \\[0.3em]

    \textbf{LoRA-Pro}~\cite{wang2024lorapro} &
    2024-07 & 
    2025-04 & 
    ICLR & 
    {\renewcommand{\arraystretch}{\arraystretchval}%
    \begin{tabular}[c]{@{}l@{}}Llama-1-7B\\[-0.3em]Llama-2-7B\\[-0.3em]Llama-3-8B\\[-0.3em]Llama-3.1-8B\end{tabular}} & 
    {\renewcommand{\arraystretch}{\arraystretchval}%
    \begin{tabular}[c]{@{}l@{}}Math\\[-0.3em]Code\\[-0.3em]Code\\[-0.3em]Instruction Following\end{tabular}} & 
    \ccmark & \xxmark & \ccmark \\[0.3em]
    
    {\renewcommand{\arraystretch}{\arraystretchval}%
    \begin{tabular}[c]{@{}l@{}}
    \textbf{LoRA-Dash}\\[-0.3em]
    \cite{si2024unleashing}
    \end{tabular}} &
    2024-09 & 
    2025-04 & 
    ICLR & 
    {\renewcommand{\arraystretch}{\arraystretchval}%
    \begin{tabular}[c]{@{}l@{}}Llama-1-7B\\[-0.3em]Llama-2-7B\\[-0.3em]Llama-3-8B\\[-0.3em]Qwen2.5-7B\end{tabular}} & 
    {\renewcommand{\arraystretch}{\arraystretchval}%
    \begin{tabular}[c]{@{}l@{}}GLUE\\[-0.3em]Commonsense\end{tabular}} & 
    \xxmark & \xxmark & \ccmark \\[0.3em]

    \textbf{KaSA}~\cite{wang2024kasa} & 
    2024-09 & 
    2025-04 & 
    ICLR & 
    {\renewcommand{\arraystretch}{\arraystretchval}%
    \begin{tabular}[c]{@{}l@{}}Llama-1-7B\\[-0.3em]Llama-2-7B\\[-0.3em]Llama-3-8B\\[-0.3em]Qwen2.5-7B\end{tabular}} & 
    {\renewcommand{\arraystretch}{\arraystretchval}%
    \begin{tabular}[c]{@{}l@{}}GLUE\\[-0.3em]Commonsense\\[-0.3em]Instruction Following\end{tabular}} & 
    \xxmark & \xxmark & \ccmark \\[0.3em]

    \textbf{RandLoRA}~\cite{albert2025randlora} & 
    2025-02 & 
    2025-04 & 
    ICLR & 
    {\renewcommand{\arraystretch}{\arraystretchval}%
    \begin{tabular}[c]{@{}l@{}}GPT-2 Medium\\[-0.3em]Qwen2-0.5B\\[-0.3em]Phi3-3B\\[-0.3em]Llama3-8B\end{tabular}} & 
    {\renewcommand{\arraystretch}{\arraystretchval}%
    \begin{tabular}[c]{@{}l@{}}NLG\\[-0.3em]Commonsense\end{tabular}} & 
    \xxmark & \xxmark & \ccmark \\[0.3em]

    \textbf{DeLoRA}~\cite{bini2025decoupling} & 
    2025-03 & 
    2025-04 & 
    ICLR & 
    {\renewcommand{\arraystretch}{\arraystretchval}%
    \begin{tabular}[c]{@{}l@{}}Llama-2-7B\\[-0.3em]Llama3-8B\end{tabular}} & 
    Commonsense & 
    \ccmark & \xxmark & \ccmark \\[0.3em]

    \textbf{HiRA}~\cite{huang2025hira} & 
    -- & 
    2025-04 & 
    ICLR & 
    {\renewcommand{\arraystretch}{\arraystretchval}%
    \begin{tabular}[c]{@{}l@{}}Llama-2-7B\\[-0.3em]Llama-3-8B\end{tabular}} & 
    {\renewcommand{\arraystretch}{\arraystretchval}%
    \begin{tabular}[c]{@{}l@{}}Math\\[-0.3em]Commonsense\\[-0.3em]Dialogue Generation\end{tabular}} & 
    \xxmark & \xxmark & \ccmark \\[0.3em]

    \textbf{MiLoRA}~\cite{milora} & 
    2024-06 & 
    2025-04 & 
    NAACL & 
    {\renewcommand{\arraystretch}{\arraystretchval}%
    \begin{tabular}[c]{@{}l@{}}Llama-2-7B\\[-0.3em]Llama-3-8B\\[-0.3em]Qwen2.5-7B\end{tabular}} & 
    {\renewcommand{\arraystretch}{\arraystretchval}%
    \begin{tabular}[c]{@{}l@{}}Math\\[-0.3em]Commonsense\\[-0.3em]Instruction Following\end{tabular}} & 
    \xxmark & \xxmark & \xxmark \\[0.3em]

    \textbf{SSMLoRA}~\cite{yu2025ssmlora} & 
    2025-02 & 
    2025-04 & 
    NAACL & 
    {\renewcommand{\arraystretch}{\arraystretchval}%
    \begin{tabular}[c]{@{}l@{}}GPT-2\\[-0.3em]Llama-2-7B/13B\end{tabular}} & 
    GLUE & 
    \ccmark & \xxmark & \ccmark \\[0.3em]

    \textbf{LoRA-One}~\cite{lora-one} & 
    2025-02 & 
    2025-07 & 
    ICML & 
    Llama-2-7B & 
    {\renewcommand{\arraystretch}{\arraystretchval}%
    \begin{tabular}[c]{@{}l@{}}Math\\[-0.3em]Code\\[-0.3em]Instruction Following\end{tabular}} & 
    \ccmark & \ccmark & \xxmark \\[0.3em]

    \textbf{Init[AB]}~\cite{initab} & 
    2025-05 & 
    2025-07 & 
    ICML & 
    Llama-3-8B & 
    {\renewcommand{\arraystretch}{\arraystretchval}%
    \begin{tabular}[c]{@{}l@{}}Arithmetic\\[-0.3em]Commonsense\end{tabular}} & 
    \ccmark & \xxmark & \xxmark \\[0.3em]

    \textbf{Lily}~\cite{zhong2025lily} & 
    2024-07 & 
    2025-07 & 
    ACL & 
    Llama-3-8B & Commonsense & 
    \xxmark & \xxmark & \xxmark \\[0.3em]
    
    \textbf{C3A}~\cite{chen2025parameter} & 
    2024-07 & 
    2025-07 & 
    ACL & 
    {\renewcommand{\arraystretch}{\arraystretchval}%
    \begin{tabular}[c]{@{}l@{}}Llama-2-7B\\[-0.3em]Llama-3-8B/70B\\[-0.3em]Mistral-7B\\[-0.3em]Mistral-8x7B\end{tabular}} & 
    {\renewcommand{\arraystretch}{\arraystretchval}%
    \begin{tabular}[c]{@{}l@{}}Math\\[-0.3em]Code\\[-0.3em]Commonsense\end{tabular}} & 
    \xxmark & \xxmark & \xxmark \\[0.3em]

    \textbf{SuLoRA}~\cite{ding2025sulora} & 
    -- & 
    2025-07 & 
    ACL & 
    Llama-2-7B & 
    Instruction Following & 
    \xxmark & \xxmark & \ccmark \\[0.3em]

    \textbf{BiDoRA}~\cite{qin2025bidora} & 
    2024-10 & 
    2025-08 & 
    TMLR & 
    GPT-2 Medium & 
    NLG & 
    \xxmark & \xxmark & \xxmark \\[0.3em]

    \textbf{HD-PiSSA}~\cite{wang2025hd} & 
    2025-05 & 
    2025-11 & 
    EMNLP & 
    {\renewcommand{\arraystretch}{\arraystretchval}%
    \begin{tabular}[c]{@{}l@{}}Llama-2-7B\\[-0.3em]Llama-3-8B\\[-0.3em]Mistral-7b-v0.1\end{tabular}} & 
    {\renewcommand{\arraystretch}{\arraystretchval}%
    \begin{tabular}[c]{@{}l@{}}Math\\[-0.3em]Code\end{tabular}} & 
    \xxmark & \xxmark & \ccmark \\[0.3em]

    \textbf{LoSiA}~\cite{wang2025losia} & 
    2025-07 & 
    2025-11 & 
    EMNLP & 
    {\renewcommand{\arraystretch}{\arraystretchval}%
    \begin{tabular}[c]{@{}l@{}}Gemma 2B\\[-0.3em]Llama-2-7B/13B\end{tabular}} & 
    {\renewcommand{\arraystretch}{\arraystretchval}%
    \begin{tabular}[c]{@{}l@{}}Math\\[-0.3em]Code\\[-0.3em]Commonsense\\[-0.3em]Instruction Following\end{tabular}} & 
    \ccmark & \xxmark & \ccmark \\[0.3em]

    {\renewcommand{\arraystretch}{\arraystretchval}%
    \begin{tabular}[c]{@{}l@{}}
    \textbf{Sensitivity-LoRA}\\[-0.3em]
    \cite{zhang2025sensitivity}
    \end{tabular}} &
    2025-09 & 
    2025-11 & 
    EMNLP & 
    {\renewcommand{\arraystretch}{\arraystretchval}%
    \begin{tabular}[c]{@{}l@{}}GPT-2 Large\\[-0.3em]Qwen2.5-7B/32B\\[-0.3em]Llama-3.1-8B\end{tabular}} & 
    {\renewcommand{\arraystretch}{\arraystretchval}%
    \begin{tabular}[c]{@{}l@{}}NLG\\[-0.3em]Instruction Following\end{tabular}} & 
    \xxmark & \xxmark & \xxmark \\[0.3em]

    \textbf{OHoRA}~\cite{zhang2025orthogonal} & 
    -- & 
    2025-11 & 
    EMNLP & 
    {\renewcommand{\arraystretch}{\arraystretchval}%
    \begin{tabular}[c]{@{}l@{}}Llama-2-7B\\[-0.3em]Llama-3-8B\\[-0.3em]Gemma-7B\\[-0.3em]Llama-3.1-8B-Inst\end{tabular}} & 
    {\renewcommand{\arraystretch}{\arraystretchval}%
    \begin{tabular}[c]{@{}l@{}}Math\\[-0.3em]Code\\[-0.3em]Commonsense\\[-0.3em]Instruction Following\end{tabular}} & 
    \xxmark & \xxmark & \ccmark \\[0.3em]

    \textbf{EVA}~\cite{paischer2024parameter} & 
    2024-10 & 
    2025-12 & 
    NeurIPS & 
    {\renewcommand{\arraystretch}{\arraystretchval}%
    \begin{tabular}[c]{@{}l@{}}Llama-2-7B\\[-0.3em]Gemma-2-70B\\[-0.3em]Llama-3.1-8B/70B\end{tabular}} & 
    {\renewcommand{\arraystretch}{\arraystretchval}%
    \begin{tabular}[c]{@{}l@{}}Math\\[-0.3em]Code\\[-0.3em]Commonsense\end{tabular}} & 
    \ccmark & \xxmark & \ccmark \\[0.3em]

    \textbf{GoRA}~\cite{he2026gora} & 
    2025-02 & 
    2025-12 & 
    NeurIPS & 
    {\renewcommand{\arraystretch}{\arraystretchval}%
    \begin{tabular}[c]{@{}l@{}}Llama-3.1-8B\\[-0.3em]Llama-2-7B\end{tabular}} & 
    {\renewcommand{\arraystretch}{\arraystretchval}%
    \begin{tabular}[c]{@{}l@{}}Math\\[-0.3em]Code\\[-0.3em]Instruction Following\end{tabular}} & 
    \xxmark & \xxmark & \xxmark \\[0.3em]

    \textbf{AuroRA}~\cite{dong2025aurora} & 
    2025-05 & 
    2025-12 & 
    NeurIPS & 
    Llama-3-8B & 
    Commonsense & 
    \xxmark & \xxmark & \ccmark \\[0.3em]

    \textbf{GraLoRA}~\cite{gralora} & 
    2025-05 & 
    2025-12 & 
    NeurIPS & 
    {\renewcommand{\arraystretch}{\arraystretchval}%
    \begin{tabular}[c]{@{}l@{}}Llama-3.1-8B/70B\\[-0.3em]Llama-3.2-3B\\[-0.3em]Qwen-2.5-1B/7B\end{tabular}} & 
    {\renewcommand{\arraystretch}{\arraystretchval}%
    \begin{tabular}[c]{@{}l@{}}Math\\[-0.3em]Code\\[-0.3em]Commonsense\end{tabular}} & 
    \xxmark & \xxmark & \ccmark \\[0.3em]

    \textbf{FlyLoRA}~\cite{zou2025flylora} & 
    2025-10 & 
    2025-12 & 
    NeurIPS & 
    {\renewcommand{\arraystretch}{\arraystretchval}%
    \begin{tabular}[c]{@{}l@{}}Llama-3.1-8B\\[-0.3em]Qwen-2.5-7B/14B\end{tabular}} & 
    {\renewcommand{\arraystretch}{\arraystretchval}%
    \begin{tabular}[c]{@{}l@{}}MMLU\\[-0.3em]Science\\[-0.3em]Math\\[-0.3em]Code\end{tabular}} & 
    \xxmark & \xxmark & \ccmark \\[0.3em]

    \textbf{DropLoRA}~\cite{zhang2025droplora} & 
    2025-08 & 
    -- & 
    arXiv & 
    {\renewcommand{\arraystretch}{\arraystretchval}%
    \begin{tabular}[c]{@{}l@{}}Llama-2-7B\\[-0.3em]Llama-3-8B\end{tabular}} & 
    {\renewcommand{\arraystretch}{\arraystretchval}%
    \begin{tabular}[c]{@{}l@{}}Math\\[-0.3em]Code\\[-0.3em]Commonsense\\[-0.3em]Instruction Following\end{tabular}} & 
    \xxmark & \xxmark & \ccmark \\[0.3em]

    \textbf{PrunedLoRA}~\cite{yu2025prunedlora} & 
    2025-09 & 
    -- &
    arXiv & 
    Llama-3-8B & 
    {\renewcommand{\arraystretch}{\arraystretchval}%
    \begin{tabular}[c]{@{}l@{}}Math\\[-0.3em]Commonsense\end{tabular}} & 
    \ccmark & \xxmark & \ccmark \\[0.3em]

    \textbf{LoRA-DA}~\cite{zhang2025lora-DA} & 
    2025-10 & 
    -- & 
    arXiv & 
    Llama-2-7B & 
    {\renewcommand{\arraystretch}{\arraystretchval}%
    \begin{tabular}[c]{@{}l@{}}Math\\[-0.3em]Commonsense\end{tabular}} & 
    \xxmark & \xxmark & \ccmark \\[0.3em]

    \textbf{ABM-LoRA}~\cite{lee2025abm} & 
    2025-11 & 
    -- & 
    arXiv & 
    Llama-2-7B & 
    Instruction Following & 
    \xxmark & \xxmark & \ccmark \\[0.3em]
    \textbf{MiSS}~\cite{kang2025missrevisitingtradeofflora} & 
    2024-09 & 
    2026-04 & 
    ICLR & 
    {\renewcommand{\arraystretch}{\arraystretchval}%
    \begin{tabular}[c]{@{}l@{}}Llama-2-7B/13B\\[-0.3em]Mistral-7B\\[-0.3em]Qwen3-4B\\[-0.3em]Llama-3.2-3B\end{tabular}} & 
    {\renewcommand{\arraystretch}{\arraystretchval}%
    \begin{tabular}[c]{@{}l@{}}Math\\[-0.3em]Code\\[-0.3em]Instruction Following\end{tabular}} & 
    \xxmark & \xxmark & \ccmark \\[0.3em]

    \textbf{LoFT}~\cite{tastan2025loft} & 
    2025-05 & 
    2026-04 & 
    ICLR & 
    {\renewcommand{\arraystretch}{\arraystretchval}%
    \begin{tabular}[c]{@{}l@{}}GPT-2-base/Large\\[-0.3em]Llama-1-7B\\[-0.3em]Llama-2-7B\\[-0.3em]Llama-3-8B\\[-0.3em]Llama-3.1-70B\end{tabular}} & 
    {\renewcommand{\arraystretch}{\arraystretchval}%
    \begin{tabular}[c]{@{}l@{}}NLG\\[-0.3em]Math\\[-0.3em]Code\\[-0.3em]Commonsense\end{tabular}} & 
    \xxmark & \xxmark & \ccmark \\[0.3em]
    
    \textbf{FlexLoRA}~\cite{liu2026flexlora} & 
    2026-01 & 
    2026-04 & 
    ICLR & 
    Llama-3-8B & 
    Commonsense & 
    \xxmark & \xxmark & \xxmark \\[0.3em]

    \textbf{Stable-LoRA}~\cite{wu2026stablelora} & 
    2026-03 & 
    2026-04 & 
    ICLR & 
   {\renewcommand{\arraystretch}{\arraystretchval}%
    \begin{tabular}[c]{@{}l@{}}Qwen-2-0.5B/1B\\[-0.3em]Llama-1-7B\\[-0.3em]Llama-3.1-8B\\[-0.3em]Llama-3.2-1B/3B\\[-0.3em]\end{tabular}} & 
    {\renewcommand{\arraystretch}{\arraystretchval}%
    \begin{tabular}[c]{@{}l@{}}Math\\[-0.3em]Commonsense\end{tabular}} & 
    \ccmark & \xxmark & \xxmark \\[0.3em]

    \textbf{RaLoRA}~\cite{ye2026gradient} & 
    -- & 
    2026-04 & 
    ICLR & 
    LLaMA-3.1-8B & 
    {\renewcommand{\arraystretch}{\arraystretchval}%
    \begin{tabular}[c]{@{}l@{}}Math\\[-0.3em]Code\\[-0.3em]Instruction Following\end{tabular}} & 
    \xxmark & \xxmark & \ccmark \\[0.3em]

    \textbf{GiVA}~\cite{gangwar2026giva} & 
    2026-04 & 
    2026-05 & 
    AISTATS & 
   {\renewcommand{\arraystretch}{\arraystretchval}%
    \begin{tabular}[c]{@{}l@{}}Qwen-2-0.5B\\[-0.3em]OLMo-2-7B\\[-0.3em]Phi-3-3.8B\\[-0.3em]Mistral-7B\\[-0.3em]\end{tabular}} & 
    {\renewcommand{\arraystretch}{\arraystretchval}%
    \begin{tabular}[c]{@{}l@{}}Math\\[-0.3em]Code\\[-0.3em]Commonsense\\[-0.3em]Instruction Following\end{tabular}} & 
    \ccmark & \xxmark & \xxmark \\[0.3em]

    \textbf{PEANuT}~\cite{zhong2026peanut} & 
    2024-10 & 
    2026-08 & 
    KDD & 
   {\renewcommand{\arraystretch}{\arraystretchval}%
    \begin{tabular}[c]{@{}l@{}}Llama-2-7B\\[-0.3em]Llama-3-8B\\[-0.3em]Qwen-3-8B\end{tabular}} & 
    {\renewcommand{\arraystretch}{\arraystretchval}%
    \begin{tabular}[c]{@{}l@{}}Math\\[-0.3em]Commonsense\end{tabular}} & 
    \xxmark & \xxmark & \xxmark \\[0.3em]

    \bottomrule

\end{longtable}
}
\endgroup

\clearpage
\section{Detailed Formulas of Selected LoRA Variants}\label{sec:variants-detailed-formulas}

We describe the detailed formulas of LoRA variants we select in this paper in the following. For readers who wish to explore more LoRA variants, we refer them to Appendix Table~\ref{tab:prior_studies} and existing LoRA surveys~\cite{zhu2025survey,he2026unified,mao2025survey,yang2024low,liang2025low} for more comprehensive coverage.

\subsection{Initialization Variants}\label{sec:variants-init-detailed-formulas}

\paragraph{OLoRA.}
\citet{buyukakyuz2024olora} proposed to improve the convergence behavior of LoRA by initializing the adaptation subspace with an orthonormal basis obtained from QR decomposition of pretrained weight matrices. 
OLoRA first performs QR decomposition:
\begin{equation*}
    W_{\text{pre}} = QR,
\end{equation*}
where $Q \in \mathbb{R}^{m \times m}$ is orthogonal and $R \in \mathbb{R}^{m \times n}$ is upper triangular. 
Let $Q_r \in \mathbb{R}^{m \times r}$ denote the first $r$ columns of $Q$, and $R_r \in \mathbb{R}^{r \times n}$ denote the first $r$ rows of $R$. 
OLoRA initializes LoRA adapters as
\begin{equation}
    B_0 = Q_r, \quad A_0 = R_r,
\end{equation}
such that the initialized adapter component satisfies
\begin{equation}
    B_0A_0 = Q_rR_r.
\end{equation}
To start fine-tuning from the pretrained weights, the \emph{residual matrix} is defined as
\begin{equation}
    W_{\text{res}} = W_{\text{pre}} -  B_0A_0 
    = W_{\text{pre}} - Q_rR_r,
\end{equation}
and the forward pass becomes:
\begin{equation}
    h = W_{\text{res}}x + BAx.
\end{equation}

\paragraph{PiSSA.} 
Aiming to address the potential slow convergence of LoRA,~\citet{pissa} proposed to initialize $BA$ with the top-$r$ principal components of the pretrained weight matrix and showed that this approach achieves faster convergence with loss and gradient norm curves similar to those of full fine-tuning. Suppose $W_{\text{pre}}$ admits SVD into $\sum_{i} \sigma_i u_i v_i^T$ where $\sigma_i$ are singular values in descending order, the LoRA adapter is initialized as:
\begin{equation*}
    B_0 A_0 = \sum_{i=1}^{r} \sigma_i u_i v_i^T,
\end{equation*}
with
\begin{equation}\label{eq:spectral_B0A0}
    B_0 = \sum_{i=1}^{r} \sqrt{\sigma_i} u_i e_i^T, \quad A_0 = \sum_{i=1}^{r} \sqrt{\sigma_i} e_i v_i^T,
\end{equation}
where $e_i \in \mathbb{R}^r$ denotes the $i$-th standard basis vector. To start fine-tuning from the pretrained weights, the \emph{residual matrix} is defined as
\begin{equation*}
    W_{\text{res}} = W_{\text{pre}} - B_0A_0 = \sum_{i=r+1}^{\min(m,n)} \sigma_i u_i v_i^T.
\end{equation*}

\paragraph{MiLoRA.} Concurrent to PiSSA,~\citet{milora} proposed an analogous approach targeting adaptation to new tasks while maximally retaining the pretrained model's knowledge. Instead of using principal components, MiLoRA initializes the low-rank adapters using bottom-$r$ minor components:
\begin{equation*}
    B_0 A_0 = \sum_{i=\min(m,n)-r+1}^{\min(m,n)} \sigma_i u_i v_i^T,
\end{equation*}
with $B_0$ and $A_0$ defined analogously to Eq.~\ref{eq:spectral_B0A0}. 
The residual matrix retains  principal components:
\begin{equation*}
    W_{\text{res}} = W_{\text{pre}} - B_0A_0 = \sum_{i=1}^{\min(m,n)-r} \sigma_i u_i v_i^T.
\end{equation*}
\citet{milora} showed experimentally that MiLoRA achieves superior downstream performance with less catastrophic forgetting.

\paragraph{Init[AB].} Several works have theoretically analyzed the initialization strategies of LoRA~\cite{hayou2024impact, xu2025understanding}. In particular,~\citet{hayou2024impact} confirmed that Init[A] (i.e., randomly initializing $A$ only, the default LoRA setting) generally leads to better performance than Init[B] (randomly initializing $B$ only). \citet{initab} further showed that initializing both matrices (i.e., Init[AB]) can provide even better advantage by balancing stability, training efficiency, and hyperparameter robustness. Specifically, Init[AB] initializes both matrices as $B_0 \sim \mathcal{N}(0, \sigma^2)$ and $A_0 \sim \mathcal{N}(0, \sigma^2)$. Since $B_0A_0 \neq 0$ in this case, the residual matrix $W_{\text{res}}$ is similarly introduced and utilized. Note that~\citet{initab} also proposed a variant, Init[AB+], which does not require $W_{\text{res}}$ and shows no discernible performance difference.

\paragraph{LoRA-GA.}
\citet{loraga} proposed to accelerate LoRA convergence by aligning the first-step update of the low-rank product with the full fine-tuning gradient. 
Let $G=\nabla_W\mathcal{L}(W_{\text{pre}})$ denote the gradient of the loss with respect to the full weight matrix, computed on a sampled downstream batch. 
Instead of decomposing the pretrained weight matrix as in PiSSA and MiLoRA, LoRA-GA performs SVD on the gradient matrix:
\begin{equation*}
    G = U\Sigma V^T.
\end{equation*}
The goal is to initialize $A$ and $B$ such that the update of the low-rank product approximates the full fine-tuning update:
\begin{equation*}
    \Delta(BA) \approx \zeta \Delta W,
\end{equation*}
for some positive scaling constant $\zeta$. 
Following the solution derived in LoRA-GA, the adapter matrices are initialized using selected singular directions of $G$:
\begin{equation*}
    B_0 = \frac{\sqrt{\zeta}}{\eta} U_{\mathcal{I}_B}, 
    \quad 
    A_0 = \frac{\sqrt{\zeta}}{\eta} V_{\mathcal{I}_A}^{T},
\end{equation*}
where $\eta$ denotes the scaling factor in LoRA-GA, and $\mathcal{I}_A$ and $\mathcal{I}_B$ are selected index sets of singular directions with $|\mathcal{I}_A|=|\mathcal{I}_B|=r$. 
Since $B_0A_0\neq0$, the residual matrix is introduced as
\begin{equation*}
    W_{\text{res}} = W_{\text{pre}} - \eta B_0A_0,
\end{equation*}
and the forward pass becomes:
\begin{equation*}
    h = W_{\text{res}}x + \eta BAx.
\end{equation*}
A notable follow-up is LoRA-One~\cite{lora-one}, which aims to address several limitations of LoRA-GA identified through theoretical analysis by using the top-$r$ singular directions of $G$ together with additional preconditioning and scaling choices.
\subsection{Architectural Modifications}\label{sec:variants-arch-detailed-formulas}

\paragraph{DoRA.} \citet{dora} proposed to learn \emph{magnitude}
and \emph{directional} updates of $W_{\text{pre}}$ separately.
Formally, the modified forward pass becomes:
\begin{equation}\label{eq:dora}
    h = {\gamma_r}\left( \frac{m}{\|W_{\text{pre}} + BA\|_c} \odot (W_{\text{pre}} + BA) \right) x,
\end{equation}
where $B$ and $A$ are responsible for directional updates and initialized as vanilla LoRA, while $m \in \mathbb{R}^{1 \times n}$ is an additional trainable magnitude vector initialized with $m_0 = \|W_{\text{pre}}\|_c$.
Note that $\|.\|_c$ denotes taking the column-wise norm of a matrix, while $\odot$ denotes element-wise multiplication with broadcasting across columns.
With only a marginal increase in trainable parameters introduced by the vector $m$, DoRA has been shown to consistently outperform LoRA, especially in regimes where the rank is small.

\paragraph{GraLoRA.} \citet{gralora} proposed GraLoRA to address the structural bottleneck of vanilla LoRA, where increasing the rank does not always lead to better performance and may even distort gradient propagation under skewed input-channel statistics. 
Instead of applying a single global low-rank adapter to the entire weight matrix, GraLoRA partitions the weight matrix into a $k \times k$ grid of sub-blocks, each equipped with its own local low-rank adapter. 
Let $k$ denote the number of splits along both the output and input dimensions. 
For $W_{\text{pre}} \in \mathbb{R}^{m \times n}$, the update is written as:
\begin{equation*}
    \Delta W_{\mathrm{GraLoRA}}
    =
    \begin{bmatrix}
        B_{1,1}A_{1,1} & B_{1,2}A_{1,2} & \cdots & B_{1,k}A_{1,k} \\
        B_{2,1}A_{2,1} & B_{2,2}A_{2,2} & \cdots & B_{2,k}A_{2,k} \\
        \vdots & \vdots & \ddots & \vdots \\
        B_{k,1}A_{k,1} & B_{k,2}A_{k,2} & \cdots & B_{k,k}A_{k,k}
    \end{bmatrix},
\end{equation*}
where
\begin{equation*}
    B_{i,j} \in \mathbb{R}^{\frac{m}{k} \times \frac{r}{k}},
    \quad
    A_{i,j} \in \mathbb{R}^{\frac{r}{k} \times \frac{n}{k}}.
\end{equation*}
The forward pass becomes:
\begin{equation*}
    h = W_{\text{pre}}x + \gamma_r \Delta W_{\mathrm{GraLoRA}}x.
\end{equation*}
When $k=1$, GraLoRA reduces to vanilla LoRA. 
When $k>1$, the total number of trainable parameters remains identical to vanilla LoRA under the same overall rank:
\begin{equation*}
    k^2
    \left(
        \frac{m}{k}\frac{r}{k}
        +
        \frac{r}{k}\frac{n}{k}
    \right)
    =
    (m+n)r.
\end{equation*}
By introducing localized adapters, GraLoRA provides more fine-grained control over different subregions of the weight matrix and reduces global gradient distortion caused by input outliers. 
Moreover, the effective rank of the resulting update can increase up to $kr$, thereby enhancing expressive capacity without increasing the trainable parameter count. 
\citet{gralora} showed that this design consistently improves over vanilla LoRA across multiple tasks, especially under larger-rank settings.

\paragraph{RandLoRA.}
\citet{albert2025randlora} proposed RandLoRA to examine whether the performance gap between LoRA and full fine-tuning stems from the reduced number of trainable parameters or from the low-rank constraint itself. 
The key idea is to maintain parameter efficiency while enabling full-rank updates through fixed random low-rank bases and trainable diagonal scaling matrices. 
Following their notation, let $W_{\text{pre}} \in \mathbb{R}^{m \times n}$ with $n \leq m$. 
RandLoRA represents the update as a sum of low-rank random basis components:
\begin{equation*}
    \Delta W_{\mathrm{RandLoRA}}
    =
    \sum_{j=1}^{q}
    B_j \Lambda_j A \Gamma_j,
\end{equation*}
where $B_j \in \mathbb{R}^{m \times r_b}$ and $A \in \mathbb{R}^{r_b \times n}$ are non-trainable random matrices, while $\Lambda_j \in \mathbb{R}^{r_b \times r_b}$ and $\Gamma_j \in \mathbb{R}^{n \times n}$ are trainable diagonal matrices. 
The forward pass becomes:
\begin{equation*}
    h = W_{\text{pre}}x + \gamma_r \Delta W_{\mathrm{RandLoRA}}x.
\end{equation*}
Since only the diagonal scaling matrices are optimized, the number of trainable parameters is substantially restricted. 
When $q=n/r_b$, RandLoRA can form a full-rank update with
\begin{equation*}
    q(r_b+n)
    =
    n+\frac{n^2}{r_b}
\end{equation*}
trainable parameters. 
This differs from vanilla LoRA, which directly learns a single rank-$r$ product $BA$ and is therefore constrained to rank at most $r$. 
RandLoRA's design rationale is to trade the flexibility of learning an arbitrary low-rank basis for broader spectral coverage through multiple fixed random bases. 
\citet{albert2025randlora} showed that RandLoRA can reduce or even eliminate the gap between LoRA and full fine-tuning in settings where full-rank updates are beneficial, while maintaining parameter and memory efficiency during training.

\subsection{Optimization Adjustments}

\paragraph{LoFT.}
\citet{tastan2025loft} proposed LoFT to reduce the optimization gap between LoRA and full fine-tuning by aligning the optimizer's internal dynamics with those of updating all model weights.
Following their notation, write the adapted weight as
\begin{equation*}
    W = W_0 + UV^T,
\end{equation*}
where $U \in \mathbb{R}^{m \times r}$ and $V \in \mathbb{R}^{n \times r}$ are trainable low-rank factors.
For a scalar loss $f(W)$, a full fine-tuning gradient descent step is:
\begin{equation*}
    W^+ = W - \eta \nabla_W f(W).
\end{equation*}
In contrast, simultaneous updates of $U$ and $V$ induce the full-space update:
\begin{equation*}
    W^+
    =
    W
    -
    \eta
    \left(
        \nabla_W f(W)VV^T
        +
        UU^T\nabla_W f(W)
    \right)
    +
    \eta^2 \nabla_W f(W)VU^T\nabla_W f(W),
\end{equation*}
where the last term is absent in full fine-tuning.
LoFT therefore uses alternating updates, updating only one low-rank factor at a time.
When updating $U$, this gives:
\begin{equation*}
    W^+ = W - \eta \nabla_W f(W)VV^T.
\end{equation*}
To address the scale ambiguity $UV^T=(cU)(V/c)^T$, LoFT applies scaled gradients:
\begin{equation*}
    \widetilde{\nabla}_U f(W)
    =
    \nabla_U f(W)(V^TV)^{-1},
    \quad
    \widetilde{\nabla}_V f(W)
    =
    \nabla_V f(W)(U^TU)^{-1},
\end{equation*}
which yield the projected full-space update
\begin{equation*}
    W^+
    =
    W
    -
    \eta \nabla_W f(W)P_V,
    \quad
    P_V = V(V^TV)^{-1}V^T.
\end{equation*}
Beyond gradient alignment, LoFT calibrates Adam-style optimizer states in the same low-rank subspace.
For the first moment, when updating $U$, it uses:
\begin{equation*}
    m_k^U
    =
    \beta_1 m_{k-1}^U C_k^V
    +
    (1-\beta_1)\widetilde{\nabla}_U f(W_k),
    \quad
    C_k^V
    =
    (V_{k-1}^T V_k)(V_k^T V_k)^{-1}.
\end{equation*}
An analogous update is applied when updating $V$.
For the second moment, LoFT similarly accumulates cross-terms for variance recalibration after projection:
\begin{equation*}
    p_k^U
    =
    \beta_2 p_{k-1}^U
    (C_k^V \otimes C_k^V)
    +
    (1-\beta_2)
    \left(
        \widetilde{\nabla}_U f(W_k)
        \bullet
        \widetilde{\nabla}_U f(W_k)
    \right),
\end{equation*}
where $\otimes$ denotes the Kronecker product and $\bullet$ denotes the transposed Khatri--Rao product.
Together, alternating updates, scaled gradients, and optimizer-state calibration allow LoFT to better approximate full fine-tuning dynamics.
\citet{tastan2025loft} showed that LoFT narrows the performance gap between LoRA and full fine-tuning without increasing inference cost.

\section{Fine-tuning Implementation Details}\label{sec:implementation_details}

\subsection{Models}
Following standard practice in PEFT research to ensure results purely reflect the training data, we use base versions (not instruction-tuned) for all models.
Specifically, we utilize the official checkpoints hosted on Hugging Face~\cite{wolf2019huggingface}: \texttt{Qwen3-0.6B-Base}\footnote{\url{https://huggingface.co/Qwen/Qwen3-0.6B-Base}}, \texttt{gemma-3-1b-pt}\footnote{\url{https://huggingface.co/google/gemma-3-1b-pt}}, \texttt{Llama-2-7b-hf}\footnote{\url{https://huggingface.co/meta-llama/Llama-2-7b-hf}}, and \texttt{Llama-2-13b-hf}\footnote{\url{https://huggingface.co/meta-llama/Llama-2-13b-hf}}.


\subsection{Training Hyperparameter Search Ranges}
See Table~\ref{tab:model_task_config} for hyperparameter search ranges for each model-task combination. For all experiments on Qwen and Gemma, we conduct three independent trainings and report the mean and standard deviation.

\begin{table}[h]
\centering
\footnotesize
\renewcommand{\arraystretch}{1.35} 
\setlength{\tabcolsep}{3.5pt} 
\setlength{\aboverulesep}{0pt}
\setlength{\belowrulesep}{0pt}
\adjustbox{max width=0.85\textwidth}{%
\begin{tabular}{c|c|c|c|c}
\toprule
\textbf{Model} & 
\textbf{Task} & 
\makecell{\textbf{Rank} \textbf{($r$)}} & 
\makecell{\textbf{Batch} \textbf{($B$)}} & 
\makecell{\textbf{Learning Rate} \textbf{($\eta$)}} \\
\midrule 
\multirow{3}{*}{\textbf{Qwen3-0.6B}}
& \multirow{2}{*}{Math}
& 8 
& \{16, 64, 128\} 
& 1.1247e-5 -- 6.3246e-3 \\ 
& 
& 128 
& 64 
& 2.0000e-6 -- 2.0000e-3 \\
\cmidrule{2-5} 
& Inst
& 128 
& 64
& 2.0000e-6 -- 2.0000e-3 \\

\cmidrule[0.5pt]{1-5} 

\multirow{4}{*}{\textbf{Gemma-3-1B}}
& \multirow{1}{*}{CS}
& 4
& 64 
& 2.0000e-6 -- 2.0000e-2 \\
\cmidrule{2-5} 
& \multirow{2}{*}{Math}
& \{4, 8, 16, 32, 64, 128, 256\}
& 64 
& 1.1247e-5 -- 6.3246e-3 \\
&
& 128 
& \{16, 64, 128\} 
& 1.1247e-5 -- 6.3246e-3 \\
\cmidrule{2-5} 
& \multirow{1}{*}{Code}
& \{4, 8, 16, 32, 64, 128, 256\}
& 64 
& 1.1247e-5 -- 6.3246e-3 \\
\cmidrule[0.5pt]{1-5}

\multirow{2}{*}{\textbf{Llama-2-7B}}
& \multirow{1}{*}{Math}
& \{8, 32, 128\}
& \{16, 128\}
& 2.0000e-5 -- 3.5566e-3 \\
\cmidrule{2-5} 
& \multirow{1}{*}{Code}
& \{8, 32, 128\}
& \{16, 128\}
& 2.0000e-5 -- 3.5566e-3 \\
\cmidrule[0.5pt]{1-5}

\multirow{1}{*}{\textbf{Llama-2-13B}}
& \multirow{1}{*}{Math}
& \{8, 128\}
& 64
& 2.0000e-5 -- 2.0000e-3 \\
\bottomrule
\end{tabular}
}
\vspace{0.5em}
\caption{Summary of models, tasks, ranks, and hyperparameter search ranges. Learning rates are tuned evenly in logarithmic scale: \{1.1247e*, 2.0000e*, 3.5566e*, 6.3246e*\} per order of magnitude.}
\label{tab:model_task_config}
\end{table}

\subsection{Fixed Training Hyperparameters}
Except for tunable hyperparameters (i.e., learning rate and batch size), all other training configurations remain fixed and the same for all experiments; the values are summarized in Table~\ref{tab:other_hyperparameters}. Note that these configurations primarily follow PiSSA, thus may differ from those of other considered PEFT methods. For example, MiLoRA, DoRA, and Init[AB] employ linear decay instead of cosine annealing for learning rate scheduling. Additionally, MiLoRA and DoRA use fixed warmup steps (100) rather than 3\% of total training steps, and apply a dropout rate of 0.05 instead of no dropout. Furthermore, while we place low-rank adapters on all linear layers, these methods exclude output projections (\texttt{out\_proj}) or gate matrices (\texttt{gate\_proj}) in several of their experiments.

\begin{table}[h]
\centering
\footnotesize
\renewcommand{\arraystretch}{1.35}
\setlength{\tabcolsep}{3.5pt} 
\adjustbox{max width=0.65\textwidth}{
\begin{tabular}{l|l}
\hline
\textbf{Configuration} & \textbf{Value} \\
\hline
Epoch & 1 \\
$\alpha$ & $r$ \\
Optimizer & AdamW~\cite{adamw} \\
LR scheduler & Cosine annealing with warmup \\
Warmup ratio & 3\% \\
Dropout & None \\
Weight Decay & None \\
Adapter placement & All linear layers (except for the LM head) \\
Base model precision & BFloat16~\citep{bfloat16}\tablefootnote{Following PiSSA's codebase, normalization layers and \texttt{gate\_proj} matrices of the pretrained model are converted back to Float32 after the BFloat16 sampling and before training.} \\
Adapter precision & Float32 \\
Max sequence length & 512 \\
\hline
\end{tabular}
}
\vspace{0.5em}
\caption{Fixed training configurations across all experiments. 
$\alpha$ equals the LoRA rank $r$, resulting in a scaling factor $\gamma_r=1$.
Adapters are applied to all linear projection layers except for the final language modeling head (i.e., \texttt{q\_proj}, \texttt{k\_proj}, \texttt{v\_proj}, \texttt{o\_proj}, \texttt{gate\_proj}, \texttt{up\_proj}, \texttt{down\_proj}).}
\label{tab:other_hyperparameters}
\end{table}

\FloatBarrier



\subsection{Data, Code, Libraries, and Hardware}\label{sec:data_code_lib_hard}
We use the PiSSA codebase~\cite{pissa} as the core framework and extend it into a unified implementation covering all ten LoRA-based methods considered in this study. Specifically, LoRA, OLoRA, PiSSA, LoRA-GA, DoRA, GraLoRA, and RandLoRA are implemented using the built-in interfaces of the official PEFT library~\cite{peft}. For MiLoRA, Init[AB], and LoFT, we refer to their official codebases\footnote{
\begin{tabular}[t]{@{}l@{}}
\url{https://github.com/sufenlp/MiLoRA} \\
\url{https://github.com/Leopold1423/non_zero_lora-icml25} \\
\url{https://github.com/tnurbek/loft}
\end{tabular}
} 
for essential functions and implementation details, which are then integrated into the same unified experimental codebase.

Note that while the original LoRA paper used Kaiming Normal initialization, we follow its official implementation and the widely-used PEFT library to use Kaiming Uniform instead in our experiments. The results are expected to be similar (cf.~\citet[Table 2]{pissa}). Additionally, while~\citet{initab} also proposed a variant, Init[AB+], which does not require $W_{\text{res}}$ and shows no discernible performance difference, we chose to implement the default Init[AB].

For commonsense reasoning tasks, we leveraged the dataset released by~\citet{hu2023llm}\footnote{\url{https://github.com/AGI-Edgerunners/LLM-Adapters/blob/main/ft-training_set/commonsense_15k.json}}.
For mathematical reasoning and code generation tasks, we used the preprocessed dataset released by~\citet{pissa}\footnote{\url{https://huggingface.co/datasets/fxmeng/pissa-dataset}}.
For instruction-following tasks, we trained models on Alpaca-cleaned, which is publicly available on HuggingFace\footnote{\url{https://huggingface.co/datasets/yahma/alpaca-cleaned}}.

PyTorch~\cite{paszke2019pytorch} version 2.7.1 is used for implementation. All experiments are conducted on four GPUs (either 4$\times$ Nvidia RTX 3090 or 4$\times$ Nvidia A6000). We employ DeepSpeed~\cite{rasley2020deepspeed} for parallel training and vLLM~\cite{kwon2023efficient} for parallel inference.
During inference, we apply greedy decoding (i.e., temperature set to 0), and utilize the EvalPlus~\cite{evalplus} framework to evaluate pass@1 for code generation tasks.

All fine-tuning experiments (except those for Llama) are conducted with three independent runs and reported with means and standard deviations. 
The sources of randomness are controlled by explicitly fixing random seeds across Python, NumPy, and PyTorch using the code snippet shown below.

\begin{center}
    \begin{minipage}{0.55\linewidth}
        \begin{lstlisting}[
            frame=single, 
            basicstyle=\ttfamily\fontsize{7.5}{9}\selectfont, 
            breaklines=true
        ]
def seed_everything(seed):
    random.seed(seed)
    os.environ['PYTHONHASHSEED'] = str(seed)
    np.random.seed(seed)
    torch.manual_seed(seed)
    torch.cuda.manual_seed(seed)
    torch.backends.cudnn.deterministic = True
    torch.backends.cudnn.benchmark = True
        \end{lstlisting}
    \end{minipage}
\end{center}

\section{On LoRA Scaling Factor}\label{sec:on_scaling_factor}

The configuration of the LoRA alpha parameter ($\alpha$) generally follows two paradigms: (1) setting it to a fixed constant across LoRA ranks (typically 32 or 64), or (2) scaling it with the LoRA rank, often following $\alpha=r$ or $\alpha=2r$, which results in a scaling factor $\gamma_r$ of $1$ or $2$, respectively. The configurations adopted for decoder-LLMs in the considered core LoRA variants are summarized as follows:
\begin{itemize}
    \item \textbf{PiSSA:} $\gamma_r = 1$ ($\alpha=r$ for all $r$).
    \item \textbf{MiLoRA:} $\gamma_r = 2$ for vanilla LoRA; $\gamma_r=1$ for both MiLoRA and PiSSA.
    \item \textbf{Init[AB]:} $\gamma_r = 1$ ($r=16$ and $\alpha=16$).
    \item \textbf{DoRA:} $\gamma_r = 2$ ($\alpha=2r$ for all $r$).
\end{itemize}

Evidently, the methods considered in this paper adhere to the second paradigm. Consequently, we adopt the setting $\alpha = r$ ($\gamma_r=1$) for all our experiments. We refer readers to several prior studies that have explored the optimal LoRA scaling factor: 
\citet{kalajdzievski2023rank} argued that $\gamma_r$ should scale with the square root of $r$ ($\gamma_r = \alpha/\sqrt{r}$), rather than linearly ($\gamma_r = \alpha/r$), though the optimal $\alpha$ setup remains unclear.
Empirically, \citet{biderman2024lora} demonstrated through joint sweeps of $\alpha$ and learning rates that $\alpha=2r$ ($\gamma_r=2$) is the optimal choice, with $\alpha=r$ ($\gamma_r=1$) performing only marginally worse (cf.~\citet[Appendix B.2, Figure S3]{biderman2024lora}). Notably,~\citet{zhang2025primacy} recently unified the learning rate, scaling factor, and initialization under a single theoretical framework, suggesting that tuning the learning rate is theoretically equivalent to tuning the scaling factor~\cite{fan2025make}. This further validates our decision to fix the scaling factor in our experiments.

\clearpage

\clearpage
\FloatBarrier

\section{Details of Hyperparameter Search Results}\label{sec:detail_table}
Sections~\ref{sec:table_qwen}, \ref{sec:table_gemma}, and \ref{sec:table_llama} present several detailed hyperparameter search results on Qwen3-0.6B, Gemma-3-1B, and Llama-2-7B, respectively. Note that Tables~\ref{tab:qwen_math_r128}, \ref{tab:llama_math_r128}, and \ref{tab:llama_code_r128} correspond to the detailed numerical results for Figures~\ref{fig:main-evidence}, \ref{fig:llama-main-math}, and \ref{fig:llama-main-code} in the main text, respectively. Additionally, Tables~\ref{tab:llama_math_r8}, \ref{tab:llama_math_r32}, and \ref{tab:llama_math_r128} provide the hyperparameter search details across adapter ranks for Appendix Figure~\ref{fig:perf_across_ranks_llama_math}, while Tables~\ref{tab:llama_code_r8}, \ref{tab:llama_code_r32}, and \ref{tab:llama_code_r128} correspond to Appendix Figure~\ref{fig:perf_across_ranks_llama_code}.

\subsection{Qwen3-0.6B}\label{sec:table_qwen}
\subsubsection{Mathematical Reasoning}

\begin{table*}[h]
\centering
\renewcommand{\arraystretch}{1.75} 
\setlength{\tabcolsep}{0.7pt}
\setlength{\aboverulesep}{0pt}
\setlength{\belowrulesep}{0pt}
\adjustbox{max width=\textwidth}{%
\begin{tabular}{l|c|*{12}{c}}
\toprule
\multirow{2}{*}{\centering\arraybackslash \textbf{Methods}} &
\multirow{2}{*}{\centering\arraybackslash \makecell{\textbf{Batch}\\\textbf{Size}}} &
\multicolumn{12}{c}{\raisebox{0.4ex}{\textbf{Learning Rate}}} \\
\cline{3-14}
& &
\textbf{1.12e-05} & \textbf{2.00e-05} & \textbf{3.56e-05} & \textbf{6.32e-05} & 
\textbf{1.12e-04} & \textbf{2.00e-04} & \textbf{3.56e-04} & \textbf{6.32e-04} & 
\textbf{1.12e-03} & \textbf{2.00e-03} & \textbf{3.56e-03} & \textbf{6.32e-03} \\
\midrule
\multirow{3}{*}{\textbf{LoRA}}
& \textbf{16} \small
& 38.94$_{\pm 1.02}$ & 46.49$_{\pm 1.17}$ & 47.78$_{\pm 0.47}$ & 47.85$_{\pm 0.42}$
& 48.13$_{\pm 0.40}$ & 48.90$_{\pm 0.41}$ & 48.41$_{\pm 0.59}$ & \textbf{49.05}$_{\pm 0.49}$
& 47.64$_{\pm 0.86}$ & 44.00$_{\pm 0.45}$ & 26.59$_{\pm 21.82}$ & 5.03$_{\pm 5.33}$ \\
\normalsize & \textbf{64} \small
& 29.53$_{\pm 0.18}$ & 33.02$_{\pm 0.28}$ & 39.15$_{\pm 0.33}$ & 46.49$_{\pm 0.25}$
& 48.16$_{\pm 0.24}$ & 48.39$_{\pm 0.29}$ & 48.95$_{\pm 0.23}$ & \textbf{48.99}$_{\pm 0.40}$
& 48.73$_{\pm 0.10}$ & 48.14$_{\pm 0.42}$ & 43.92$_{\pm 0.48}$ & 1.28$_{\pm 0.05}$ \\
\normalsize & \textbf{128} \small
& 22.30$_{\pm 0.30}$ & 30.22$_{\pm 1.63}$ & 33.64$_{\pm 0.13}$ & 40.64$_{\pm 1.48}$
& 47.88$_{\pm 0.61}$ & 48.38$_{\pm 0.19}$ & 48.38$_{\pm 0.01}$ & 48.69$_{\pm 0.57}$
& \textbf{48.72}$_{\pm 0.04}$ & 48.33$_{\pm 0.54}$ & 31.32$_{\pm 26.20}$ & 1.13$_{\pm 0.58}$ \\
\cline{1-14}

\multirow{3}{*}{\textbf{DoRA}}
\normalsize & \textbf{16} \small
& 42.03$_{\pm 1.63}$ & 47.36$_{\pm 0.59}$ & 48.10$_{\pm 0.12}$ & 48.10$_{\pm 0.48}$
& 48.29$_{\pm 0.08}$ & \textbf{48.80}$_{\pm 0.64}$ & 48.60$_{\pm 0.17}$ & 48.70$_{\pm 0.05}$
& 46.30$_{\pm 0.29}$ & 42.67$_{\pm 0.19}$ & 35.93$_{\pm 0.44}$ & 1.31$_{\pm 0.19}$ \\
\normalsize & \textbf{64} \small
& 38.69$_{\pm 1.24}$ & 37.60$_{\pm 1.22}$ & 40.65$_{\pm 1.58}$ & 47.06$_{\pm 0.50}$
& 48.41$_{\pm 0.31}$ & 48.03$_{\pm 0.18}$ & \textbf{49.07}$_{\pm 0.03}$ & 48.87$_{\pm 0.82}$
& 48.55$_{\pm 0.43}$ & 47.31$_{\pm 0.48}$ & 44.61$_{\pm 0.00}$ & 1.10$_{\pm 0.31}$ \\
\normalsize & \textbf{128} \small
& 33.37$_{\pm 1.56}$ & 36.85$_{\pm 0.98}$ & 36.56$_{\pm 0.41}$ & 43.03$_{\pm 1.11}$
& 48.46$_{\pm 0.54}$ & 47.94$_{\pm 0.30}$ & 48.30$_{\pm 0.24}$ & 48.41$_{\pm 0.72}$
& \textbf{48.59}$_{\pm 0.10}$ & 47.75$_{\pm 0.47}$ & 46.08$_{\pm 0.10}$ & 14.91$_{\pm 23.56}$ \\

\cline{1-14}

\multirow{3}{*}{\textbf{Init[AB]}}
\normalsize & \textbf{16} \small
& 36.53$_{\pm 2.20}$ & 41.67$_{\pm 1.99}$ & 45.47$_{\pm 1.30}$ & 48.07$_{\pm 0.70}$
& 48.28$_{\pm 0.72}$ & \textbf{48.66}$_{\pm 0.31}$ & 48.53$_{\pm 0.48}$ & 48.18$_{\pm 0.49}$
& 46.79$_{\pm 0.19}$ & 42.32$_{\pm 0.39}$ & 38.65$_{\pm 0.80}$ & 20.67$_{\pm 27.96}$ \\
\normalsize & \textbf{64} \small
& 35.78$_{\pm 0.54}$ & 35.15$_{\pm 1.09}$ & 37.85$_{\pm 0.08}$ & 40.04$_{\pm 1.50}$
& 45.03$_{\pm 1.13}$ & 48.34$_{\pm 0.29}$ & 48.53$_{\pm 0.07}$ & 48.45$_{\pm 0.50}$
& \textbf{48.68}$_{\pm 0.15}$ & 47.11$_{\pm 0.43}$ & 43.13$_{\pm 0.80}$ & 1.36$_{\pm 0.06}$ \\
\normalsize & \textbf{128} \small
& 31.34$_{\pm 1.30}$ & 32.44$_{\pm 0.70}$ & 36.21$_{\pm 0.32}$ & 35.29$_{\pm 1.90}$
& 41.73$_{\pm 1.33}$ & 47.06$_{\pm 1.29}$ & 48.38$_{\pm 0.60}$ & \textbf{48.57}$_{\pm 0.39}$
& 48.12$_{\pm 0.01}$ & 48.34$_{\pm 0.66}$ & 46.79$_{\pm 0.22}$ & 0.99$_{\pm 0.52}$ \\
\cline{1-14}

\multirow{3}{*}{\textbf{MiLoRA}}
\normalsize & \textbf{16} \small
& 39.42$_{\pm 0.87}$ & 45.09$_{\pm 0.19}$ & 44.76$_{\pm 0.78}$ & 45.92$_{\pm 0.81}$
& 49.16$_{\pm 0.37}$ & \textbf{49.36}$_{\pm 0.09}$ & 48.93$_{\pm 0.37}$ & 48.08$_{\pm 0.06}$
& 46.09$_{\pm 0.55}$ & 43.39$_{\pm 0.52}$ & 25.53$_{\pm 20.93}$ & 1.47$_{\pm 0.07}$ \\
\normalsize & \textbf{64} \small
& 32.25$_{\pm 1.20}$ & 38.33$_{\pm 1.24}$ & 45.72$_{\pm 0.88}$ & 44.08$_{\pm 1.50}$
& 47.32$_{\pm 0.05}$ & 48.69$_{\pm 0.32}$ & \textbf{49.40}$_{\pm 0.01}$ & 49.06$_{\pm 0.18}$
& 48.90$_{\pm 0.37}$ & 47.02$_{\pm 0.38}$ & 43.98$_{\pm 1.06}$ & 1.07$_{\pm 0.50}$ \\
\normalsize & \textbf{128} \small
& 30.94$_{\pm 0.32}$ & 33.71$_{\pm 0.34}$ & 35.03$_{\pm 0.69}$ & 40.49$_{\pm 0.39}$
& 44.27$_{\pm 0.12}$ & 48.08$_{\pm 0.42}$ & 48.65$_{\pm 0.18}$ & \textbf{49.37}$_{\pm 0.19}$
& 49.33$_{\pm 0.52}$ & 48.12$_{\pm 0.74}$ & 46.91$_{\pm 0.06}$ & 1.24$_{\pm 0.33}$ \\
\cline{1-14}

\multirow{3}{*}{\textbf{PiSSA}}
\normalsize & \textbf{16} \small
& 47.10$_{\pm 0.28}$ & 44.80$_{\pm 0.71}$ & 46.45$_{\pm 0.61}$ & \textbf{48.37}$_{\pm 0.36}$
& 48.30$_{\pm 0.26}$ & 47.42$_{\pm 0.22}$ & 45.47$_{\pm 0.36}$ & 42.43$_{\pm 0.03}$
& 38.83$_{\pm 0.40}$ & 33.14$_{\pm 0.41}$ & 24.63$_{\pm 0.98}$ & 14.82$_{\pm 19.10}$ \\
\normalsize & \textbf{64} \small
& 44.20$_{\pm 0.12}$ & 44.12$_{\pm 0.69}$ & 47.54$_{\pm 0.50}$ & 48.25$_{\pm 0.14}$
& \textbf{48.46}$_{\pm 0.59}$ & 48.43$_{\pm 0.12}$ & 47.94$_{\pm 0.39}$ & 47.23$_{\pm 0.19}$
& 43.94$_{\pm 0.04}$ & 40.15$_{\pm 0.20}$ & 35.37$_{\pm 0.55}$ & 18.90$_{\pm 15.52}$ \\
\normalsize & \textbf{128} \small
& 39.49$_{\pm 2.26}$ & 43.64$_{\pm 0.51}$ & 43.61$_{\pm 0.50}$ & 46.42$_{\pm 0.10}$
& \textbf{48.51}$_{\pm 0.31}$ & 48.24$_{\pm 0.18}$ & 48.06$_{\pm 0.72}$ & 47.70$_{\pm 0.36}$
& 45.56$_{\pm 0.15}$ & 43.54$_{\pm 0.41}$ & 38.85$_{\pm 0.18}$ & 33.36$_{\pm 0.44}$ \\

\bottomrule
\end{tabular}
}
\caption{Performance of \textbf{Qwen3-0.6B} fine-tuned on \textbf{mathematical reasoning} tasks with \textbf{rank = 8}.
}
\label{tab:qwen_math_r8}
\end{table*}

\begin{table*}[h]
\centering
\renewcommand{\arraystretch}{1.75} 
\setlength{\tabcolsep}{0.85pt}
\setlength{\aboverulesep}{0pt}
\setlength{\belowrulesep}{0pt}
\adjustbox{max width=\textwidth}{
\begin{tabular}{l|c|*{13}{c}}
\toprule
\multirow{2}{*}{\centering\arraybackslash \textbf{Methods}} &
\multirow{2}{*}{\centering\arraybackslash \makecell{\textbf{Batch}\\\textbf{Size}}} &
\multicolumn{13}{c}{\raisebox{0.4ex}{\textbf{Learning Rate}}} \\ 
\cline{3-15}
& &
\textbf{2.00e-06} & \textbf{3.56e-06} & \textbf{6.32e-06} & \textbf{1.12e-05} & \textbf{2.00e-05} & \textbf{3.56e-05} & \textbf{6.32e-05} &
\textbf{1.12e-04} & \textbf{2.00e-04} & \textbf{3.56e-04} & \textbf{6.32e-04} &
\textbf{1.12e-03} & \textbf{2.00e-03} \\
\midrule
\textbf{LoRA}
& \textbf{64} \small
& 21.48$_{\pm 0.53}$
& 24.52$_{\pm 0.44}$
& 31.94$_{\pm 0.58}$
& 43.07$_{\pm 1.16}$
& 48.37$_{\pm 0.19}$
& 48.81$_{\pm 0.14}$
& 49.27$_{\pm 0.34}$
& 49.46$_{\pm 0.56}$
& \textbf{49.60}$_{\pm 0.18}$
& 48.95$_{\pm 0.20}$
& 47.08$_{\pm 0.22}$
& 40.76$_{\pm 0.96}$
& 1.27$_{\pm 0.15}$ \\
\cline{1-15}

\textbf{DoRA}
& \textbf{64} \small
& 36.84$_{\pm 0.86}$
& 38.19$_{\pm 1.74}$
& 39.29$_{\pm 0.36}$
& 44.72$_{\pm 0.96}$
& 48.09$_{\pm 0.74}$
& 49.01$_{\pm 0.44}$
& 49.25$_{\pm 0.33}$
& \textbf{49.45}$_{\pm 0.25}$
& 49.33$_{\pm 0.45}$
& 49.32$_{\pm 0.71}$
& 46.92$_{\pm 0.28}$
& 40.24$_{\pm 0.62}$
& 10.14$_{\pm 15.64}$ \\
\cline{1-15}

\textbf{Init[AB]}
& \textbf{64} \small
& 33.40$_{\pm 2.16}$
& 35.81$_{\pm 1.72}$
& 41.20$_{\pm 1.83}$
& 47.27$_{\pm 0.89}$
& 49.27$_{\pm 0.13}$
& 49.02$_{\pm 0.11}$
& 48.81$_{\pm 0.17}$
& \textbf{49.29}$_{\pm 0.23}$
& 48.51$_{\pm 0.44}$
& 47.37$_{\pm 0.39}$
& 44.81$_{\pm 0.34}$
& 39.41$_{\pm 0.49}$
& 0.93$_{\pm 0.41}$  \\
\cline{1-15}

\textbf{MiLoRA}
& \textbf{64} \small
& 33.62$_{\pm 0.39}$
& 38.35$_{\pm 0.46}$
& 43.37$_{\pm 0.41}$
& 48.30$_{\pm 0.27}$
& 49.08$_{\pm 0.23}$
& 48.74$_{\pm 0.32}$
& \textbf{49.17}$_{\pm 0.38}$
& 49.08$_{\pm 0.14}$
& 48.22$_{\pm 0.20}$
& 46.57$_{\pm 0.73}$
& 43.91$_{\pm 0.58}$
& 38.70$_{\pm 0.34}$
& 9.48$_{\pm 14.40}$ \\ 
\cline{1-15}

\textbf{PiSSA}
& \textbf{64} \small
& 38.13$_{\pm 0.84}$
& 44.51$_{\pm 0.26}$
& 48.11$_{\pm 0.14}$
& 48.77$_{\pm 0.12}$
& \textbf{49.43}$_{\pm 0.19}$
& 49.09$_{\pm 0.16}$
& 48.44$_{\pm 0.12}$
& 47.10$_{\pm 0.32}$
& 43.84$_{\pm 0.28}$
& 39.66$_{\pm 0.36}$
& 34.37$_{\pm 0.39}$
& 27.18$_{\pm 1.29}$
& 17.35$_{\pm 0.57}$ \\
\bottomrule
\end{tabular}
}
\caption{Performance of \textbf{Qwen3-0.6B} fine-tuned on \textbf{mathematical reasoning} tasks with \textbf{rank = 128}.}
\label{tab:qwen_math_r128}
\end{table*}

\FloatBarrier

\clearpage
\FloatBarrier

\subsection{Gemma-3-1B}\label{sec:table_gemma}
\subsubsection{Mathematical Reasoning}
\begin{table*}[h]
\centering
\renewcommand{\arraystretch}{1.75} 
\setlength{\tabcolsep}{0.85pt} 
\setlength{\aboverulesep}{0pt}
\setlength{\belowrulesep}{0pt}
\adjustbox{max width=\textwidth}{%
\begin{tabular}{l|c|*{12}{c}}
\toprule
\multirow{2}{*}{\centering\arraybackslash \textbf{Methods}} & 
\multirow{2}{*}{\centering\arraybackslash \makecell{\textbf{Batch}\\\textbf{Size}}} & 
\multicolumn{12}{c}{\raisebox{0.4ex}{\textbf{Learning Rate}}} \\
\cline{3-14}
& & 
\textbf{1.12e-5} & \textbf{2.00e-5} & \textbf{3.56e-5} & \textbf{6.32e-5} & 
\textbf{1.12e-4} & \textbf{2.00e-4} & \textbf{3.56e-4} & \textbf{6.32e-4} & 
\textbf{1.12e-3} & \textbf{2.00e-3} & \textbf{3.56e-3} & \textbf{6.32e-3} \\
\midrule
\multirow{3}{*}{\textbf{LoRA}}
& \textbf{16} \small
& 9.78$_{\pm 0.36}$
& 11.16$_{\pm 0.28}$
& 13.58$_{\pm 0.18}$
& 15.48$_{\pm 0.15}$
& 18.43$_{\pm 0.14}$
& \textbf{20.00}$_{\pm 0.26}$
& 19.93$_{\pm 0.65}$
& 17.99$_{\pm 0.55}$
& 11.71$_{\pm 0.49}$
& 1.52$_{\pm 0.19}$
& 1.27$_{\pm 0.59}$
& 1.07$_{\pm 0.27}$ \\
\normalsize & \textbf{64} \small
& 6.88$_{\pm 0.04}$
& 9.12$_{\pm 0.39}$
& 10.79$_{\pm 0.37}$
& 13.23$_{\pm 0.25}$
& 15.65$_{\pm 0.57}$
& 17.54$_{\pm 0.29}$
& 19.73$_{\pm 0.16}$
& \textbf{20.46}$_{\pm 0.79}$
& 19.83$_{\pm 0.91}$
& 13.33$_{\pm 0.81}$
& 1.48$_{\pm 0.48}$
& 0.00$_{\pm 0.00}$ \\
\normalsize & \textbf{128} \small
& 5.70$_{\pm 0.34}$
& 6.95$_{\pm 0.23}$
& 9.41$_{\pm 0.44}$
& 11.43$_{\pm 0.40}$
& 13.68$_{\pm 0.77}$
& 15.92$_{\pm 0.45}$
& 18.58$_{\pm 0.44}$
& 19.60$_{\pm 0.09}$
& \textbf{20.32}$_{\pm 0.28}$
& 16.95$_{\pm 2.70}$
& 0.09$_{\pm 0.16}$
& 0.00$_{\pm 0.00}$ \\
\cline{1-14}
\multirow{3}{*}{\textbf{DoRA}}
\normalsize & \textbf{16} \small
& 9.89$_{\pm 0.24}$
& 11.16$_{\pm 0.51}$
& 13.84$_{\pm 0.41}$
& 15.61$_{\pm 0.11}$
& 18.21$_{\pm 0.45}$
& 20.11$_{\pm 0.26}$
& \textbf{20.96}$_{\pm 0.57}$
& 18.34$_{\pm 0.20}$
& 11.90$_{\pm 0.29}$
& 4.89$_{\pm 0.99}$
& 0.93$_{\pm 0.12}$
& 1.16$_{\pm 0.15}$ \\
\normalsize & \textbf{64} \small
& 6.72$_{\pm 0.09}$
& 9.19$_{\pm 0.19}$
& 10.53$_{\pm 0.20}$
& 13.45$_{\pm 0.31}$
& 15.72$_{\pm 0.32}$
& 17.66$_{\pm 0.20}$
& 19.96$_{\pm 0.05}$
& \textbf{20.82}$_{\pm 0.32}$
& 19.87$_{\pm 0.91}$
& 13.53$_{\pm 1.64}$
& 1.52$_{\pm 0.45}$
& 0.34$_{\pm 0.23}$ \\
\normalsize & \textbf{128} \small
& 5.55$_{\pm 0.11}$
& 7.21$_{\pm 0.18}$
& 9.72$_{\pm 0.17}$
& 11.58$_{\pm 0.25}$
& 13.98$_{\pm 0.33}$
& 16.19$_{\pm 0.46}$
& 18.25$_{\pm 0.23}$
& 19.67$_{\pm 0.71}$
& \textbf{20.33}$_{\pm 0.64}$
& 12.86$_{\pm 10.03}$
& 0.13$_{\pm 0.23}$
& 0.02$_{\pm 0.03}$ \\
\cline{1-14}
\multirow{3}{*}{\textbf{Init[AB]}}
\normalsize & \textbf{16} \small
& 9.73$_{\pm 0.35}$
& 12.10$_{\pm 0.14}$
& 14.41$_{\pm 0.49}$
& 16.73$_{\pm 0.37}$
& 18.38$_{\pm 0.53}$
& 20.39$_{\pm 0.38}$
& \textbf{20.55}$_{\pm 0.40}$
& 18.34$_{\pm 0.48}$
& 11.94$_{\pm 0.31}$
& 1.48$_{\pm 0.24}$
& 1.16$_{\pm 0.31}$
& 1.45$_{\pm 0.17}$ \\
\normalsize & \textbf{64} \small
& 6.51$_{\pm 0.22}$
& 9.15$_{\pm 0.12}$
& 11.28$_{\pm 0.20}$
& 13.20$_{\pm 0.24}$
& 15.88$_{\pm 0.39}$
& 17.89$_{\pm 0.30}$
& 20.08$_{\pm 0.26}$
& \textbf{20.98}$_{\pm 0.33}$
& 19.31$_{\pm 0.75}$
& 13.97$_{\pm 0.03}$
& 2.74$_{\pm 3.83}$
& 0.07$_{\pm 0.12}$ \\
\normalsize & \textbf{128} \small
& 6.06$_{\pm 0.35}$
& 7.05$_{\pm 0.33}$
& 9.53$_{\pm 0.22}$
& 11.81$_{\pm 0.08}$
& 13.98$_{\pm 0.79}$
& 16.46$_{\pm 0.39}$
& 18.36$_{\pm 0.21}$
& 20.37$_{\pm 0.39}$
& \textbf{20.66}$_{\pm 0.39}$
& 17.85$_{\pm 0.84}$
& 4.40$_{\pm 7.46}$
& 0.00$_{\pm 0.00}$ \\
\cline{1-14}

\multirow{3}{*}{\textbf{MiLoRA}}
\normalsize & \textbf{16} \small
& 12.44$_{\pm 0.07}$
& 13.77$_{\pm 0.25}$
& 16.28$_{\pm 0.24}$
& 18.45$_{\pm 0.47}$
& 20.04$_{\pm 0.19}$
& \textbf{20.63}$_{\pm 0.67}$
& 19.40$_{\pm 0.80}$
& 15.72$_{\pm 0.49}$
& 10.22$_{\pm 0.42}$
& 2.03$_{\pm 0.95}$
& 1.35$_{\pm 0.43}$
& 1.56$_{\pm 0.65}$ \\
\normalsize & \textbf{64} \small
& 8.82$_{\pm 0.40}$
& 11.25$_{\pm 0.20}$
& 13.16$_{\pm 0.11}$
& 15.54$_{\pm 0.29}$
& 17.43$_{\pm 0.24}$
& 19.56$_{\pm 0.33}$
& \textbf{20.03}$_{\pm 0.59}$
& 19.60$_{\pm 0.78}$
& 17.93$_{\pm 0.90}$
& 13.65$_{\pm 0.07}$
& 4.97$_{\pm 0.40}$
& 0.00$_{\pm 0.00}$ \\
\normalsize & \textbf{128} \small
& 7.32$_{\pm 0.33}$
& 9.57$_{\pm 0.24}$
& 11.76$_{\pm 0.33}$
& 13.54$_{\pm 0.12}$
& 16.02$_{\pm 0.16}$
& 18.39$_{\pm 0.26}$
& 19.70$_{\pm 0.34}$
& \textbf{19.99}$_{\pm 0.66}$
& 19.53$_{\pm 0.47}$
& 16.83$_{\pm 0.73}$
& 7.45$_{\pm 1.00}$
& 0.57$_{\pm 0.81}$ \\
\cline{1-14}
\multirow{3}{*}{\textbf{PiSSA}}
\normalsize & \textbf{16} \small
& 14.30$_{\pm 0.18}$
& 16.10$_{\pm 0.27}$
& 18.31$_{\pm 0.12}$
& 19.90$_{\pm 0.21}$
& \textbf{20.61}$_{\pm 0.28}$
& 19.09$_{\pm 0.20}$
& 16.10$_{\pm 0.64}$
& 13.25$_{\pm 0.55}$
& 8.41$_{\pm 0.13}$
& 4.67$_{\pm 0.29}$
& 2.50$_{\pm 1.27}$
& 0.96$_{\pm 0.15}$ \\
\normalsize & \textbf{64} \small
& 11.11$_{\pm 0.05}$
& 13.67$_{\pm 0.17}$
& 15.56$_{\pm 0.33}$
& 18.11$_{\pm 0.23}$
& 19.52$_{\pm 0.48}$
& \textbf{20.68}$_{\pm 0.77}$
& 20.59$_{\pm 0.32}$
& 19.11$_{\pm 0.86}$
& 15.53$_{\pm 0.37}$
& 9.57$_{\pm 0.72}$
& 5.78$_{\pm 0.37}$
& 0.33$_{\pm 0.46}$ \\
\normalsize & \textbf{128} \small
& 9.42$_{\pm 0.38}$
& 11.80$_{\pm 0.28}$
& 14.40$_{\pm 0.11}$
& 16.23$_{\pm 0.38}$
& 18.60$_{\pm 0.21}$
& 19.61$_{\pm 0.44}$
& \textbf{20.65}$_{\pm 0.44}$
& 19.21$_{\pm 1.15}$
& 16.91$_{\pm 0.19}$
& 13.87$_{\pm 0.97}$
& 6.28$_{\pm 0.49}$
& 1.19$_{\pm 0.36}$ \\
\bottomrule
\end{tabular}
}
\caption{Performance of \textbf{Gemma-3-1B} fine-tuned on \textbf{mathematical reasoning} tasks with \textbf{rank=128}.}
\label{tab:gemma_math_r128}
\end{table*}

\FloatBarrier

\subsection{Llama-2-7B}\label{sec:table_llama}
\subsubsection{Mathematical Reasoning}
\begin{table*}[h]
\centering
\footnotesize 
\renewcommand{\arraystretch}{1.75} 
\setlength{\tabcolsep}{1.85pt} 
\setlength{\aboverulesep}{0pt}
\setlength{\belowrulesep}{0pt}
\adjustbox{max width=0.85\textwidth}{%
\begin{tabular}{l|c|*{10}{c}}
\toprule
\multirow{2}{*}{\centering\arraybackslash \textbf{Methods}} &
\multirow{2}{*}{\centering\arraybackslash \makecell{\textbf{Batch}\\\textbf{Size}}} &
\multicolumn{10}{c}{\raisebox{0.4ex}{\textbf{Learning Rate}}} \\
\cline{3-12}
& &
\textbf{2.00e-05} & \textbf{3.56e-05} & \textbf{6.32e-05} & \textbf{1.12e-04} &
\textbf{2.00e-04} & \textbf{3.56e-04} & \textbf{6.32e-04} & \textbf{1.12e-03} & \textbf{2.00e-03} & \textbf{3.56e-03} \\
\midrule

\multirow{2}{*}{\textbf{LoRA}}
& \textbf{16} \small
& 23.16
& 24.55
& 26.05
& 28.73
& 30.70
& 32.18
& \textbf{32.94}
& 32.02 
& 27.71
& 0.00 \\

\normalsize & \textbf{128} \small
& 16.12
& 18.60
& 21.46
& 23.49
& 25.91
& 28.21
& 30.31
& 32.28 
& \textbf{32.78} 
& 1.97 \\
\cline{1-12}

\multirow{2}{*}{\textbf{DoRA}}
& \textbf{16} \small
& 22.74
& 24.34
& 26.20
& 28.44
& 30.62
& \textbf{33.20}
& 32.71
& 32.43 
& 1.59 
& 1.98 \\

\normalsize & \textbf{128} \small
& 16.07
& 18.98
& 21.57
& 23.70
& 26.16
& 28.54
& 30.20
& 32.80 
& \textbf{33.62} 
& 0.00  \\
\cline{1-12}

\multirow{2}{*}{\textbf{Init[AB]}}
& \textbf{16} \small
& 20.89
& 23.36
& 27.08
& 29.25
& 31.24
& \textbf{33.30}
& 32.78
& 31.34 
& 27.38 
& 0.04 \\

\normalsize & \textbf{128} \small
& 15.79 
& 17.64
& 20.17
& 22.96
& 25.42
& 28.03
& 30.45
& 32.32
& \textbf{32.96} 
& 31.08 \\
\cline{1-12}

\multirow{2}{*}{\textbf{MiLoRA}}
& \textbf{16} \small
& 21.12
& 23.45
& 25.61
& 28.38
& 30.59
& 32.49
& \textbf{33.22}
& 32.46 
& 27.56 
& 0.00 \\

\normalsize & \textbf{128} \small
& 15.72
& 18.51
& 20.70
& 22.58
& 25.32
& 26.76
& 29.87
& 31.48 
& \textbf{33.55} 
& 0.26 \\
\cline{1-12}

\multirow{2}{*}{\textbf{PiSSA}}
& \textbf{16} \small
& 22.66
& 26.30
& 28.20
& 30.12
& \textbf{31.91}
& 31.62
& 30.57
& 28.76 
& 0.84 
& 0.92 \\

\normalsize & \textbf{128} \small
& 18.94
& 21.60
& 22.80
& 26.23
& 28.14
& 30.61
& 31.64
& \textbf{31.86} 
& 31.53 
& 0.48 \\

\bottomrule
\end{tabular}
}
\caption{Performance of \textbf{Llama-2-7B} fine-tuned on \textbf{mathematical reasoning} with \textbf{rank=8}. 
}
\label{tab:llama_math_r8}
\end{table*}

\begin{table*}[h]
\centering
\footnotesize 
\renewcommand{\arraystretch}{1.75} 
\setlength{\tabcolsep}{1.85pt} 
\setlength{\aboverulesep}{0pt}
\setlength{\belowrulesep}{0pt}
\adjustbox{max width=0.78\textwidth}{%
\begin{tabular}{l|c|*{9}{c}}
\toprule
\multirow{2}{*}{\centering\arraybackslash \textbf{Methods}} &
\multirow{2}{*}{\centering\arraybackslash \makecell{\textbf{Batch}\\\textbf{Size}}} &
\multicolumn{9}{c}{\raisebox{0.4ex}{\textbf{Learning Rate}}} \\
\cline{3-11}
& &
\textbf{2.00e-05} & \textbf{3.56e-05} & \textbf{6.32e-05} & \textbf{1.12e-04} &
\textbf{2.00e-04} & \textbf{3.56e-04} & \textbf{6.32e-04} & \textbf{1.12e-03} & \textbf{2.00e-03} \\
\midrule

\multirow{2}{*}{\textbf{LoRA}}
& \textbf{16} \small
& 25.03
& 27.77
& 29.67
& 32.11
& 33.73
& 33.84
& \textbf{34.18}
& 27.01
& 1.31 \\

\normalsize & \textbf{128} \small
& 19.93
& 22.25 
& 24.08 
& 26.29 
& 29.13 
& 32.19 
& 33.24 
& \textbf{34.62 }
& 0.00 \\
\cline{1-11}

\multirow{2}{*}{\textbf{DoRA}}
& \textbf{16} \small
& 25.41
& 27.35
& 29.78
& 31.96
& 34.14
& \textbf{35.16}
& 33.26
& 28.92 
& 0.55 \\

\normalsize & \textbf{128} \small
& 20.60
& 22.41 
& 24.02
& 26.66
& 29.71
& 31.96 
& 33.41
& \textbf{34.25} 
& 0.00 \\
\cline{1-11}

\multirow{2}{*}{\textbf{Init[AB]}}
& \textbf{16} \small
& 24.06 
& 27.68 
& 28.71 
& 32.52 
& 34.10 
& \textbf{34.92}
& 34.12
& 27.16 
& 0.77 \\

\normalsize & \textbf{128} \small
& 17.83 
& 20.96 
& 23.15 
& 26.62 
& 28.34 
& 30.79 
& 33.38 
& \textbf{34.59}
& 0.97 \\

\cline{1-11}
\multirow{2}{*}{\textbf{MiLoRA}}
& \textbf{16} \small
& 24.53
& 27.23
& 29.23
& 31.44
& 33.97
& \textbf{34.85}
& 33.85
& 28.35
& 0.62 \\

\normalsize & \textbf{128} \small
& 18.23 
& 21.45 
& 23.65 
& 26.54 
& 27.97 
& 30.29 
& 32.39 
& \textbf{34.59}
& 0.00 \\
\cline{1-11}

\multirow{2}{*}{\textbf{PiSSA}}
& \textbf{16} \small
& 29.42
& 30.68
& 32.75
& \textbf{33.66}
& 33.62
& 32.66
& 29.30
& 18.36 
& 0.15 \\

\normalsize & \textbf{128} \small
& 23.98
& 27.02 
& 29.44 
& 30.01 
& 32.89 
& 33.42 
& \textbf{34.31} 
& 32.92 
& 0.65 \\
\bottomrule
\end{tabular}
}
\caption{Performance of \textbf{Llama-2-7B} fine-tuned on \textbf{mathematical reasoning} with \textbf{rank=32}.}
\label{tab:llama_math_r32}
\end{table*}

\begin{table*}[h]
\centering
\footnotesize 
\renewcommand{\arraystretch}{1.75} 
\setlength{\tabcolsep}{1.85pt} 
\setlength{\aboverulesep}{0pt}
\setlength{\belowrulesep}{0pt}
\adjustbox{max width=0.78\textwidth}{%
\begin{tabular}{l|c|*{9}{c}}
\toprule
\multirow{2}{*}{\centering\arraybackslash \textbf{Methods}} &
\multirow{2}{*}{\centering\arraybackslash \makecell{\textbf{Batch}\\\textbf{Size}}} &
\multicolumn{9}{c}{\raisebox{0.4ex}{\textbf{Learning Rate}}} \\
\cline{3-11}
& &
\textbf{2.00e-05} & \textbf{3.56e-05} & \textbf{6.32e-05} & \textbf{1.12e-04} &
\textbf{2.00e-04} & \textbf{3.56e-04} & \textbf{6.32e-04} & \textbf{1.12e-03} & \textbf{2.00e-03} \\
\midrule

\multirow{2}{*}{\textbf{LoRA}}
& \textbf{16} \small
& 29.21
& 31.30
& 33.25
& 35.45
& \textbf{35.91}
& 35.10
& 27.41
& 0.97 
& 0.00 \\

\normalsize & \textbf{128} \small
& 22.69
& 24.95
& 27.79
& 30.74
& 32.62
& 34.85
& \textbf{35.66}
& 0.00
& 0.00 \\
\cline{1-11}

\multirow{2}{*}{\textbf{DoRA}}
& \textbf{16} \small
& 29.43
& 30.75
& 33.14
& 35.73
& \textbf{36.41}
& 34.54
& 1.28
& 0.94 
& 0.79 \\

\normalsize & \textbf{128} \small
& 23.27
& 25.63
& 27.87
& 30.11
& 33.00
& 35.10
& \textbf{35.57}
& 0.38 
& 0.00 \\
\cline{1-11}

\multirow{2}{*}{\textbf{Init[AB]}}
& \textbf{16} \small
& 29.04
& 31.52 
& 31.96 
& 34.81 
& \textbf{36.72} 
& 35.41 
& 27.98 
& 1.54 
& 0.00 \\

\normalsize & \textbf{128} \small
& 22.01 
& 25.03 
& 28.11
& 30.47
& 31.80
& 34.78
& \textbf{35.57}
& 34.45
& 0.00 \\
\cline{1-11}

\multirow{2}{*}{\textbf{MiLoRA}}
& \textbf{16} \small
& 28.23
& 31.11
& 33.42
& 35.22
& \textbf{36.02}
& 34.71
& 28.03
& 0.38 
& 0.00 \\

\normalsize & \textbf{128} \small
& 22.14
& 24.84
& 27.88
& 30.33
& 31.29
& 33.67
& \textbf{35.23}
& 0.00 
& 0.00 \\
\cline{1-11}

\multirow{2}{*}{\textbf{PiSSA}}
& \textbf{16} \small
& 33.35
& 35.03
& 35.27
& \textbf{35.83}
& 32.89
& 27.90
& 16.75
& 1.45 
& 0.00 \\

\normalsize & \textbf{128} \small
& 29.84
& 31.64
& 33.44
& 34.45
& \textbf{35.31}
& 34.99
& 31.59
& 27.83 
& 0.00 \\

\bottomrule
\end{tabular}
}
\caption{Performance of \textbf{Llama-2-7B} fine-tuned on \textbf{mathematical reasoning} with \textbf{rank=128}.}
\label{tab:llama_math_r128}
\end{table*}

\FloatBarrier
\subsubsection{Code Generation}
\begin{table*}[h]
\centering
\footnotesize 
\renewcommand{\arraystretch}{1.75} 
\setlength{\tabcolsep}{1.85pt} 
\setlength{\aboverulesep}{0pt}
\setlength{\belowrulesep}{0pt}
\adjustbox{max width=0.85\textwidth}{%
\begin{tabular}{l|c|*{10}{c}}
\toprule
\multirow{2}{*}{\centering\arraybackslash \textbf{Methods}} &
\multirow{2}{*}{\centering\arraybackslash \makecell{\textbf{Batch}\\\textbf{Size}}} &
\multicolumn{10}{c}{\raisebox{0.4ex}{\textbf{Learning Rate}}} \\
\cline{3-12}
& &
\textbf{2.00e-05} & \textbf{3.56e-05} & \textbf{6.32e-05} & \textbf{1.12e-04} &
\textbf{2.00e-04} & \textbf{3.56e-04} & \textbf{6.32e-04} & \textbf{1.12e-03} & \textbf{2.00e-03} & \textbf{3.56e-03} \\
\midrule

\multirow{2}{*}{\textbf{LoRA}}
& \textbf{16} \small
& 27.20
& 29.40
& 27.55
& 30.00
& 31.40
& 33.25
& \textbf{36.10}
& 32.65 
& 32.30
& 0.00 \\

\normalsize & \textbf{128} \small
& 24.50
& 25.60
& 27.05
& 28.15
& 29.30
& 30.90
& 30.45
& \textbf{35.35} 
& 33.35 
& 0.00 \\
\cline{1-12}
\multirow{2}{*}{\textbf{DoRA}}
& \textbf{16} \small
& 28.05
& 28.45
& 28.90
& 31.15
& 31.80
& 33.15
& \textbf{33.80}
& 32.00 
& 29.50 
& 0.00 \\

\normalsize & \textbf{128} \small
& 24.65
& 26.95
& 27.25
& 29.65
& 29.15
& 30.15
& 33.15
& 33.15 
& \textbf{33.35} 
& 32.55 \\

\cline{1-12}

\multirow{2}{*}{\textbf{Init[AB]}}
& \textbf{16} \small
& 26.25 
& 27.30 
& 30.30 
& 29.90 
& 32.90 
& 34.15 
& \textbf{35.60}
& 32.70
& 31.20 
& 0.00 \\

\normalsize & \textbf{128} \small
& 23.00
& 25.50
& 25.55
& 29.15
& 30.50
& 32.05
& 33.55
& \textbf{34.20}
& 33.25 
& 0.00 \\
\cline{1-12}

\multirow{2}{*}{\textbf{MiLoRA}}
& \textbf{16} \small
& 26.80
& 28.60
& 27.55
& 28.75
& 31.60
& \textbf{33.95}
& 32.25
& 33.30 
& 29.85 
& 0.00 \\

\normalsize & \textbf{128} \small
& 22.50
& 25.50
& 26.40
& 27.20
& 29.55
& 29.30
& 32.35
& 32.40
& \textbf{33.35}
& 0.00 \\
\cline{1-12}

\multirow{2}{*}{\textbf{PiSSA}}
& \textbf{16} \small
& 29.05
& 29.90
& 32.55
& 31.90
& \textbf{34.05}
& 30.75
& 29.85
& 29.60 
& 24.35
& 0.00 \\

\normalsize & \textbf{128} \small
& 27.45
& 27.70
& 30.15
& 30.45
& 31.45
& 30.10
& \textbf{33.55}
& 31.75 
& 30.75 
& 29.35 \\

\bottomrule
\end{tabular}
}
\caption{Performance of \textbf{Llama-2-7B} fine-tuned on \textbf{code generation} with \textbf{rank=8}.
}
\label{tab:llama_code_r8}
\end{table*}

\begin{table*}[h]
\centering
\footnotesize 
\renewcommand{\arraystretch}{1.75} 
\setlength{\tabcolsep}{1.85pt} 
\setlength{\aboverulesep}{0pt}
\setlength{\belowrulesep}{0pt}
\adjustbox{max width=0.78\textwidth}{
\begin{tabular}{l|c|*{9}{c}}
\toprule
\multirow{2}{*}{\centering\arraybackslash \textbf{Methods}} &
\multirow{2}{*}{\centering\arraybackslash \makecell{\textbf{Batch}\\\textbf{Size}}} &
\multicolumn{9}{c}{\raisebox{0.4ex}{\textbf{Learning Rate}}} \\
\cline{3-11}
& &
\textbf{2.00e-05} & \textbf{3.56e-05} & \textbf{6.32e-05} & \textbf{1.12e-04} &
\textbf{2.00e-04} & \textbf{3.56e-04} & \textbf{6.32e-04} & \textbf{1.12e-03} & \textbf{2.00e-03} \\
\midrule

\multirow{2}{*}{\textbf{LoRA}}
& \textbf{16} \small
& 28.30
& 29.10
& 30.55
& 35.30
& 37.05
& \textbf{37.65}
& 36.85
& 28.75
& 0.00 \\

\normalsize & \textbf{128} \small
& 27.60
& 27.25
& 29.25
& 29.30
& 30.85
& 34.55
& 35.45
& \textbf{35.65}
& 32.20 \\
\cline{1-11}
\multirow{2}{*}{\textbf{DoRA}}
& \textbf{16} \small
& 29.00
& 29.10
& 30.30
& 34.75
& 34.20
& 36.40
& \textbf{37.40}
& 29.35
& 0.00 \\

\normalsize & \textbf{128} \small
& 26.75
& 27.15
& 29.00
& 30.30
& 31.20
& 35.25
& \textbf{35.45}
& 35.15
& 0.00 \\
\cline{1-11}

\multirow{2}{*}{\textbf{Init[AB]}}
& \textbf{16} \small
& 27.75
& 29.70
& 31.10
& 34.05
& \textbf{35.40}
& 35.20
& 33.05
& 31.25
& 0.00 \\

\normalsize & \textbf{128} \small
& 26.10
& 27.00
& 27.75
& 29.00
& 31.60
& 35.70
& \textbf{37.00}
& 36.70
& 0.00 \\
\cline{1-11}
\multirow{2}{*}{\textbf{MiLoRA}}
& \textbf{16} \small
& 27.15
& 28.55
& 30.55
& 32.75
& 34.10
& 35.90
& \textbf{36.10}
& 27.40
& 0.00 \\

\normalsize & \textbf{128} \small
& 25.70
& 26.50
& 26.90
& 28.90
& 31.05
& 31.40
& 33.30
& \textbf{34.75}
& 0.00 \\
\cline{1-11}
\multirow{2}{*}{\textbf{PiSSA}}
& \textbf{16} \small
& 32.10
& 33.00
& 34.00
& 32.10
& \textbf{34.45}
& 32.30
& 27.75
& 18.55
& 0.00 \\

\normalsize & \textbf{128} \small
& 29.80
& 30.05
& 31.75
& 32.10
& 33.55
& \textbf{35.00}
& 33.20
& 32.60
& 0.00 \\
\bottomrule
\end{tabular}
}
\caption{Performance of \textbf{Llama-2-7B} fine-tuned on \textbf{code generation} tasks with \textbf{rank = 32}.}
\label{tab:llama_code_r32}
\end{table*}

\begin{table*}[h]
\centering
\footnotesize 
\renewcommand{\arraystretch}{1.75} 
\setlength{\tabcolsep}{1.85pt} 
\setlength{\aboverulesep}{0pt}
\setlength{\belowrulesep}{0pt}
\adjustbox{max width=0.78\textwidth}{%
\begin{tabular}{l|c|*{9}{c}}
\toprule
\multirow{2}{*}{\centering\arraybackslash \textbf{Methods}} &
\multirow{2}{*}{\centering\arraybackslash \makecell{\textbf{Batch}\\\textbf{Size}}} &
\multicolumn{9}{c}{\raisebox{0.4ex}{\textbf{Learning Rate}}} \\
\cline{3-11}
& &
\textbf{2.00e-05} & \textbf{3.56e-05} & \textbf{6.32e-05} & \textbf{1.12e-04} &
\textbf{2.00e-04} & \textbf{3.56e-04} & \textbf{6.32e-04} & \textbf{1.12e-03} & \textbf{2.00e-03} \\
\midrule

\multirow{2}{*}{\textbf{LoRA}}
& \textbf{16} \small
& 30.72
& 31.87
& 34.23
& \textbf{37.55}
& 37.40
& 36.68
& 29.20
& 0.00 
& 0.00 \\

\normalsize & \textbf{128} \small
& 29.37
& 29.82
& 31.18
& 33.40
& 35.70
& 36.48
& \textbf{36.68}
& 13.05 
& 0.00 \\
\cline{1-11}

\multirow{2}{*}{\textbf{DoRA}}
& \textbf{16} \small
& 30.93
& 32.50
& 34.72
& 36.70
& \textbf{38.30}
& 35.82
& 30.50
& 0.00 
& 0.00 \\

\normalsize & \textbf{128} \small
& 29.12
& 30.17
& 31.22
& 32.80
& 36.12
& \textbf{38.12}
& 37.50
& 0.00 
& 0.00 \\
\cline{1-11}

\multirow{2}{*}{\textbf{Init[AB]}}
& \textbf{16} \small
& 30.73
& 31.48
& 34.47
& 36.75
& \textbf{38.07}
& 35.82
& 30.72
& 0.00
& 0.00 \\

\normalsize & \textbf{128} \small
& 28.32
& 30.00
& 30.15
& 32.80
& 35.83
& 36.97
& \textbf{38.43}
& 0.00
& 0.00 \\
\cline{1-11}
\multirow{2}{*}{\textbf{MiLoRA}}
& \textbf{16} \small
& 29.72
& 32.18
& 33.55
& 36.53
& \textbf{37.08}
& 35.95
& 30.03
& 0.00 
& 0.00 \\

\normalsize & \textbf{128} \small
& 28.67
& 29.20
& 29.72
& 31.93
& 34.67
& \textbf{37.42}
& 37.07
& 0.00 
& 0.00 \\
\cline{1-11}
\multirow{2}{*}{\textbf{PiSSA}}
& \textbf{16} \small
& 35.47
& \textbf{37.35}
& 35.98
& 36.08
& 33.92
& 28.43
& 17.53
& 0.00 
& 0.00 \\

\normalsize & \textbf{128} \small
& 31.90
& 32.33
& 34.80
& 35.42
& \textbf{36.77}
& 36.67
& 34.50
& 26.90 
& 0.00\\

\bottomrule
\end{tabular}
}
\caption{Performance of \textbf{Llama-2-7B} fine-tuned on \textbf{code generation} tasks with \textbf{rank = 128}.}
\label{tab:llama_code_r128}
\end{table*}
\clearpage
\section{Example Model Responses}\label{sec:example_model_response}
We examine the responses of Gemma-3-1B fine-tuned on the mathematical reasoning task using LoRA and PiSSA ($r=128, B=16$) under various learning rates. Figure~\ref{fig:math_example_input} presents a randomly selected testing example from the MATH dataset, with the corresponding model responses organized in Table~\ref{tab:model_response}. In this example, it is easily observed that the two PEFT methods operate in distinct effective learning rate regimes. Specifically, LoRA produces correct reasoning paths within the range of $2 \times 10^{-4}$ to $6.32 \times 10^{-4}$, whereas PiSSA achieves accurate results in a slightly lower range of $6.32 \times 10^{-5}$ to $2 \times 10^{-4}$. Moreover, the LLM under PiSSA fine-tuning tends to continue responding reasonably at larger learning rates, as evident when $\eta = 2 \times 10^{-3}$ and $3.56 \times 10^{-3}$, where the model produces answers of ``6'' and ``4'', while under LoRA fine-tuning, the model diverges to output 
repetitive gibberish.
These qualitative findings correspond to our discussion in Sec.~\ref{sec:similar_perf_level}.

\begin{figure}[h]
    \centering
    \begin{minipage}{0.85\linewidth} 
        \small
        \begin{examplecard_2}[Selected MATH Testing Set Input]
            Below is an instruction that describes a task. Write a response that appropriately completes the request.\\\\ 
            \#\#\# Instruction: \\
            There are eight boys and six girls who are members of the trumpet section in the band. From the members of the trumpet section, a quintet is being formed. If the quintet must consist of three boys and two girls, how many quintets are possible? \\\\
            \#\#\# Response: Let's think step by step.
        \end{examplecard_2}
    \end{minipage}
    
    \vspace{-0.1cm}
    \caption{An example input from the MATH dataset used for testing.}
    \label{fig:math_example_input}
\end{figure}

\begin{longtable}{ 
    >{\centering\arraybackslash}m{1.6cm} | 
    p{0.41\textwidth} |                    
    p{0.41\textwidth}                      
}\caption{Qualitative comparison of LoRA and PiSSA reasoning paths across learning rates.}\label{tab:model_response} \\
    \hline
    \textbf{LR ($\eta$)} & 
    \multicolumn{1}{c|}{\textbf{LoRA Response}} & 
    \multicolumn{1}{c}{\textbf{PiSSA Response}} 
    \\
    \hline
    \endfirsthead
    
    \hline
    \textbf{LR ($\eta$)} & 
    \multicolumn{1}{c|}{\textbf{LoRA Response}} & 
    \multicolumn{1}{c}{\textbf{PiSSA Response}} 
    \\
    \hline
    \endhead

    \multirow{2}{*}{{\small $1.12 \times 10^{-5}$}} 
    & 
    \llmtext{
        We are given that there are 8 boys and 6 girls in the trumpet section.\\
        To form a quintet, we need to choose 3 boys and 2 girls.\\
        So, we need to choose 3 boys from the 8 boys and 2 girls from the 6 girls.\\
        This can be done in $\binom{8}{2} = \frac{8!}{2!6!} = \frac{8 \cdot 7}{2 \cdot 1} = \boxed{16}$ ways.\\
        The answer is: 16
    } 
    & 
    \llmtext{
        We have 8 boys and 6 girls, so we need to choose 3 boys from the 8 boys and 2 girls from the 6 girls.\\
        This can be done in $\binom{8}{3} \cdot \binom{6}{2} = \frac{8!}{3!3!} \cdot \frac{6!}{2!2!} = \frac{8 \cdot 7}{3 \cdot 2} = \boxed{14}$ ways.\\
        The answer is: 14
    } 
    \\* 
     & \hfill \xmark  & \hfill \xmark \\ 
    \hline

    \multirow{2}{*}{{\small $2.00 \times 10^{-5}$}} 
    & 
    \llmtext{
        We are given that there are 8 boys and 6 girls in the trumpet section.\\
        To form a quintet, we need to choose 3 boys and 2 girls.\\
        So, the number of quintets that can be formed is $\frac{8}{3} \times \frac{6}{2} = \frac{8 \times 6}{3 \times 2} = \frac{48}{6} = \boxed{8}$.\\
        The answer is: 8
    } 
    & 
    \llmtext{
        We have 8 boys and 6 girls, so we need to choose 3 boys from the 8 boys and 2 girls from the 6 girls.\\
        This can be done in $\binom{8}{3} \cdot \binom{6}{2} = \frac{8!}{3!3!} = \frac{8 \cdot 7}{3 \cdot 2} = 28$ ways.\\
        Therefore, there are $\boxed{28}$ quintets possible.\\
        The answer is: 28
    } 
    \\* 
     & \hfill \xmark & \hfill \xmark \\ 
    \hline
    \multirow{2}{*}{{\small $3.56 \times 10^{-5}$}} 
    & 
    \llmtext{
        We have 8 boys and 6 girls, so there are a total of 8 + 6 = 14 members in the trumpet section.\\
        To form a quintet, we need to choose 3 boys and 2 girls from the 14 members.\\
        The number of quintets that can be formed is the number of ways to choose 3 boys and 2 girls from a group of 14 members.\\
        Using the combination formula, this is $\binom{14}{3} = \frac{14!}{3!11!} = \frac{14 \times 13 \times 12}{3 \times 2 \times 1} = \boxed{105}$.\\
        The answer is: 105
    } 
    & 
    \llmtext{
        We have 8 boys and 6 girls in the trumpet section.\\
        To form a quintet, we need to choose 3 boys and 2 girls from the 8 boys and 6 girls.\\
        This can be done in $\binom{8}{3} \cdot \binom{6}{2} = \frac{8!}{3!3!} = \frac{8 \cdot 7}{3 \cdot 2} = 28$ ways.
        Therefore, there are $\boxed{28}$ possible quintets.\\
        The answer is: 28
    } 
    \\* 
     & \hfill \xmark & \hfill \xmark \\ 
    \hline
    \multirow{2}{*}{{\small $6.32 \times 10^{-5}$}} 
    & 
    \llmtext{
        We have 8 boys and 6 girls, so we need to choose 3 boys from the 8 boys and 2 girls from the 6 girls.\\
        This can be done in $\binom{8}{3} \cdot \binom{6}{2} = \frac{8!}{3!3!} \cdot \frac{6!}{2!4!} = \frac{8 \cdot 7 \cdot 6}{3 \cdot 2 \cdot 1} = \frac{216}{6} = \boxed{36}$.\\
        The answer is: 36
    } 
    & 
    \llmtext{
        We need to choose 3 boys from the 8 members of the trumpet section, and 2 girls from the 6 members of the trumpet section.\\
        The number of ways to choose 3 boys from 8 is $\binom{8}{3} = 56$.\\
        The number of ways to choose 2 girls from 6 is $\binom{6}{2} = 15$.\\
        Therefore, the total number of quintets is $56 \times 15 = \boxed{840}$.\\
        The answer is: 840     
    } 
    \\* 
     & \hfill \xmark & \hfill \cmark \\ 
    \hline
    \multirow{2}{*}{{\small $1.12 \times 10^{-4}$}} 
    & 
    \llmtext{
        We have 8 boys and 6 girls, so we need to choose 3 boys from the 8 boys and 2 girls from the 6 girls.\\
        This can be done in $\binom{8}{3} \cdot \binom{6}{2} = \frac{8!}{3!3!} \cdot \frac{6!}{2!2!} = \frac{8 \cdot 7 \cdot 6}{3 \cdot 2 \cdot 1} = 56$ ways.\\
        Therefore, there are $\boxed{56}$ possible quintets.\\
        The answer is: 56
    } 
    & 
    \llmtext{
        We need to choose 3 boys from the 8 available boys, and 2 girls from the 6 available girls.\\
        The number of ways to choose 3 boys from 8 is $\binom{8}{3} = 56$.\\
        The number of ways to choose 2 girls from 6 is $\binom{6}{2} = 15$.\\
        Therefore, the total number of quintets is $56 \times 15 = \boxed{840}$.\\
        The answer is: 840   
    } 
    \\* 
     & \hfill \xmark & \hfill \cmark \\ 
    \hline
    \multirow{2}{*}{{\small $2.00 \times 10^{-4}$}} 
    & 
    \llmtext{
        We need to choose 3 boys from the 8 available, and 2 girls from the 6 available.\\
        This can be done in $\binom{8}{3} = 56$ ways.\\
        Similarly, we need to choose 2 girls from the 6 available, and 3 boys from the 3 available.\\
        This can be done in $\binom{6}{2} = 15$ ways.\\
        Therefore, there are $56 \times 15 = \boxed{840}$ possible quintets.\\
        The answer is: 840
    } 
    & 
    \llmtext{
        First, we choose 3 boys from the 8 available boys. This can be done in $\binom{8}{3} = 56$ ways.\\
        Next, we choose 2 girls from the 6 available girls. This can be done in $\binom{6}{2} = 15$ ways.\\
        Therefore, the total number of quintets is $56 \times 15 = \boxed{840}$.\\
        The answer is: 840     
    } 
    \\* 
     & \hfill \cmark & \hfill \cmark \\ 
    \hline
    \multirow{2}{*}{{\small $3.56 \times 10^{-4}$}} 
    & 
    \llmtext{
        We need to choose 3 boys from the 8 available, and 2 girls from the 6 available.\\
        The number of ways to choose 3 boys from 8 is $\binom{8}{3} = 56$.\\
        The number of ways to choose 2 girls from 6 is $\binom{6}{2} = 15$.\\
        Therefore, the total number of quintets is $56 \times 15 = \boxed{840}$.\\
        The answer is: 840
    } 
    & 
    \llmtext{
        For the boys, we have 8 choices for the quintet.\\
        For the girls, we have 6 choices for the quintet.\\
        So, the total number of quintets is $8 \times 6 = \boxed{48}$.\\
        The answer is: 48
    } 
    \\* 
     & \hfill \cmark & \hfill \xmark \\ 
    \hline
    \multirow{2}{*}{{\small $6.32 \times 10^{-4}$}} 
    & 
    \llmtext{
       We have 8 boys and 6 girls, so we need to choose 3 boys and 2 girls.\\
        The number of ways to choose 3 boys from 8 is $\binom{8}{3} = 56$.\\
        The number of ways to choose 2 girls from 6 is $\binom{6}{2} = 15$.\\
        Therefore, the total number of quintets is $56 \times 15 = \boxed{840}$.\\
        The answer is: 840
    } 
    & 
    \llmtext{
        There are 8 boys and 6 girls, so there are a total of 8 + 6 = 14 members in the trumpet section.\\
        We need to choose 3 boys from the 8 boys, so there are 8 boys.\\
        We also need to choose 2 girls from the 6 girls, so there are 6 girls.\\
        Therefore, the total number of quintets is $8 + 3 + 6 = \boxed{17}$.\\
        The answer is: 17
    } 
    \\* 
     & \hfill \cmark & \hfill \xmark \\ 
    \hline
    \multirow{2}{*}{{\small $1.12 \times 10^{-3}$}} 
    & 
    \llmtext{
       From the members of the trumpet section, there are 8 boys and 6 girls.\\
       From the members of the quintet, we need to choose 3 boys from the 8 boys and 2 girls from the 6 girls.\\
       This can be done in $\binom{8}{3} = \frac{8!}{3!5!} = \frac{8 \cdot 7 \cdot 6}{3 \cdot 2 \cdot 1} = \boxed{35}$ ways.\\
       The answer is: 35
    } 
    & 
    \llmtext{
        There are 8 boys and 6 girls in the band, so there are a total of $8+6=14$ members in the band.\\
        We want to find the number of quintets that are possible.\\
        Since the quintet must consist of three boys and two girls, we need to choose 3 boys out of the 14 members.\\
        This can be done in $\binom{14}{3}=46$ ways.\\
        Therefore, there are $46-14=\boxed{32}$ possible quintets.\\
        The answer is: 32
    } 
    \\* 
     & \hfill \xmark & \hfill \xmark \\ 
    \hline
    \multirow{2}{*}{{\small $2.00 \times 10^{-3}$}} 
    & 
    \llmtext{
       Since the number of boys is x, we can set up the equation:\\
        Number of boys = x\\
        Number of boys with the number of boys with the two boys with the two boys with the two boys with the two boys...(repetitive gibberish)
    } 
    & 
    \llmtext{
        There are 8 boys and 6 girls, so there are a total of 8 + 6 = 14 people in the band.\\
        There are 6 girls and 2 girls, so there are a total of 6 + 2 = 8 people in the band.\\
        There are 8 quintet members in the band.\\
        Therefore, the number of quintets possible is 14 - 8 = 6.\\
        \#\#\#\# 6\\
        The answer is: 6
    } 
    \\* 
     & \hfill \xmark & \hfill \xmark \\ 
    \hline
    \multirow{2}{*}{{\small $3.56 \times 10^{-3}$}} 
    & 
    \llmtext{
       The total number of students is 1, so the number of people is 1, 1, 1, 1, 1, 1, 1, 1, 1, 1, 1, 1, 1, 1, 1, 1, 1, 1, 1, 1, 1, 1, 1, 1, 1, 1, 1, 1, 1...(repetitive gibberish)
    } 
    & 
    \llmtext{
        There are 8 boys and 6 girls, so there are a total of $8+6=14$ possible outcomes.
        Out of the 14 girls, 6 are girls, so there are $14-6=18$ possible outcomes.\\
        Therefore, there are $18-14=\boxed{4}$ possible outcomes.\\
        The answer is: 4
    } 
    \\* 
     & \hfill \xmark & \hfill \xmark \\ 
    \hline
    \multirow{2}{*}{{\small $6.32 \times 10^{-3}$}} 
    & 
    \llmtext{
       The number of the first circle is 12, 10 = 10.\\
        The total number of the number of the number of the number of the number of the number of the number of the first two sides of the first two sides.\\
        The area of the first two sides is 10 + 1 = 120 = 100\\
        The total number of the first two sides is 1000.\\
        The total number of the number of the first two sides is 1000, so the total number of the first 100000.\\
        The answer is 10000000000...(repetitive gibberish)
    } 
    & 
    \llmtext{
        We know that the total amount of the water is 1.\\
        We also know that the total number of eggs is 1.\\
        The total number of boys who has not the number of students is 1.\\
        We can set up the equation as follows:
        Number of students who has not the pool + Number of students who has not the number of students who in the pool...(repetitive gibberish)

    } 
    \\* 
     & \hfill \xmark & \hfill \xmark \\ 
    \hline
\end{longtable}

\clearpage
\section{Practical Heuristics for LoRA Hyperparameter Tuning}\label{sec:practical-heuristic}

Based on the extensive hyperparameter tuning experiments, we derive five practical heuristics for tuning LoRA-based methods.
While general strategies such as early stopping unpromising learning rate ranges and using sparse grid searches before zooming into promising ranges remain useful, 
our LoRA-specific guidelines provide additional heuristics on how learning rate interacts with batch size, LoRA rank, training duration, and loss curvature.
We summarize them below.

\textbf{I. Prioritize Learning Rate Tuning.} 
Based on the joint optimization of batch size and learning rate across diverse model--task combinations 
(Table~\ref{tab:gemma-main}, Appendix Tables~\ref{tab:qwen_math_r8},~\ref{tab:gemma_math_r128},~\ref{tab:llama_math_r8}--\ref{tab:llama_code_r128}), 
we suggest that, under limited computational resources, practitioners may prioritize learning rate tuning while fixing the batch size. 
Importantly, when the batch size is set too large, the best achievable performance under learning rate tuning may start to decay (Appendix Sec.~\ref{sec:addition-batch-size}). 
We therefore suggest using a small or medium batch size as the default choice.

\textbf{II. Mind Batch Size Scaling.} 
If additional resources are available and practitioners wish to explore different batch sizes, 
further performance gains are likely to be marginal once the learning rate has been properly tuned for each batch size
(as shown in Table~\ref{tab:gemma-main}). 
In practice, however, practitioners should still account for the scaling relationship between batch size and learning rate (discussed in Sec.~\ref{sec:similar_perf_level}), 
as it provides a useful initial guess for the learning rate when changing the batch size.

\textbf{III. Select Learning Rate based on Hessian.} 
As described in Sec.~\ref{sec:hessian-analysis} and Appendix Sec.~\ref{sec:additional_hessians_gemma_llama}, the maximum eigenvalue of the loss Hessian can serve as a useful indicator of a variant's relative operating learning rate range compared with vanilla LoRA. 
In Appendix Sec.~\ref{sec:detailed_lambda}, we further show that Hessian trends across different matrix types and Transformer layers are typically consistent, in the sense that they are generally either larger or smaller than those of vanilla LoRA. 
Critically, 
Hessian estimation for LoRA adapters of a single layer requires only around 10 minutes on a single RTX A6000. 
Hence, practitioners with sufficient resources may use Hessian analysis to guide 
initial learning rate tuning ranges
before conducting a large scale search.

We also note that, based on our broad experiments, a given variant typically exhibits a stable relationship in optimal learning rate relative to LoRA across different
model--task combinations,
in terms of being either higher or lower (e.g., as discussed in Sec.~\ref{sec:addition-model-task} for Figure~\ref{fig:main-evidence} and~\ref{fig:main-evidence-gemma-cs-r4}). 
Practitioners may therefore estimate the Hessian once and leverage the known learning-rate range relationships of specific variants across model--task combinations, 
without re-running the Lanczos algorithm every time they switch to a new setting.

\textbf{IV. Increase LoRA Ranks.} 
When sufficient effort has been invested in learning rate tuning at a given rank but the resulting downstream performance remains unsatisfactory, increasing the LoRA rank can be a reliable way to further improve performance, as shown in Figures~\ref{fig:perf_across_ranks} and~\ref{fig:perf_across_ranks_llama} for various
methods.
After switching to a higher rank, however, one should still perform learning rate tuning to elicit the best achievable performance at that rank.

To this end, 
Tables~\ref{tab:llama_math_r8}--\ref{tab:llama_math_r128} and Tables~\ref{tab:llama_code_r8}--\ref{tab:llama_code_r128} report results for $r=\{8,32,128\}$ on Llama-2-7B for math and code, respectively.
These results suggest that the optimal learning rate generally decreases as the rank increases.
With this, we also visualize vanilla LoRA's performance trends across learning rates for varying adapter ranks 
in Figure~\ref{fig:gemma-lr-rank-sample-a} in the next page,
with $r=4$ requiring a $5.6\times$ larger learning rate than $r=256$.
This observation can help practitioners conduct more efficient learning rate tuning across different ranks.

Although recommending larger ranks for better performance may seem straightforward, we highlight that this trend may not be observed in practice if the learning rate is not properly configured for each rank.
In fact, we find that many prior LoRA studies fail to demonstrate such a consistent performance improvement trend as the LoRA rank increases, partly because a fixed learning rate setting was applied (e.g., cf.~DoRA~\citep[Figure~5]{dora},
LoFT~\citep[Figure~3]{tastan2025loft},
GraLoRA~\citep[Table~2]{gralora}). 

\textbf{V. Prolong Training Duration.}
If practitioners have computational resources to further improve LoRA performance after increasing the LoRA rank with proper learning-rate tuning, we suggest prolonging the training duration as a final step. 
In particular, one can increase the training duration by using more training samples or training epochs. 
In both cases, we show in Appendix Sec.~\ref{sec:vary_train_duration} that, with proper learning rate tuning, various LoRA methods typically have the capacity to further improve their performance. 
Interestingly, we observe that the optimal learning-rate range shifts slightly downward as the amount of training increases, consistent with the tendency described in \emph{practical heuristic \textbf{IV}}. However, the magnitude of this decay is relatively modest, as shown in Figure~\ref{fig:gemma-lr-rank-sample-b}: the optimal learning rates for $N=5k$ and $N=395k$ differ by only $3.1 
\times$.
This suggests that, when extending training duration, practitioners may start from the learning rate range already identified under the shorter training setting, 
rather than restarting the search from scratch.


\begin{figure}[h]
    \centering
    \begin{subfigure}[b]{0.48\linewidth}
        \centering
        \includegraphics[width=\linewidth]{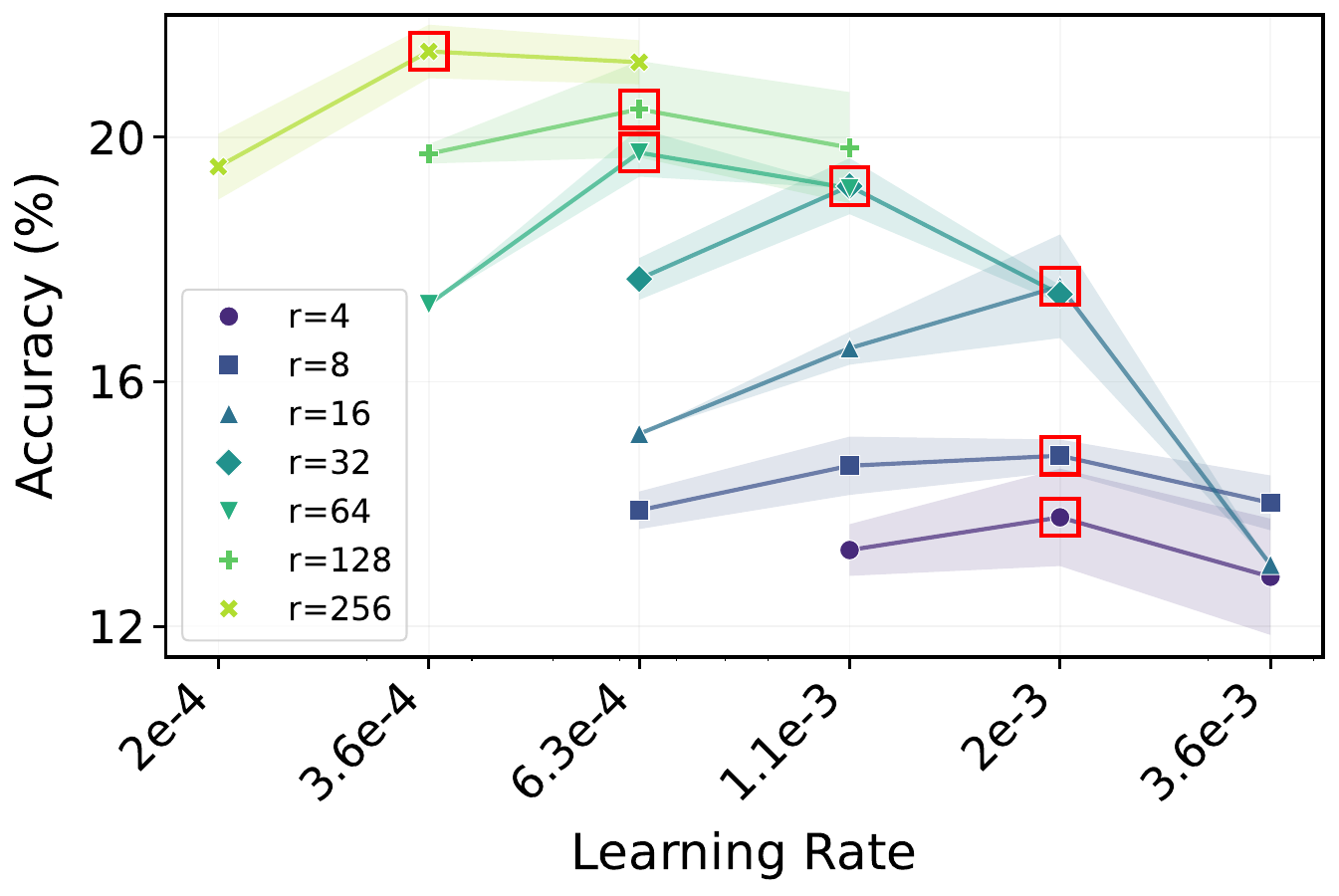}
        \caption{Varying LoRA Ranks}\label{fig:gemma-lr-rank-sample-a}
    \end{subfigure}
    \hfill
    \begin{subfigure}[b]{0.48\linewidth}
        \centering
        \includegraphics[width=\linewidth]{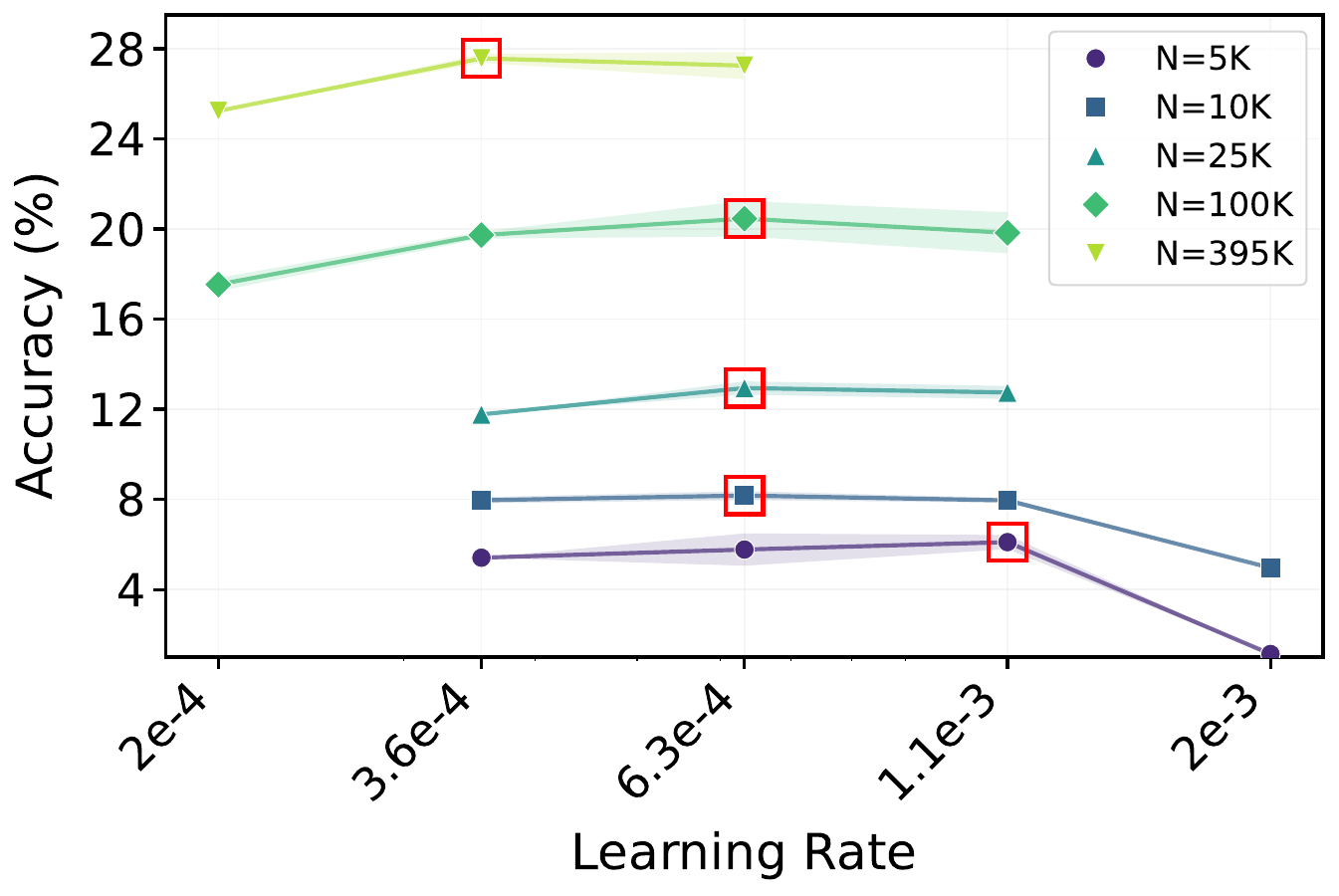}
        \caption{Varying Number of Training Samples}\label{fig:gemma-lr-rank-sample-b}
    \end{subfigure}
    \caption{Vanilla LoRA performance across learning rates with (a) varying ranks and (b) varying training-set sizes on mathematical reasoning with Gemma-3-1B ($B=64$). Note that panels (a) and (b) re-plot the data behind Figures~\ref{fig:performance_across_rank_math} and~\ref{fig:number_training_samples}, respectively, with the learning rate on the $x$-axis. Red boxes indicate the learning rate yielding the highest accuracy for each configuration.}
    \label{fig:gemma-lr-rank-sample}
\end{figure}
\clearpage
\section{Hessian Computation Details}\label{sec:hessian_implementation_details}

We select the Lanczos algorithm over the Power Iteration method for the presented eigenvalue problem, as the latter converges to eigenvalues in descending order of magnitude, whereas our focus is on probing the algebraically largest eigenvalue of the Hessian. Sec.~\ref{sec:lanczos_algo_implement} explains implementation details of the Lanczos algorithm, and Sec.~\ref{sec:additional_hessians_gemma_llama} and~\ref{sec:detailed_lambda} provide additional Hessian results for diverse model scales and matrix types, respectively.

\subsection{Lanczos Algorithm Implementation Details}\label{sec:lanczos_algo_implement}

Our implementation is built upon several Hessian-related frameworks, such as PyHessian\footnote{\url{https://github.com/amirgholami/PyHessian}}~\cite{yao2020pyhessian} and LLM-Hessian\footnote{\url{https://github.com/vectozavr/llm-hessian}}~\cite{ilin2025hessian}, with several modifications to suit our custom scenario. Algorithm~\ref{alg:lanczos} summarizes our implementation of the Lanczos Algorithm for estimating $\lambda_{\max}(\mathbf{H})$. We set the Lanczos iterations $m=100$ and tolerance $\epsilon=5 \times 10^{-3}$. 
At each Lanczos iteration step, the Hessian-Vector Product (HVP) is applied to calculate $\mathbf{H}\mathbf{q}_k$ without explicitly forming $\mathbf{H}$ (Algorithm~\ref{alg:hvp_calc}). 

We strictly ensure that the loss is calculated identically to that in supervised fine-tuning, rendering the resulting curvature information 
meaningful. In particular, we ensure that (1) the input prompt (i.e., instruction or question) tokens are masked out from the loss calculation, and (2) the loss is averaged over each token instead of each sentence\footnote{This is the default way of calculating loss during LLM supervised fine-tuning; see \url{https://github.com/huggingface/transformers/issues/34510}}. To ensure computational feasibility, a fixed subset of $N=500$ training samples from MetaMathQA is selected for loss calculation, and a batch size of $B=5$ is utilized. Figure~\ref{fig:hessian_esitimate_samples} validates that this sample size is sufficient for reliably estimating the Hessian of the downstream task. 
Due to the numerical instability of the Lanczos algorithm in finite precision arithmetic~\cite{cahill2000numerical}, we use Float32 precision for both the base model and adapters and incorporate re-orthogonalization steps~\cite{paige1970practical, golub1972lanczos}.

\begin{figure}[h]
    \centering
    \includegraphics[width=0.64\linewidth]{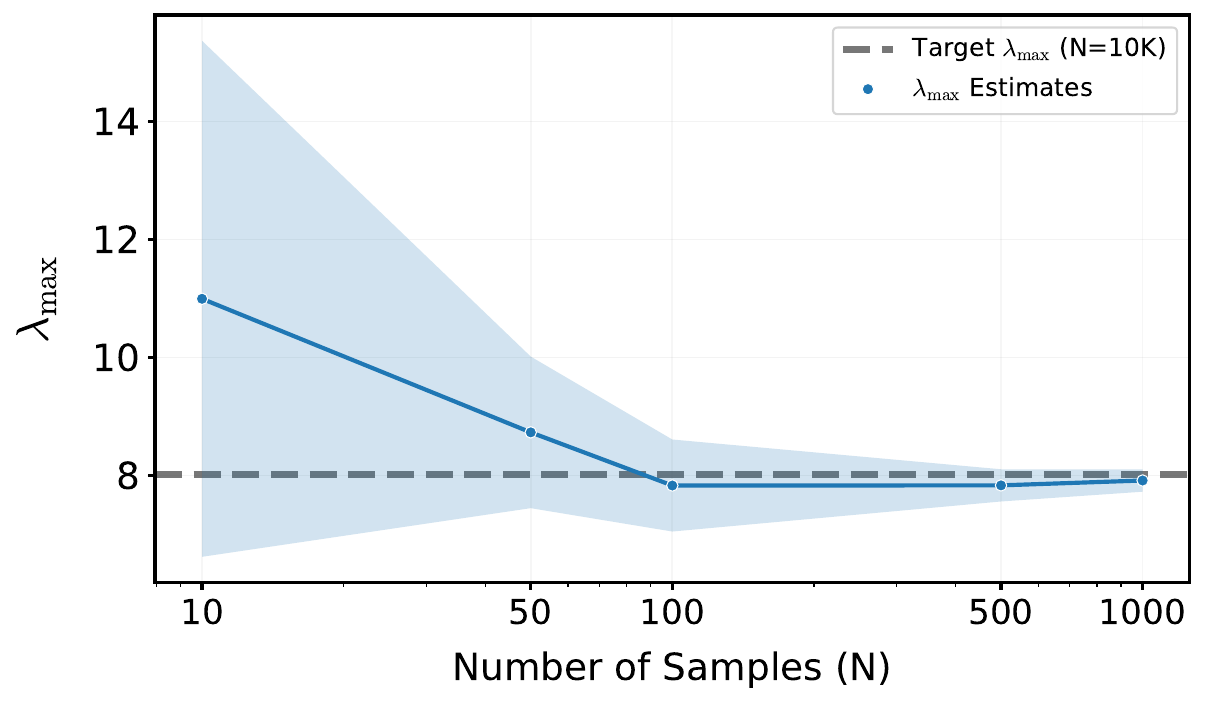}
    \caption{Approximately 500 training samples are sufficient for stable Hessian estimation. The figure reports the estimated $\lambda_{\max}$
    for PiSSA
    on the first Query projection matrix of Qwen3-0.6B ($r=128$). We track these estimates across varying sample sizes ($N$) from the MetaMathQA dataset, using $N=10k$ as the reference baseline.  
    Results represent the mean and standard deviation over 5 randomly selected subsets for each $N$.
    }    \label{fig:hessian_esitimate_samples}
\end{figure}

\begin{algorithm}[tb]
   \caption{Estimating Maximum Eigenvalues of Hessian by Lanczos Iterations}
   \label{alg:lanczos}
   \small
    \begin{algorithmic}
       \STATE {\bfseries Input:} LoRA parameters $\theta = \{B_0, A_0\}$, downstream dataset $\mathcal{D}$,  sample size $N$, iterations $m$, initial vector $\mathbf{b}$, tolerance $\epsilon$.
       \STATE {\bfseries Output:} Approximation of the maximum eigenvalue $\lambda_{\max}(\mathbf{H})$.
       \STATE {\bfseries Sampling:} Sample a subset $\mathcal{S}$ of size $N$ from $\mathcal{D}$.
       \STATE {\bfseries Initialization:}
       \STATE Set $\beta_0 = 0$, $\mathbf{q}_0 = \mathbf{0}$, $\lambda_{\text{prev}} = -\infty$.
       \STATE Normalize initial vector: $\mathbf{q}_1 = \mathbf{b} / \|\mathbf{b}\|_2$.
       \STATE {\bfseries Lanczos Iteration:}
       \FOR{$k = 1$ {\bfseries to} $m$}
       \STATE Compute Hessian-Vector Product:
       \STATE $\mathbf{v} = \text{HVP}(\theta, \mathcal{S}, \mathbf{q}_k)$ \hfill // See Algorithm \ref{alg:hvp_calc}
       \STATE Compute diagonal element of $T$:
       \STATE $\alpha_k = \mathbf{q}_k^\top \mathbf{v}$
       \STATE Orthogonalize (Gram-Schmidt):
       \STATE $\mathbf{v} = \mathbf{v} - \beta_{k-1}\mathbf{q}_{k-1} - \alpha_k\mathbf{q}_k$
       \STATE Compute off-diagonal element of $T$:
       \STATE $\beta_k = \|\mathbf{v}\|_2$
       \STATE {\bfseries Convergence Check:}
       \STATE Construct symmetric tridiagonal matrix $T_k \in \mathbb{R}^{k \times k}$ using $\alpha_{1:k}, \beta_{1:k-1}$.
       \STATE $\texttt{eig\_vals} \leftarrow \texttt{torch.linalg.eigvalsh}(T_k)$
       \STATE $\lambda_{\text{curr}} = \max(\texttt{eig\_vals})$
       \IF{$|\lambda_{\text{curr}} - \lambda_{\text{prev}}| < \epsilon$}
       \STATE {\bfseries Return} $\lambda_{\text{curr}}$
       \ENDIF
       \STATE $\lambda_{\text{prev}} \leftarrow \lambda_{\text{curr}}$
       \IF{$\beta_k \approx 0$}
       \STATE {\bfseries Return} $\lambda_{\text{curr}}$ \hfill // The Krylov subspace is invariant
       \ENDIF
       \STATE Normalize: $\mathbf{q}_{k+1} = \mathbf{v} / \beta_k$
       \ENDFOR
       \STATE {\bfseries Return} $\lambda_{\text{curr}}$
    \end{algorithmic}
\end{algorithm}

\begin{algorithm}[tb]
   \caption{Hessian-Vector Product (HVP) Calculation}
   \label{alg:hvp_calc}
   \small
    \begin{algorithmic}
       \STATE {\bfseries Input:} LoRA parameters $\theta = \{B_0, A_0\}$, vector $\mathbf{q}_k$, sample subset $\mathcal{S}$, batch size $B$.
       \STATE {\bfseries Output:} The Hessian-Vector product $\mathbf{H}\mathbf{q}_k$.
       \STATE {\bfseries Initialization:}
       \STATE Set accumulator $\mathbf{u} = \mathbf{0}$.
       \STATE Set total token counter $C_{\text{total}} = 0$.
       \STATE {\bfseries Batch Processing:}
       \FOR{each mini-batch $\mathcal{B}$ of size $B$ from $\mathcal{S}$}
       \STATE Count supervised tokens in batch: $c_{\mathcal{B}} = \text{CountTokens}(\mathcal{B})$.
       \STATE $C_{\text{total}} \leftarrow C_{\text{total}} + c_{\mathcal{B}}$.
       \STATE {\bfseries Forward Pass for Loss Calculation:}
       \STATE Compute sum of Cross-Entropy losses over all supervised tokens in $\mathcal{B}$:
       \STATE $\mathcal{L}_{\text{batch}}(\theta) = \sum_{(x, y) \in \mathcal{B}} \ell(f(x; \theta), y)$ \hfill // \texttt{torch.nn.CrossEntropyLoss(sum)}
       \STATE {\bfseries Double Backward for HVP:}
       \STATE $\mathbf{g} = \nabla_\theta \mathcal{L}_{\text{batch}}$
       \STATE $s = \mathbf{g}^\top \mathbf{q_k}$
       \STATE $\mathbf{h}_{\mathcal{B}} = \nabla_\theta s$ \hfill // Implemented via \texttt{torch.autograd.functional.vhp}
       \STATE {\bfseries Accumulate:}
       \STATE $\mathbf{u} \leftarrow \mathbf{u} + \mathbf{h}_{\mathcal{B}}$ \hfill // Summing up batch-wise contributions
       \ENDFOR
       \STATE {\bfseries Normalize:}
       \STATE $\mathbf{u} \leftarrow \mathbf{u} / C_{\text{total}}$ \hfill // Average over total supervised tokens
       \STATE {\bfseries Return} $\mathbf{u}$
    \end{algorithmic}
\end{algorithm}

\FloatBarrier


\FloatBarrier

\subsection{Hessian Results on Gemma and Llama}\label{sec:additional_hessians_gemma_llama}

As discussed earlier in Sec.~\ref{sec:hessian-analysis}, Figure~\ref{fig:main_hessian} presents the distributions of the top loss Hessian eigenvalues of LoRA variants relative to vanilla LoRA on Qwen3-0.6B, providing a theoretical explanation for the optimal learning rate trends observed in Figure~\ref{fig:main-evidence}. Analogously, Figure~\ref{fig:hessian_gemma_llama_math_r128_dist} below shows the corresponding distributions on Gemma-3-1B and Llama-2-7B. The associated learning rate tuning results for these two models on mathematical reasoning tasks under rank $r=128$ are reported earlier in Table~\ref{tab:gemma-main} and Figure~\ref{fig:llama-main-math}, respectively. 

\begin{figure*}[h]
    \centering
    \includegraphics[width=1.00\linewidth]{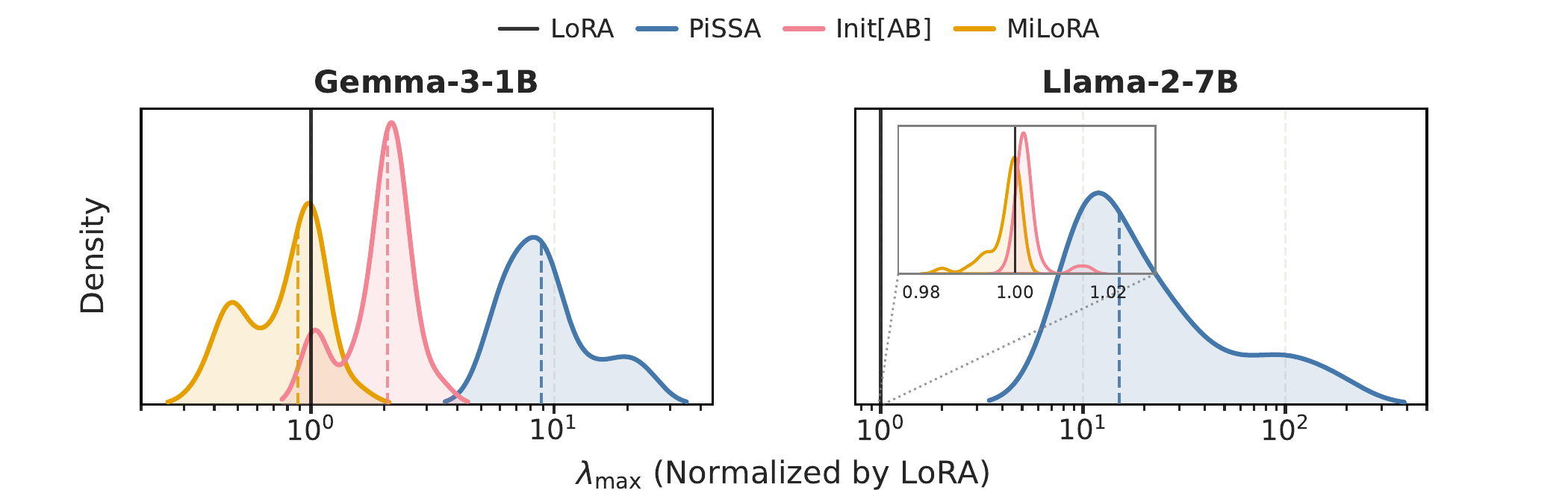}
    \caption{Distributions of the ratios of the top loss Hessian eigenvalues relative to LoRA for Query projection matrices across Transformer layers ($r = 128$). Dashed lines indicate the medians.}  \label{fig:hessian_gemma_llama_math_r128_dist}
\end{figure*}

It is evident that the Hessian relationship presented here is, again, generally negatively correlated with the optimal learning rate of each method. 
Specifically, PiSSA exhibits substantially larger $\lambda_{\max}$ on both models, aligning with its requirement for $1.8\mbox{--}2\times$ smaller learning rates in Table~\ref{tab:gemma-main} and Figure~\ref{fig:llama-main-math}.  
MiLoRA and Init[AB], on the other hand, have $\lambda_{\max}$ values that largely overlap with those of vanilla LoRA (especially on Llama-2-7B), which explains their similar optimal learning rate ranges to vanilla LoRA.  
These findings further support the use of relative Hessian magnitude as a useful indicator for explaining the observed performance differences and optimal learning rate ranges across different model scales.

\subsection{Detailed $\lambda_{\max}$ Values}\label{sec:detailed_lambda}

Figure~\ref{fig:hessian_qwen_math_r128_q} presents the detailed $\lambda_{\max}$ values of the Query projection matrix for Qwen across Transformer layers, providing a layer-wise breakdown of the results shown in Figure~\ref{fig:main_hessian}. We observe intriguing patterns in which all methods tend to exhibit high or low values at similar layer locations. For example, at layer 20, $\lambda_{\max}=
\{4.7, 8.5, 8.3, 53.8, 297.3, 264.7\}$ for LoRA, Init[AB], MiLoRA, PiSSA, OLoRA, and LoRA-GA, respectively, whereas at layer 26, these values drop to $\{0.2, 0.7, 0.7, 2.3, 12.3, 17.9\}$. However, at any given layer, OLoRA and LoRA-GA consistently exhibit substantially larger $\lambda_{\max}$ than LoRA, by around two orders of magnitude. PiSSA, in contrast, is larger than LoRA by roughly one order of magnitude.
Similar trends for the Key projection matrix are presented in Figure~\ref{fig:hessian_qwen_math_r128_k}.

\begin{figure*}[h]
    \centering
    \includegraphics[width=1.00\linewidth]{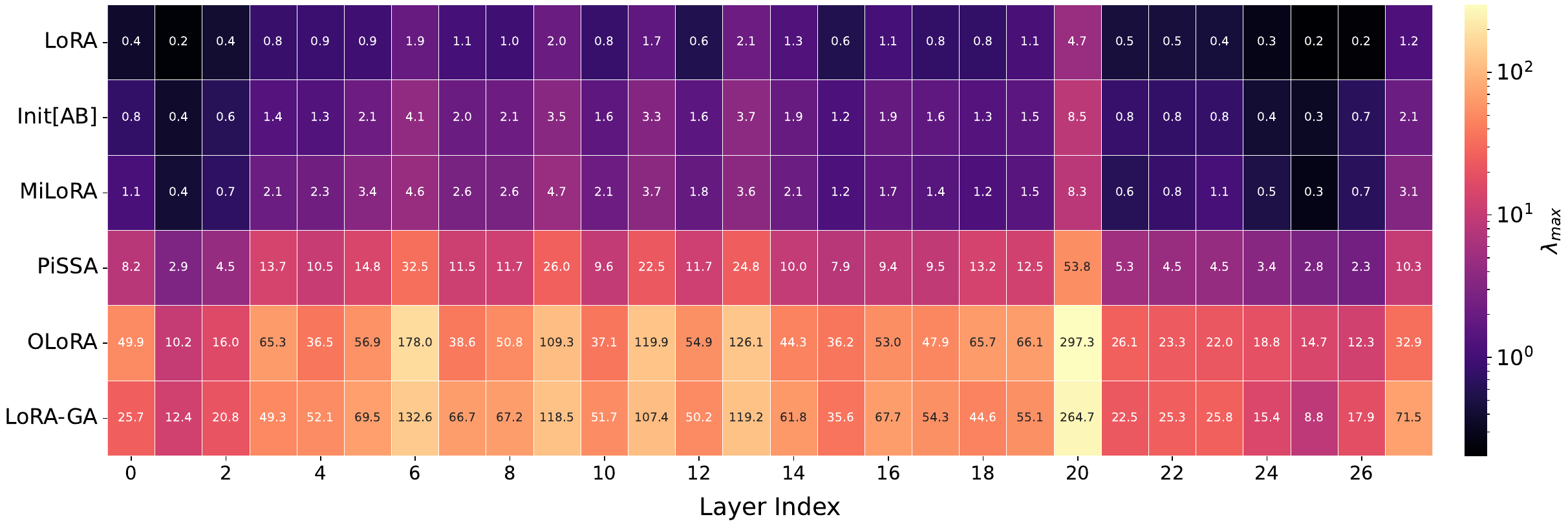}
    \caption{Heatmap of the top eigenvalues of the \textbf{Query projection matrix} across Transformer layers, i.e., $\lambda_{\max}^{Q,i}$ for $i=1,\ldots, L$, for Qwen3-0.6B on MetaMathQA ($r=128$). All methods exhibit similar distributional patterns across layers, with PiSSA, OLoRA and LoRA-GA consistently exhibiting significantly larger values compared to vanilla LoRA.}  \label{fig:hessian_qwen_math_r128_q}
\end{figure*}

\begin{figure*}[h]
    \centering
    \includegraphics[width=1.00\linewidth]{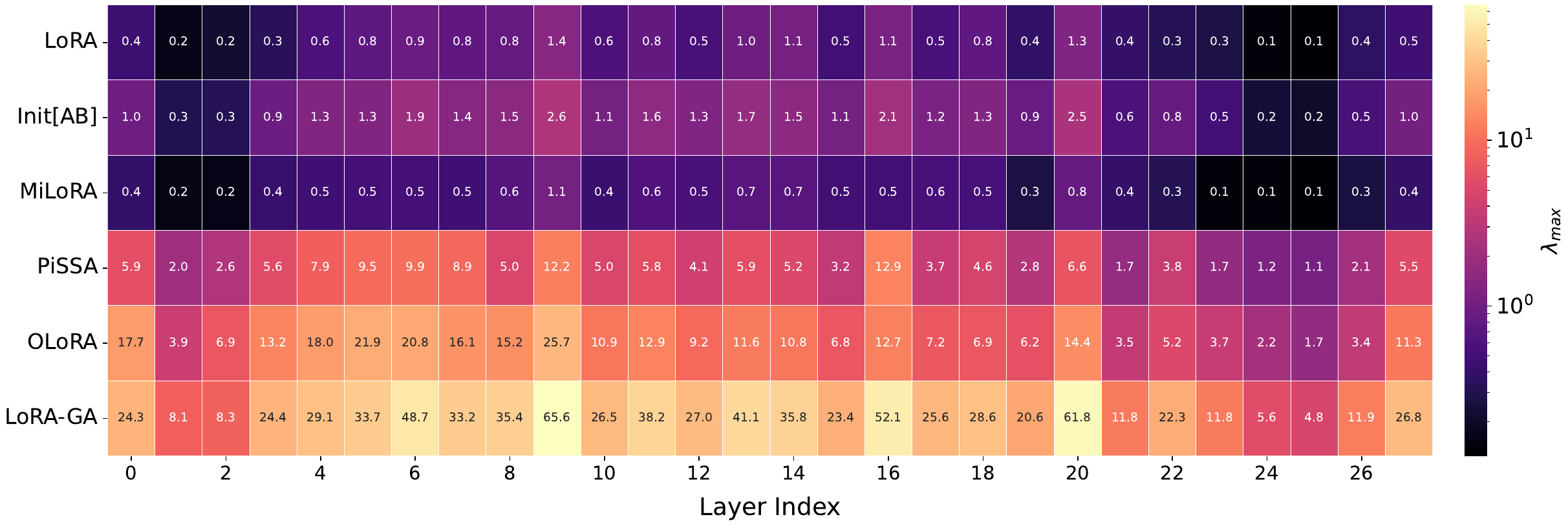}
    \caption{Heatmap of the top eigenvalues of the \textbf{Key projection matrix} across Transformer layers, i.e., $\lambda_{\max}^{K,i}$ for $i=1,\ldots, L$, for Qwen3-0.6B on MetaMathQA ($r=128$). All methods exhibit similar distributional patterns across layers, with PiSSA, OLoRA and LoRA-GA consistently exhibiting significantly larger values compared to vanilla LoRA.}  \label{fig:hessian_qwen_math_r128_k}
\end{figure*}

\FloatBarrier

}

\end{document}